%% file: main.tex
\documentclass[10pt,twocolumn,letterpaper]{article}

\usepackage[pagenumbers]{cvpr} 

\input{preamble}
\input{header/macro}

\definecolor{cvprblue}{rgb}{0.21,0.49,0.74}
\usepackage{booktabs}
\usepackage{multirow}
\usepackage{graphicx}
\usepackage[table]{xcolor}
\usepackage[pagebackref,breaklinks,colorlinks,allcolors=cvprblue]{hyperref}
\usepackage{acronyms}
\usepackage{subcaption}
\usepackage{comment}

\def\paperID{9045} 
\def\confName{CVPR}
\def\confYear{2026}

\title{\nickname: Sparse Voxel Rasterization for Surface Reconstruction}

\input{header/author}

\begin{document}
\input{figs/1_teaser/1_teaser}  
\input{sec/0_abstract}
\vspace{-8mm}
\input{sec/1_introduction}
\input{sec/2_related_work}

\input{sec/3_preliminary}

\input{sec/4_0_method}
\input{sec/4_1_method}
\input{sec/4_2_sparse_voxel_association}
\input{sec/4_3_sdf_encoding_in_sparse_voxel}

\input{sec/Results}
\input{sec/Conclusion}
{
    \small
    \bibliographystyle{ieeenat_fullname}
    \bibliography{main}
}

\appendix
\clearpage
\setcounter{page}{1}
\maketitlesupplementary
\input{supp/0_intro}
\input{supp/a_sparse_voxel_initialization}

\input{supp/b_loss}
\input{supp/c_additional_implementation_detail}

\input{figs/13_normal_until_end/13_normal_until_end}
\input{figs/14_real-world_result/14_real_world_result}
\input{supp/e_additional_ablation}
\input{supp/d_full_qualitative_result}

\end{document}

%% file: preamble.tex
\definecolor{yellow}{rgb}{1, 1, 0.7}
\definecolor{tableorange}{rgb}{1, 0.85, 0.7}
\definecolor{tablered}{rgb}{1, 0.7, 0.7}



%% file: header/macro.tex
\newcommand{\nickname}{SVRecon\xspace}
\newcommand{\noindentbold}[1]{\noindent\textbf{#1.}}

\definecolor{hilite}{RGB}{255,249,196} 

\newcommand{\set}[1]{\{#1\}}

%% file: header/author.tex
\author{
Seunghun Oh$^1$ \quad\;
Jaesung Choe$^2$ \quad\;
Dongjae Lee$^1$ \quad\;
Daeun Lee$^1$ \quad\;
Seunghoon Jeong$^1$,\\
Yu-Chiang Frank Wang$^2$ \quad\;
Jaesik Park$^1$\\
{$^{1}$Seoul National University} \quad
{$^{2}$NVIDIA}
}

%% file: figs/1_teaser/1_teaser.tex


\twocolumn[{%
    \maketitle
    \renewcommand\twocolumn[1][]{#1}%
    \vspace{-9mm} 
    \centering


    \includegraphics[width=\linewidth]{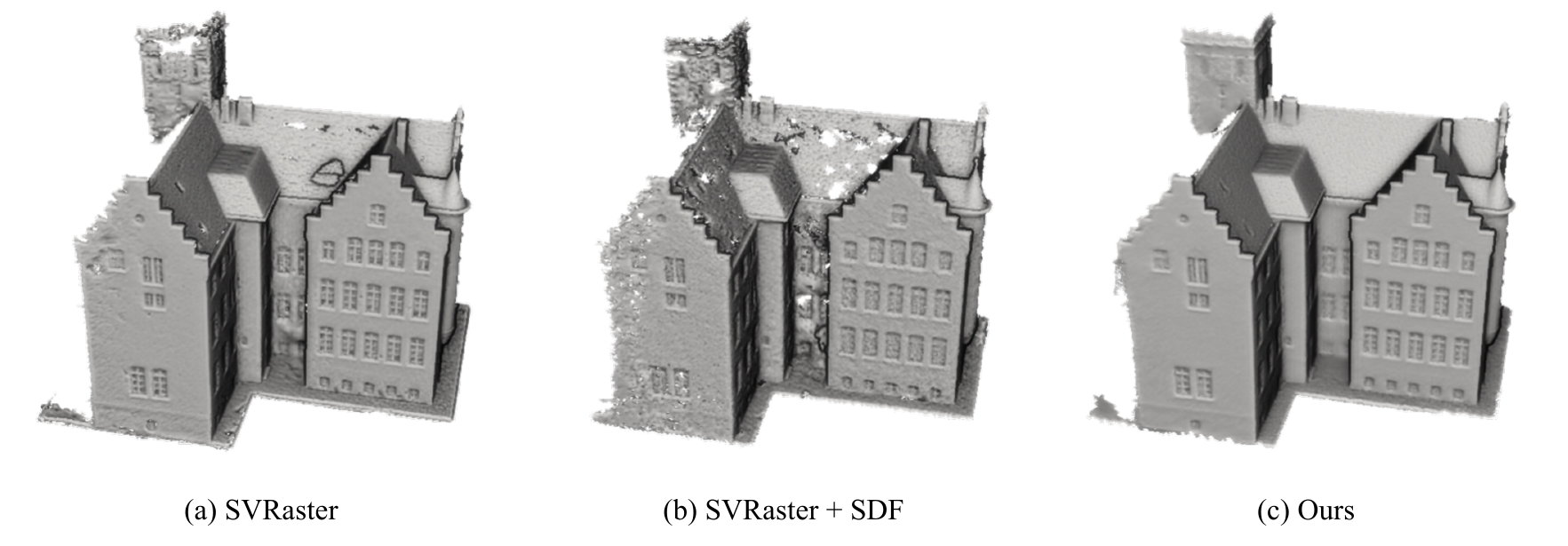}

    \vspace{-0.5em}
    \captionof{figure}{
        \nickname. This paper introduces Signed Distance Function (SDF) on top of the recent sparse voxel rasterization framework (SVRaster) for surface reconstruction. From left to right, the figure shows results from SVRaster~\cite{Sun_2025_CVPR_SVRaster}, a naive combination of SVRaster and NeuS~\cite{Wang_2021_NeurIPS_NeuS}, and our method, \nickname. While the naive extension exhibits significant high-frequency artifacts, our approach produces smooth and continuous surfaces. This improvement stems from addressing the lack of voxel-wise coherence in SVRaster.
    }
    \label{fig:teaser}    
    \vspace{7mm}
}]

%% file: sec/0_abstract.tex
\begin{abstract}
We extend the recently proposed sparse voxel rasterization paradigm to the task of high-fidelity surface reconstruction by integrating Signed Distance Function (SDF), named \textbf{\nickname}. Unlike 3D Gaussians, sparse voxels are spatially disentangled from their neighbors and have sharp boundaries, which makes them prone to local minima during optimization. Although SDF values provide a naturally smooth and continuous geometric field, preserving this smoothness across independently parameterized sparse voxels is nontrivial. To address this challenge, we promote coherent and smooth voxel-wise structure through (1) robust geometric initialization using a visual geometry model and (2) a spatial smoothness loss that enforces coherent relationships across parent-child and sibling voxel groups.
Extensive experiments across various benchmarks show that our method achieves strong reconstruction accuracy while having consistently speedy convergence. The code will be made public.
\end{abstract}

%% file: sec/1_introduction.tex
\section{Introduction}
\label{sec:introduction}

Neural rendering has rapidly evolved as a central paradigm for reconstructing and synthesizing 3D scenes from multi-view images. Since the introduction of \ac{NeRF} \cite{Mildenhall_2020_ECCV_NeRF}, learning-based scene representations have advanced toward more efficient, compact, and high-quality 3D primitives. Notably, explicit representations such as \ac{3DGS} \cite{Kerbl_2023_ACM_3DGS} have demonstrated state-of-the-art performance in real-time novel view synthesis, revealing the potential of structured primitives in neural rendering pipelines. More recently, \ac{SVRaster} \cite{Sun_2025_CVPR_SVRaster} has been proposed as an alternative representation that directly operates on sparse voxel elements and enables differentiable rasterization without the heavy reliance on neural networks. This line of work envisions the possibility of leveraging sparse voxel structures as a powerful, lightweight representation for neural rendering tasks.

The original \ac{SVRaster} framework employs `\textit{density}' as the underlying geometric quantity for representing the 3D scenes. 
While density fields offer compatibility with volumetric rendering, they are not inherently designed to capture sharp surfaces or precise geometry. Density values tend to blur object boundaries and require delicate regularization to avoid over-smoothed reconstructions. In contrast, \ac{SDF} is a widely adopted representation for surface reconstruction, as it encodes geometry in a continuous, metric-consistent manner where the zero-level set defines the target surface. 
\ac{SDF} naturally promotes smoothness and spatial coherence~\cite{Wang_2021_NeurIPS_NeuS,Fu_2023_CVPR_NeuS2,Yariv_2021_VolSDF,deepsdf}, making them an appealing alternative to density-based geometry modeling.



However, we observe that a naive extension of \ac{SVRaster}~\cite{Sun_2025_CVPR_SVRaster} to SDF-based reconstruction~\cite{Wang_2021_NeurIPS_NeuS} often falls into severe local minima during optimization as shown in \cref{fig:teaser}. A key factor behind this instability is the hierarchical subdivision strategy of the original SVRaster, which rapidly refines coarse voxels into fine-grained \ac{LoD} voxels. This progressive voxel subdivision magnifies inconsistencies when learning \ac{SDF} across independently parameterized voxel units. More importantly, sparse voxels possess distinct and isolated boundaries: each voxel maintains its own parameters without explicit constraints linking it to spatially adjacent voxels. This property sharply contrasts with 3D Gaussians, whose kernels overlap smoothly, or neural field methods, where MLPs naturally enforce spatial continuity. As a result, \ac{SDF} values in sparse voxels can become discontinuous across boundaries, causing the optimization to converge to degenerate or fragmented reconstructions results as shown in \cref{fig:teaser}-(b). 

To address these issues, we propose \nickname, a robust SDF-based extension of sparse voxel rasterization designed specifically for high-fidelity surface reconstruction. Our key observation is that spatial coherence is crucial for representing SDFs within the sparse voxel framework. First, we initialize the voxel geometry using point maps predicted by recent visual geometry models~\cite{wang2025vggt,wang2025pi3,must3r,dust3r,vipe} that provide superior geometry priors and significantly reduce ambiguity in early optimization. Second, we introduce a spatial smoothness loss that explicitly enforces consistent relationships among parent-child and sibling voxel groups, encouraging neighboring voxels to share compatible SDF values and preventing discontinuities at voxel boundaries. Unlike the original SVRaster, which treats voxels as independent rendering units, our approach leverages cross-voxel coherence as a fundamental requirement for accurate SDF modeling. Through this combination of informed initialization and structured smoothness constraints, our method achieves 
high-quality surface reconstruction results. The contributions of this paper are summarized as below:
\begin{itemize}
    \item Introduce Signed Distance Function into the sparse voxel rasterization, enabling accurate surface reconstruction.
    \item Address an inherent discontinuity issue in sparse voxels and introduce two key components to maintain coherent SDF values across voxel boundaries: (1) Voxel initialization via visual geometry models. (2) A spatial smoothness loss on parent-child and child-child voxel groups.
    \item Achieve fast, high-quality, and accurate surface reconstruction on DTU in under 5 minutes and on Tanks-and-Temples in under 15 minutes.
\end{itemize}


%% file: sec/2_related_work.tex
\section{Related works}
\label{sec:related_works}

\noindentbold{3D Primitives for Neural fields}
Neural rendering has evolved through a wide range of 3D representations, each offering distinct trade-offs between fidelity, efficiency, and scalability. Early approaches such as \ac{NeRF} model scenes implicitly through MLPs, enabling photorealistic novel view synthesis but requiring slow volumetric integration and expensive per-scene optimization. 
To address this issue, researchers have explored more explicit 3D primitives, including point clouds~\cite{pointnerf}, voxel grids~\cite{dvgo,li2023nerfacc,plenoctrees}, meshes~\cite{reiser2024binary,wang2023adaptive,yariv2023bakedsdf} or hashgrid~\cite{Muller_2022_ACM_InstantNGP,Fu_2023_CVPR_NeuS2,dong2023fast} to represent geometry in a structured manner.
Recently, Gaussian representations~\cite{Kerbl_2023_ACM_3DGS,Huang_2024_SIGGRAPH_2DGS} have shown remarkable rendering accuracy by representing scenes as anisotropic Gaussians with learnable attributes, enabling real-time rendering operation. 3D Gaussians benefit from spatial overlap and smooth blending, which naturally provide geometric continuity. 

On the other hand, a series of works~\cite{Wang_2021_NeurIPS_NeuS,Yariv_2021_VolSDF,monosdf,li2024monogsdf,geoneus} leverage SDF for the surface reconstruction task by modeling geometry as a continuous field whose zero-level set defines the surface. SDF-based methods are widely used for accurate geometry recovery due to their metric consistency and ability to represent sharp boundaries~\cite{Wang_2021_NeurIPS_NeuS}. Despite advancements, a key challenge remains: identifying a representation that is both computationally efficient and capable of expressing fine geometric detail.


\noindentbold{Sparse voxel representation}
Voxel-based methods offer a natural discretization of 3D space and have been widely used in classical graphics and more recent neural rendering pipelines. Dense voxel grids, however, suffer from cubic memory growth and limited scalability~\cite{dvgo}. Sparse voxel structures mitigate these issues by allocating voxels only where needed, allowing high-resolution modeling with manageable computational cost~\cite{Sun_2025_CVPR_SVRaster,minkunet,ptv3,liu2015sparse,choe2021deep}.
\ac{SVRaster}~\cite{Sun_2025_CVPR_SVRaster} has been recently introduced as an explicit neural rendering framework that uses sparse voxels as its fundamental 3D primitive. By enabling efficient differentiable rasterization with hierarchical structure, \ac{SVRaster} provides an alternative to implicit radiance fields and Gaussian Splatting for representing the 3D scenes. However, \ac{SVRaster} originally relies on density as measurements, which can lead to ambiguous geometry encoding.

%% file: sec/3_preliminary.tex
\section{Preliminary}
\label{sec:preliminary}

\noindent\textbf{SVRaster}~\cite{Sun_2025_CVPR_SVRaster} introduce a tile-based rasterization algorithm into the sparse voxel representation for the novel view synthesis task. Voxels are hierarchically subdivided to have different \ac{LoD}. Each voxel $v$ consists of 8 corners~$geo_v \in \mathbb{R}^{2\times 2\times 2}$ where each corner involves a density value. Then $geo_v$ is interpolated and interpreted as an $\alpha$ to be used in the volumetric rendering~\cite{Mildenhall_2020_ECCV_NeRF} as in \cref{eq:vol_color} where $N_{ray}$ is the number of transmitted voxels along the ray, and $\mathbf{c}_i$ is the color of $i$-th voxel on the ray.
\begin{equation}\label{eq:vol_color}
\mathbf{C} = \sum_{i=1}^{N_{ray}} T_i \alpha_i \mathbf{c}_i~~~\text{where}~~~T_i = \prod_{j=1}^{i-1} (1-\alpha_j).
\end{equation}

\noindent\textbf{NeuS}~\cite{Wang_2021_NeurIPS_NeuS} proposes SDF-based $\alpha$-blending into the neural surface reconstruction task unlike NeRF~\cite{Mildenhall_2020_ECCV_NeRF} that utilizes density as a geometry measurements.
Given an SDF field $f(\mathbf{x})$ from a query point~$\mathbf{x}$,
NeuS transforms the SDF value $f$ into an opacity value $\alpha_i$ along a ray using a logistic cumulative density function $\Phi_s(f)=\tfrac{1}{1+e^{-s f}}$ as:
\begin{equation} \label{eq:neus_alpha}
\alpha_i = \max \left(
\frac{\Phi_s(f(t_\text{i})) - \Phi_s(f(t_\text{i+1}))}{\Phi_s(f(t_\text{i}))},~0
\right),
\end{equation}
where $s$ is a reciprocal of standard deviation of density function $\phi_s(f)$ and $t_{\mathrm{i}}$ is the $i$-th sampled point along the ray. This closed-form $\alpha$ formulation produces accurate and consistent surface supervision, leading to high-fidelity meshes.
Then, this method is trained to minimize the rendering loss $\mathcal{L}_\mathrm{render}=|\mathbf{C} - \Tilde{\mathbf{C}}|_2$ as well as the Eikonal loss~$\mathcal{L}_\mathrm{eik}$~\cite{eikonal_loss} which is computed as
$\mathcal{L}_\mathrm{eik} = \big(\left\| \bigtriangledown_x f \right\|_2 -1 \big)^2$
where $\bigtriangledown_x f(x)$ is the partial differential of a SDF value $f(x)$ over query point x, which is the same as the surface normal at $x$.
Despite the superior results, NeuS suffers from slow training convergence. NeuS2~\cite{Fu_2023_CVPR_NeuS2} mitigates this issue by incorporating a hash-grid representation, yet it remains unexplored how the sparse voxel structure of SVRaster~\cite{Sun_2025_CVPR_SVRaster} can be combined with NeuS-style SDF modeling.

\input{figs/2_pipeline/2_pipeline}

\noindentbold{Naive combination of SVRaster and NeuS}
We first modify SVRaster to repurpose the voxel-corner storage $geo_v$ to store SDF values instead of densities. 
The SDF value $f(\mathbf{p})$ at any continuous 3D point $\mathbf{p}$ inside a voxel $v$ is then obtained by
trilinear interpolation over its $geo_v$:
\begin{equation}\label{sdf_from_voxel}
    f(\mathbf{p}) = \operatorname{interp}(geo_v, \mathbf{q})~~~\text{where}~~~\mathbf{q} = \frac{\mathbf{p} - v_{\min}}{v_l},
\end{equation}
where $v_{\min}$ denotes the voxel corner closest to the world origin, $v_l$ is the voxel side length,
and $\mathbf{q}$ is the local coordinate of $\mathbf{p}$ within $v$.

Given sparse voxels parameterized by corner SDF values as in \cref{sdf_from_voxel}, 
a straightforward extension for learning surfaces from multi-view images is to adopt the NeuS 
SDF-based $\alpha$-blending formulation~\cref{eq:neus_alpha} and evaluate $f(\cdot)$ at the ray-voxel 
entry and exit points $t_i$ and $t_{i+1}$. However, as shown in~\cref{fig:teaser}, we empirically found that such an approach demonstrates high-frequency noise, fragmented surfaces, and inconsistent level-set structures in results.

We observe that a key limitation of this naive extension stems from its independently parameterized sparse voxels in SVRaster~\cite{Sun_2025_CVPR_SVRaster}, both across spatial neighbors and hierarchical levels. Unlike implicit neural fields~\cite{Wang_2021_NeurIPS_NeuS,Yariv_2021_VolSDF,eikonal_loss} or Gaussian-based primitives~\cite{Kerbl_2023_ACM_3DGS,Huang_2024_SIGGRAPH_2DGS}, these voxels possess hard boundaries and lack any built-in spatial coupling. When SDF values are assigned per voxel, the absence of cross-voxel coherence hinders smooth geometric transitions and makes the optimization highly prone to local minima. To address this issue, we introduce spatial coherence mechanisms and structured initialization strategies that enable SDF-involving sparse voxels to act as reliable and high-fidelity geometric primitives.

%% file: figs/2_pipeline/2_pipeline.tex
\begin{figure*}[t]
    \centering
    \includegraphics[width=1.0\textwidth]{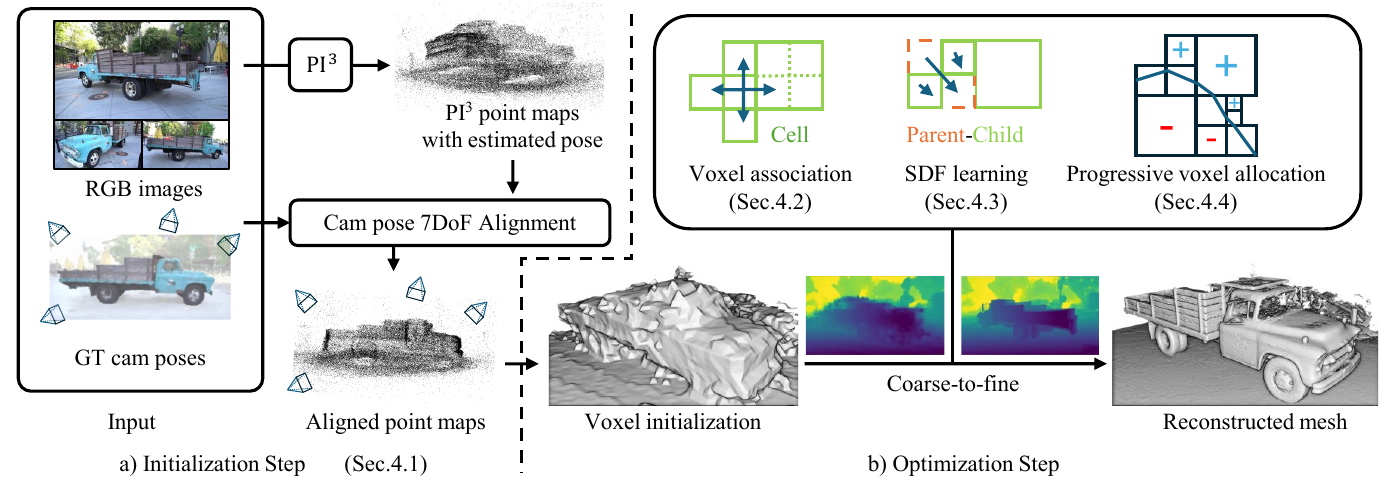}
    \vspace{-6mm}
    \caption{Our pipeline consists of two main stages: (a) Initialization and (b) Optimization.
    In (a), we initialize a coarse SDF from multi-view images by aligning PI$^3$ point maps~\cite{wang2025pi3} at estimated camera poses into points at ground truth camera poses (\cref{subsec:sparse_voxel_initialization_using_pointmap}).
    In (b), we impose 
        spatial coherence to ensure voxel continuity via inter-voxel association (\cref{subsec:sparse_voxel_association}),
        parent-child Eikonal/smoothness losses (\cref{subsec:sdf_encoding_in_sparse_voxel}),
        and progressive voxel allocation (\cref{subsec:sdf_encoding_in_sparse_voxel}). 
    Finally, our model yields a clean final mesh in minutes.
    }
    \label{fig:placeholder}
\end{figure*}

%% file: sec/4_0_method.tex
\section{Methodology}
\label{sec:methodology}

Given multi-view images and the corresponding camera poses, we aim to reconstruct 3D scenes using sparse voxel representation with \ac{SDF}. 
Our method proposes ways of imposing the surface smoothness by voxel initialization (\cref{subsec:sparse_voxel_initialization_using_pointmap}), voxel-wise association (\cref{subsec:sparse_voxel_association}).
Finally, our method is trained to minimize the losses to represent the surface geometry (\cref{subsec:sdf_encoding_in_sparse_voxel}).

%% file: sec/4_1_method.tex
\subsection{Voxel initialization}
\label{subsec:sparse_voxel_initialization_using_pointmap}

Geometry initialization plays a crucial role in optimizing neural fields, especially when employing explicit representations such as 3D Gaussians~\cite{Kerbl_2023_ACM_3DGS} or sparse voxels~\cite{Sun_2025_CVPR_SVRaster}.
While density-based methods like SVRaster typically start from an empty or near-zero space, our SDF-based formulation enables faster and more efficient convergence by directly utilizing rich geometric priors from the beginning.
To achieve this stable and physically meaningful initialization, we leverage PI$^3$~\cite{wang2025pi3}, a visual geometry model that estimates 3D points in world coordinates from unposed images. 
We transform the point maps from PI$^3$~\cite{wang2025pi3} into points at ground truth camera poses using Umeyama's method~\cite{umeyama}.
Then, we allocate the voxels at the transformed points and compute the SDF of each voxel by computing the minimum distance from the transformed points. An example of the resulting initialized SDF field is shown as part of our pipeline in \cref{fig:placeholder}. Details are included in the supplementary material. 

%% file: sec/4_2_sparse_voxel_association.tex
\subsection{Voxel Association}
\label{subsec:sparse_voxel_association}

To consistently apply spatial regularizers (e.g., Eikonal, smoothness, Laplacian) on SVRaster's multi-resolution sparse voxel structure, we introduce a novel voxel association mechanism. This is challenging because, unlike hashmap-based approaches~\cite{minkunet,ptv3} that assume a uniform grid, SVRaster's voxels are multi-resolution with different LoD.
We resolve this by reasoning on a single conceptual \emph{finest} lattice at the $G=2^L$ resolution, indexed by $\mathbf{d_L}$ where $\mathbf{d_L}{=}(i,j,k){\in}\{0,\ldots,G{-}1\}^3$ indicates a dense coordinate in the finest lattice (level $L$). We define a `cell' $c(i,j,k)$ as the unit cube volume associated with a dense coordinate $\mathbf{d_L}=(i,j,k)$ in this lattice, covering the continuous 3D region: $c(i,j,k) {=} \{ (x, y, z) {\in} \mathbb{R}^3 \mid i {\le} x {\le} i{+}1, j {\le} y {\le} j{+}1, k {\le} z {\le} k{+}1 \}$.
All neighborhood regularizers use this cell as a fundamental unit.
Note that we do not store per-cell data for the entire $G^3$ lattice. Instead, we keep a \emph{bit-level} occupancy map to compact and track only those cells that are actually covered by any voxel. In effect, the fine grid is a conceptual scaffold; computations and memory scale with the occupied region, not with $G^3$.

\noindent\textbf{Data structures.}
To manage sparse voxels across different LoDs, we assume the conceptual voxel grid at the finest LoD and introduce three lightweight data structures that link it to the sparse voxels:

\begin{enumerate}
\item \textbf{Occupancy bitmask (A)}: A bitmask of length $
G^3$ with $G{=}2^L$. Entry $\mathrm{idx}(x,y,z){=}x{\cdot}G^2{+}y{\cdot}G{+}z$ is linearized dense index for a coordinate $[i, j, k]$, which marks whether the corresponding cell $c(x,y,z)$ is occupied by an active voxel $v$. This enables constant-time queries and a single parallel scan to list all occupied grids.
\item \textbf{Index table (B)}: A sorted \textbf{lookup table} of length $M$ $(M\ll G^3)$ that holds the indices $\mathrm{idx}$ of occupied grids.
\item \textbf{Grid$\to$Voxel map (C)}: A \textbf{lookup table} of length $M$, aligned with (B), that links each occupied grid cell to its enclosing voxel. Via (B)+(C), we locate the corresponding grid and voxel for any query point, collision-free.
\end{enumerate}
\noindent With (A)-(C), we achieve fine-grid behavior while storing only $M$ occupied grid cells, which keeps memory and compute scale with $M$, not $G^3$.
The data structures are updated only when the LoD hierarchy changes.


\input{figs/5_nvs+parent-child/5_nvs+parent-child}

\noindent\textbf{Hierarchy updates.}
When the LoD hierarchy changes due to voxel subdivision, we rebuild the data structures in a simple, parallel fashion.
Each thread group processes a single active voxel, iterating over its covered fine-grid cells to compute their indices.
These indices are marked in the occupancy bitmask (A), and corresponding entries are added to the temporary buffers for the index table (B) and voxel map (C), both initially unsorted.
We then sort (B) and apply the same permutation to (C), producing the final lookup tables.
This update procedure is efficient: although it operates at fine-grid resolution, it processes only occupied cells and is fully parallelized.

\noindent\textbf{Neighboring-Voxel Search.}
For a query point $\mathbf{q}$, we first locate its containing grid cell $c(x,y,z)$ and index $\mathrm{idx}$. We then examine the six face-adjacent grid cells 6-connected neighbors): $(x\!\pm\!1,y,z)$, $(x,y\!\pm\!1,z)$, and $(x,y,z\!\pm\!1)$. Neighbor lookup proceeds as follows:
\begin{enumerate}
\item \textbf{Neighbor index}: Compute $\mathrm{idx}$ for each of the six face-adjacent neighbors.
\item \textbf{Occupancy lookup}: Use the occupancy bitmask (A) to check whether each neighbor is active. This requires a single $O(1)$ bitwise access per neighbor.
\item \textbf{Index lookup}: If $A[\mathrm{idx}] = \mathrm{true}$, the neighbor is occupied. Locate its position $i$ in the index table (B), which stores all occupied indices in sorted order. This step uses binary search and takes \(O(\log M)\) time, where $M$ is the number of occupied grids.
\item \textbf{Voxel retrieval}: Using index $i$, retrieve the corresponding voxel via $C[i]$ in $O(1)$ time.
\end{enumerate}
\noindent These processes are batched across many query points on the GPU. We will simplify this lookup process with the following notation as
$v = \mathrm{NVS}[\mathrm{idx}]$
where $\mathrm{idx}$ is the dense cell index being queried, and Neighboring Voxel Search (NVS) is the function that returns the index $v$ of the SVRaster voxel that occupies the given dense cell, implemented by the steps above.

%% file: figs/5_nvs+parent-child/5_nvs+parent-child.tex
\begin{figure*}
    \centering
    \includegraphics[width=1.0\linewidth]{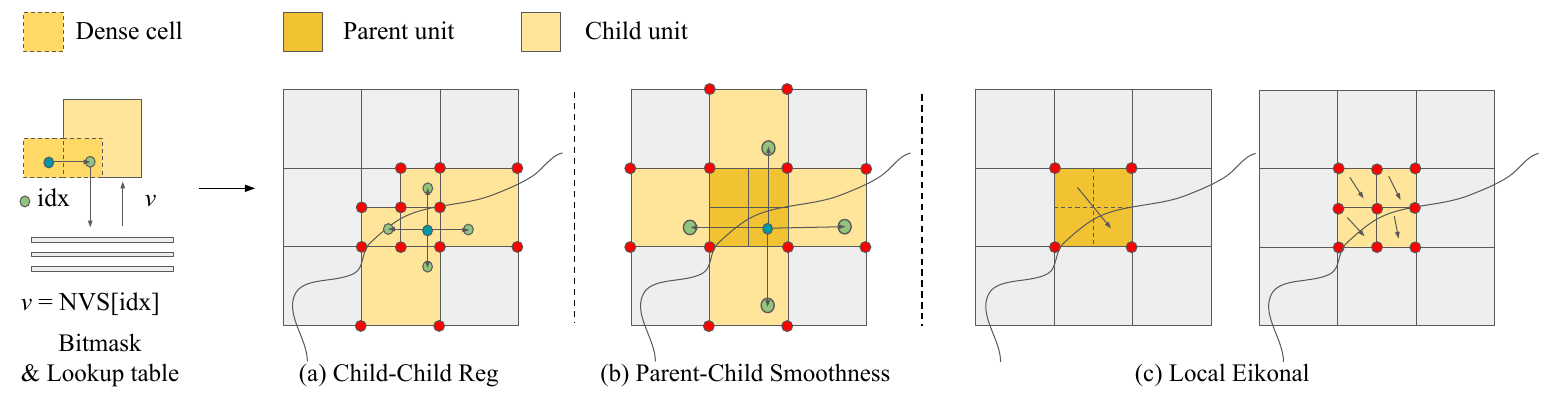}
    \vspace{-2mm}
    \caption{Coherent voxel structure. (a) illustrates child-child voxel regularization using fine cell unit. After $2^9$ resolution, all voxel corners are efficiently connected due to hierarchical regularization (b) and (c). Red voxel corners (\textcolor{red}{$\bullet$}) are associated through the continuity loss. }
    \label{fig:NVS&parent-child}
\end{figure*}

%% file: sec/4_3_sdf_encoding_in_sparse_voxel.tex
\subsection{SDF Learning in Sparse Voxels}
\label{subsec:sdf_encoding_in_sparse_voxel}

After the voxel initialization, our method follows the SDF-based alpha blending by NeuS~\cite{Wang_2021_NeurIPS_NeuS} and rasterization by SVRaster~\cite{Sun_2025_CVPR_SVRaster}.
Our method is trained to minimize the rendering loss, but we additionally introduce hierarchical regularization loss and sharpness scheduling in SDF learning.

\noindentbold{Hierarchical regularization for fine LoD}
The Voxel association mechanism described in \cref{subsec:sparse_voxel_association} relies on data structures (A, B, C) that scale with the dense grid resolution $G$. At resolutions finer than $L=9$ (e.g., $G=512$), the $G^3$ dense grid becomes prohibitively large, making the global occupancy bitmask (A) and lookup tables (B, C) intractable. To resolve this, we introduce a hierarchical regularization strategy.

For \emph{parent-child smoothness}, we cap the global association lattice and its data structures at a maximum level, $L_{cap}=9$. When SVRaster subdivides voxels beyond this level (e.g., to $L=10$), we no longer update the global data structures. While rendering operates on the fine child voxels, the smoothness loss (Laplacian) is instead accumulated at the $L=9$ parent level (where NVS is still valid) and then propagated to its descendants. Thus, the parent-level data structure needs not be subdivided beyond $L=9$, yet it still enforces cross-child consistency at finer levels.
%
Moreover, for levels $L \ge 9$ we impose a local per-voxel Eikonal loss on all active voxels in the tree. When refining from $L=10$ to $L=11$, the $L=10$ voxels also remain as internal parents and continue to receive this local regularization, so both parent and child cells are stabilized without any global NVS lookup, which is infeasible at these resolutions.

The combination of (i) parent-child smoothness and (ii) local Eikonal yields a hierarchical, memory-efficient regularizer. It allows subdivision to $L=10$ and $L=11$ resolution \emph{without losing continuity} across split faces, while keeping the primary compute and memory cost bounded to the $L=9$ association lattice and local stencils.

\input{table/05_table1}
\input{figs/7_DTU_comparision/7_DTU_comparision}

\noindent\textbf{Scheduling sharpness.}
With the NeuS logistic CDF $\Phi_s(f){=}\tfrac{1}{1+e^{-s f}}$
we define the
\emph{learning thickness} $\ell(s)$ as the surface band that captures $99\%$ of weight as $\ell(s){\approx}\frac{2\ln 199}{s}$, $\tilde{\ell}(s){=}\frac{\ell(s)}{h_{L}}$
Where $h_{L}$ is the minimum voxel size in the scene resolution $2^{L}$. We keep $\tilde{\ell}(s)$ roughly constant across coarse$\to$fine refinements by
\emph{monotonically} increasing $\log s$. Concretely, within each octree level $L$ we
ramp as
$\log s(\tau){=}\log s_L{+}0.07{\cdot}r_L(\tau)$, $r_L{:}0{\to}1$,
and when moving to the next level we update the base
$\log s_{L+1}{=}\log s_L{+}\ln\bigl(h_L/h_{L+1}\bigr)$
(so if $h$ halves, $\log s$ rises by $\ln 2$).
This keeps the $99\%$-mass band $\ell(s)$ proportional to the minimum voxel size,
yielding stable early training (thick bands) and progressively sharper surfaces.

In conclusion, our method is trained to minimize the training loss that is computed as
$L_\mathrm{total} = L_\mathrm{photo} + \lambda_n L_\mathrm{normal}  + \lambda_eL_\mathrm{eik}+ \lambda_sL_\mathrm{smooth} +\lambda_mL_\mathrm{mask}.$
Note that mask loss $L_\mathrm{mask}$ is only applied to DTU dataset following~\cite{jensen2014large} More detailed descriptions about losses is in the supplementary material.

\subsection{Progressive Voxel Allocation}
To dynamically adapt the sparse voxel structure during training, we employ a loop of pruning and subdivision steps.

\noindentbold{Pruning}
Every 1000 iterations, we prune voxels that are unlikely to contribute to the surface. A voxel $v$ is removed if it meets two conditions: (1) all eight of its corner SDF values ($geo_v$) share the same sign (i.e., it contains no zero-crossing), and (2) all corner SDF magnitudes lie outside the current learning thickness band ${\ell}(s)$ (defined in \cref{subsec:sdf_encoding_in_sparse_voxel}). As training progresses, $s$ increases, this band narrows, and pruning becomes more aggressive. To avoid over-pruning at coarse stages, we keep a small safety margin tied to the current voxel size $h_v$.

\noindentbold{Subdivision}
Conversely, we subdivide voxels every 250 iterations to add geometric detail where needed. We only split voxels that (1) are likely on the surface (either containing mixed corner signs or having corners within the learning thickness band) and (2) are not yet at the finest allowed level. To cap memory growth, this process is modified when refining from $L=9$ to $L=10$. Instead of splitting all eligible voxels, we select only the highest-loss fraction (top $q\%$) and refine them. This keeps refinement focused where it matters most while controlling the memory.



%% file: table/05_table1.tex
\begin{table*}[t]
\caption{
\textbf{Quantitative comparison on the DTU Dataset}~\cite{dtu}. We measure the Chamfer Distance for evaluation, which is lower the better.
}
\label{tab:dtu}
\vspace{-2mm}
\setlength\tabcolsep{6.5pt}
\resizebox{\linewidth}{!}{%
\begin{tabular}{l|ccccccccccccccc|c|c}
\toprule
Method & 24 & 37 & 40 & 55 & 63 & 65 & 69 & 83 & 97 & 105 & 106 & 110 & 114 & 118 & 122 & \textit{Mean} & \textit{Time} \\ 

\midrule

\multicolumn{10}{l}{\textbf{Implicit Representation.}} \\
VolSDF \cite{Yariv_2021_VolSDF} & 1.14 & 1.26 & 0.81 & 0.49 & 1.25 & 0.70 & 0.72 & 1.29 & 1.18 & 0.70 & 0.66 & 1.08 & 0.42 & 0.61 & 0.55 & 0.86 & $>$ 12h \\
NeuS \cite{Wang_2021_NeurIPS_NeuS} & 1.00 & 1.37 & 0.93 & 0.43 & 1.10 & 0.65 & 0.57 & 1.48 & 1.09 & 0.83 & 0.52 & 1.20 & 0.35 & 0.49 & 0.54 & 0.84 & $>$ 12h \\
Neuralangelo \cite{Li_2023_CVPR_Neuralangelo} & \textbf{0.37} & 0.72 & 0.35 &  \textbf{0.35} & 0.87 &  0.54 & 0.53 & 1.29 & 0.97 & 0.73 &  0.47 & 0.74 & 0.32 & 0.41 & 0.43 & 0.61 & $>$ 128h \\
GeoNeuS \cite{geoneus} & 0.38 &  \textbf{0.54} &  \textbf{0.34} & 0.36 & \textbf{0.80} &  \textbf{0.45} &  \textbf{0.41} &  \textbf{1.03} & \textbf{0.84} & \textbf{0.55} & \textbf{0.46} &  \textbf{0.47} &  \textbf{0.29} &  \textbf{0.36} &  \textbf{0.35} & \textbf{0.51} & $>$ 12h \\
MonoSDF \cite{monosdf} & 0.66 & 0.88 & 0.43 & 0.40 & 0.87 & 0.78 & 0.81 & 1.23 & 1.18 & 0.66 & 0.66 & 0.96 & 0.41 & 0.57 & 0.51 & 0.73 & 6h \\ 

\midrule

\multicolumn{10}{l}{\textbf{Explicit Representation (Gaussian).}} \\
2DGS \cite{Huang_2024_SIGGRAPH_2DGS} & 0.48 & 0.91 & 0.39 & 0.39 & 1.01 & 0.83 & 0.81 & 1.36 & 1.27 & 0.76 & 0.70 & 1.40 & 0.40 & 0.76 & 0.52 & 0.80 & 0.2h \\
GOF \cite{gof} & 0.50 & 0.82 & 0.37 & 0.37 & 1.12 & 0.74 & 0.73 & 1.18 & 1.29 & 0.68 & 0.77 & 0.90 & 0.42 & 0.66 & 0.49 & 0.74 & 1h \\
GS2Mesh \cite{wolf2024gs2mesh} & 0.59 & 0.79 & 0.70 & 0.38 & \textbf{ 0.78} & 1.00 & 0.69 & 1.25 & 0.96 &  0.59 & 0.50 & 0.68 & 0.37 & 0.50 & 0.46 & 0.68 & 0.3h \\
VCR-GauS \cite{chen2024vcr} & 0.55 & 0.91 & 0.40 & 0.43 & 0.97 & 0.95 & 0.84 & 1.39 & 1.30 & 0.90 & 0.76 & 0.92 & 0.44 & 0.75 & 0.54 & 0.80 & $\sim$1h \\
MonoGSDF \cite{li2024monogsdf} & 0.45 & 0.65 & \textbf{0.36} & 0.36 & 0.94 & 0.70 & 0.67 & 1.27 & 0.99 & 0.63 & 0.49 & 0.84 & 0.39 & 0.53 & 0.47 & 0.65 & hrs \\
PGSR \cite{chen2024pgsr} & \textbf{0.36} &  \textbf{0.57} & 0.38 & \textbf{0.33} &  \textbf{0.78} & \textbf{0.58} &  \textbf{0.50} & \textbf{1.08} &  \textbf{0.63} &  \textbf{0.59} &  \textbf{0.46} &  \textbf{0.54} &  \textbf{0.30} & \textbf{0.38} & \textbf{0.34} & \textbf{0.52} & 0.5h \\

\midrule

\multicolumn{10}{l}{\textbf{Explicit Representation (Sparse Voxel).}} \\
SVRaster \cite{Sun_2025_CVPR_SVRaster} & 0.61 & 0.74 & 0.41 & \textbf{0.36} & \textbf{0.93} & 0.75 & 0.94 & 1.33 & 1.40 & \textbf{0.61} & 0.63 & 1.19 & 0.43 & 0.57 & \textbf{0.44} & 0.76 & 5min \\
\nickname (ours) & \textbf{0.40} &  \textbf{0.67} &  \textbf{0.31} &  0.37 & \textbf{0.93} &  \textbf{0.72} &  \textbf{0.68} &  \textbf{1.32} &  \textbf{1.21} & 0.69 & \textbf{0.51} &  \textbf{0.85} &  \textbf{0.40} &  \textbf{0.55} & 0.51 & \textbf{0.67} & 5min \\ 

\bottomrule

\end{tabular}%
}
\vspace{-2mm}
\end{table*}

%% file: figs/7_DTU_comparision/7_DTU_comparision.tex

\captionsetup[subfigure]{labelformat=empty} 

\newcommand{\colhead}[1]{\begin{subfigure}{.19\textwidth}\centering\normalsize\textnormal{#1}\end{subfigure}}
\newcommand{\cellimg}[1]{%
  \begin{subfigure}[b]{.19\textwidth}\centering
    \includegraphics[width=\linewidth]{#1}
  \end{subfigure}%
}

\begin{figure*}[t]
\centering


\cellimg{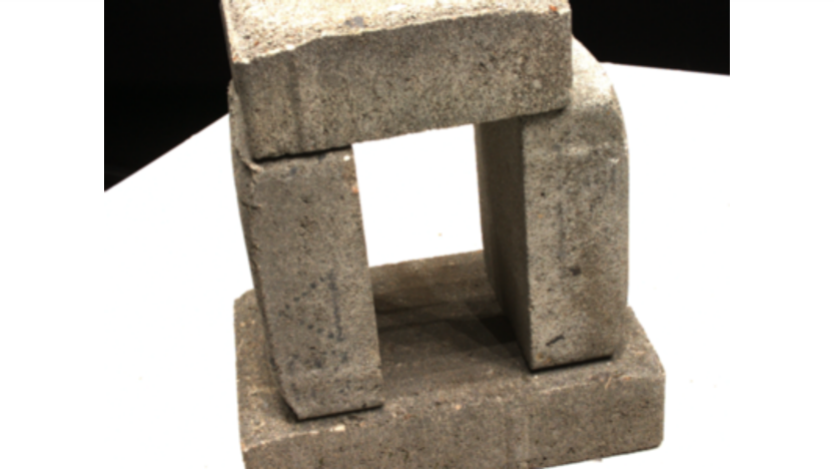}\hfill
\cellimg{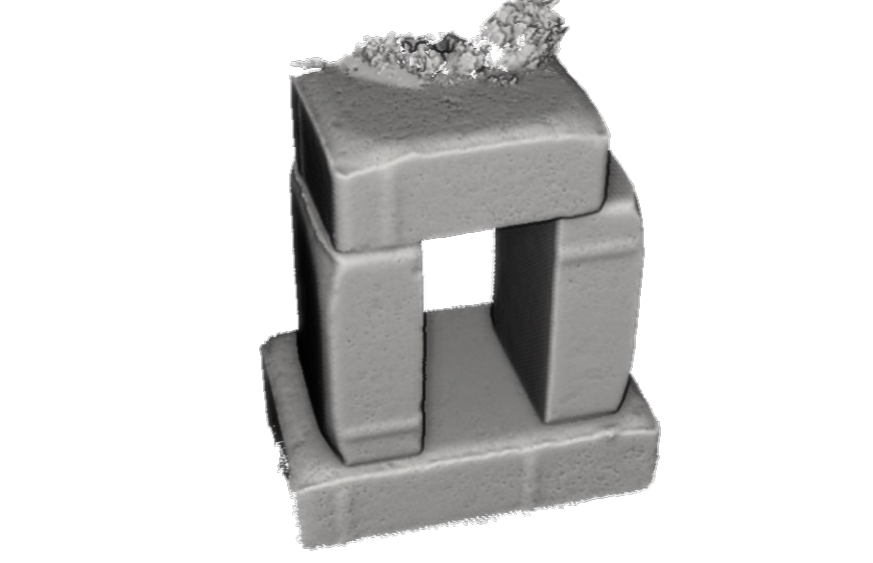}\hfill
\cellimg{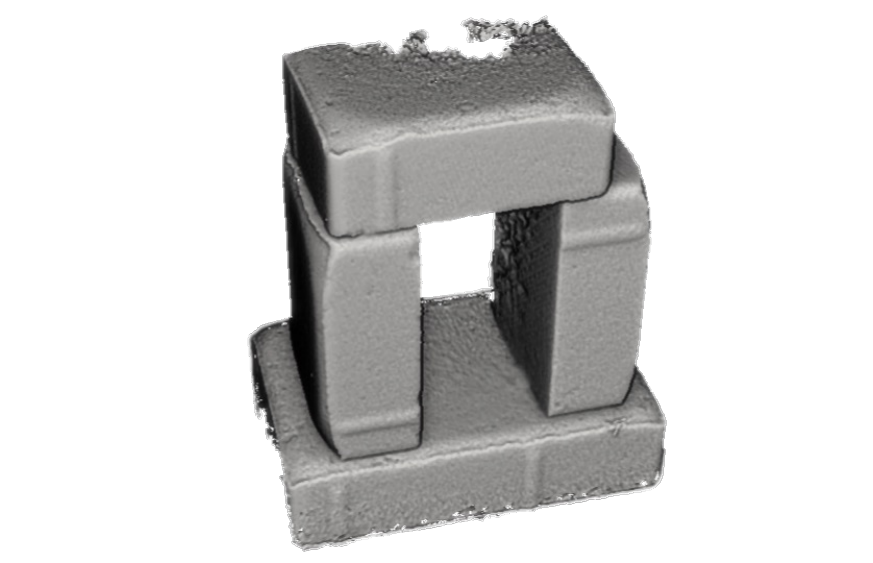}\hfill
\cellimg{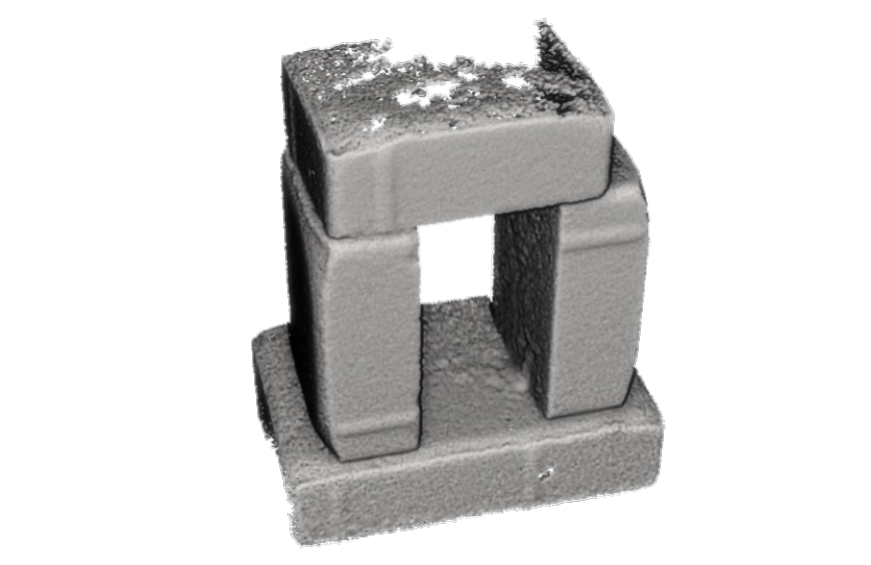}\hfill
\cellimg{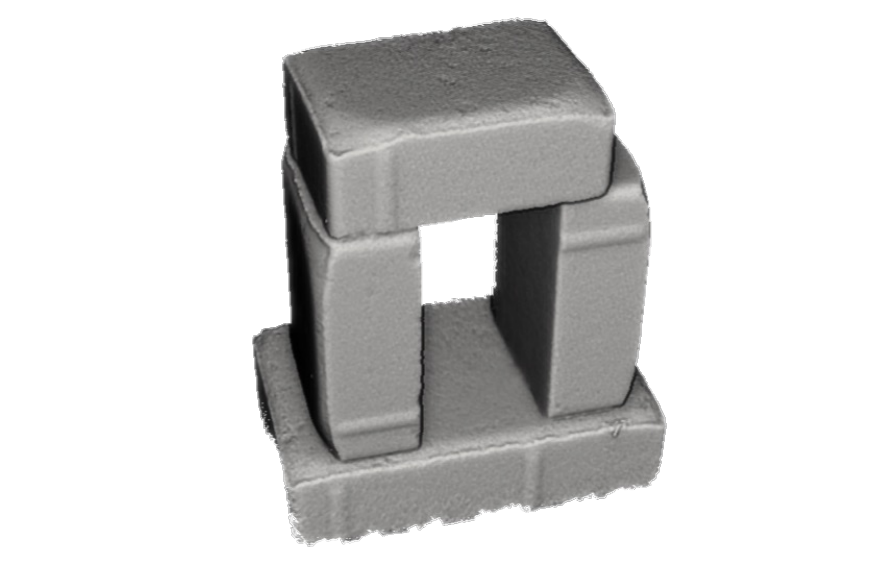}

\vspace{0.25em}
\cellimg{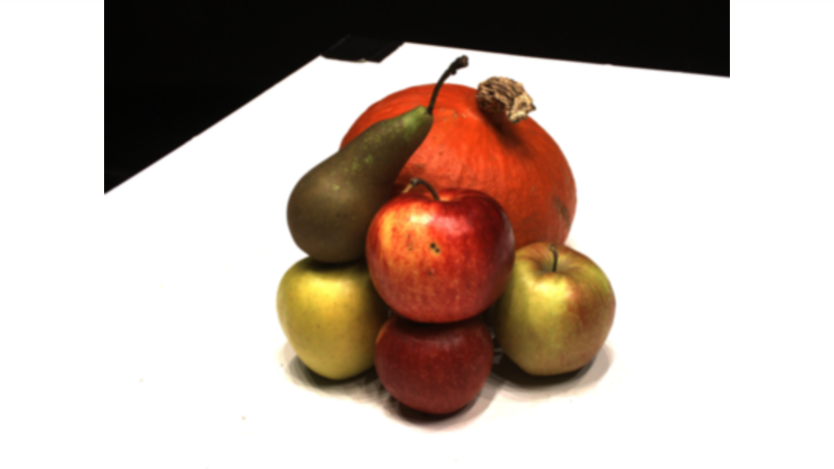}\hfill
\cellimg{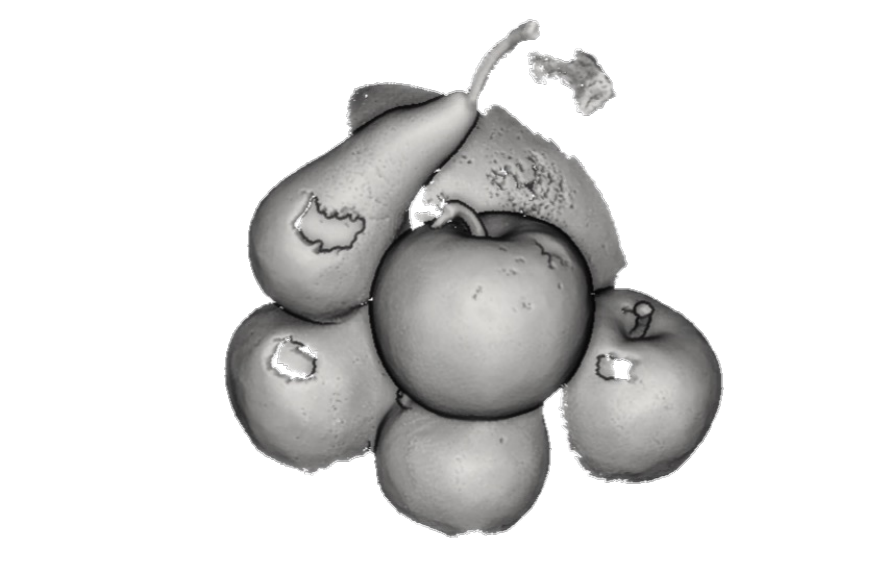}\hfill
\cellimg{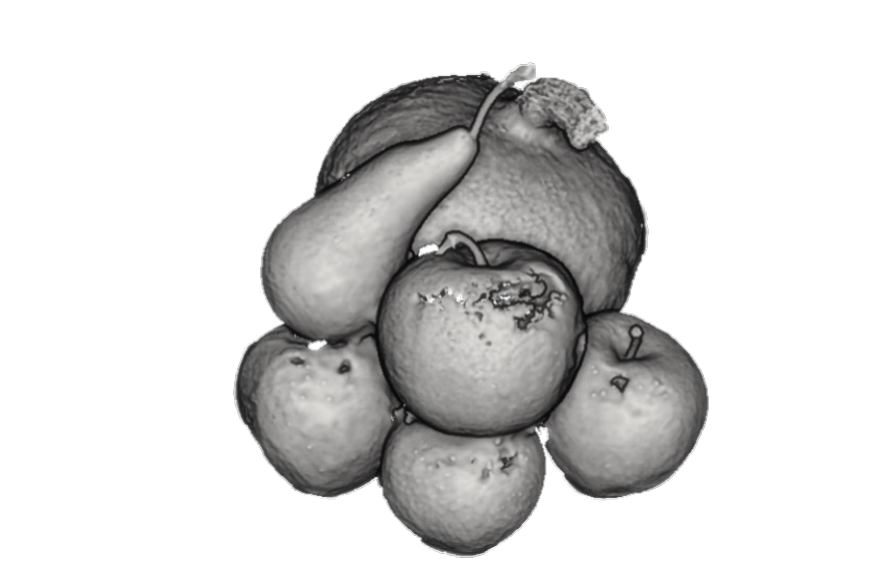}\hfill
\cellimg{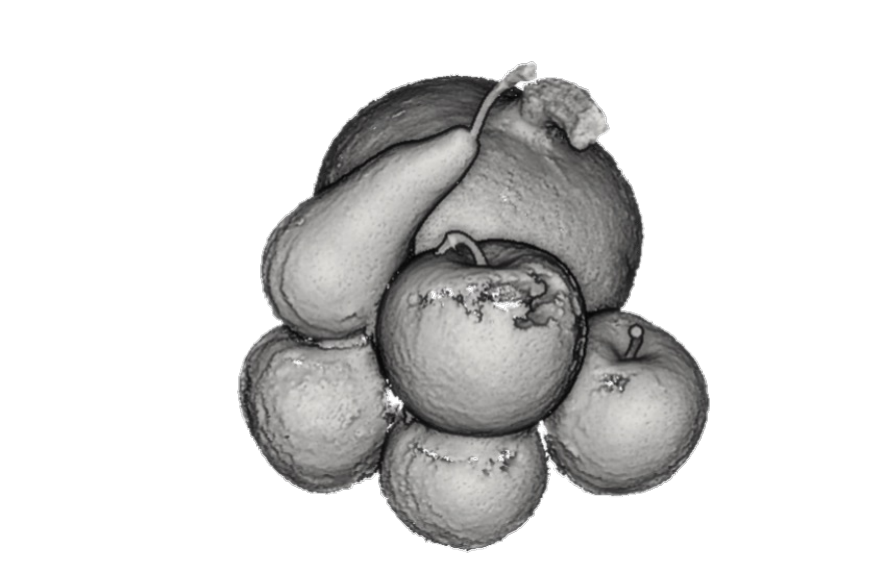}\hfill
\cellimg{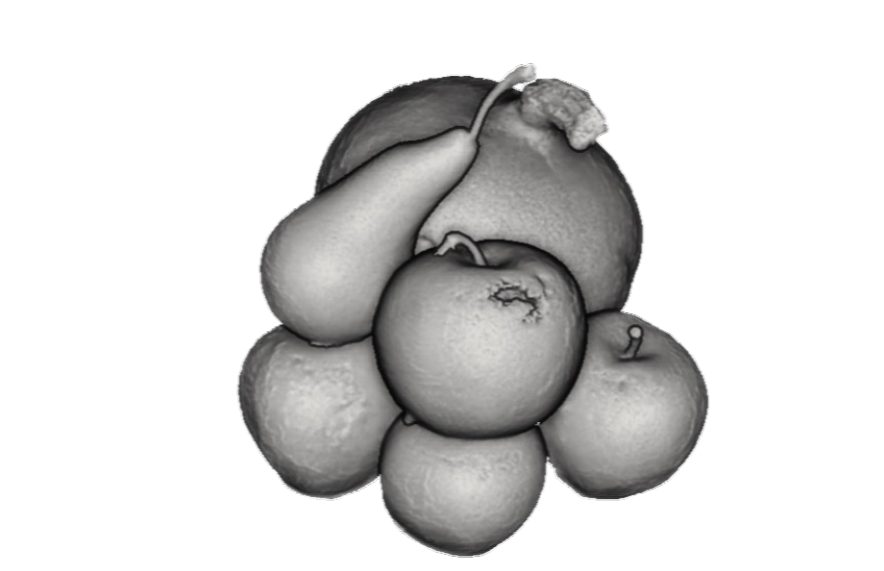}

\vspace{0.25em}

\cellimg{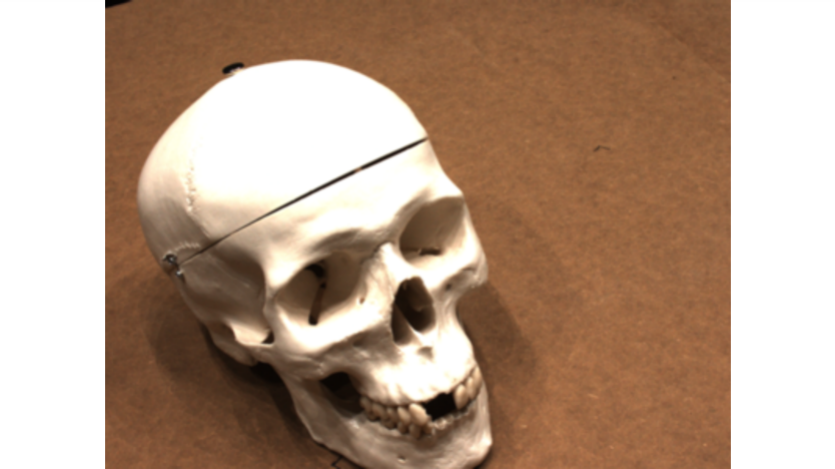}\hfill
\cellimg{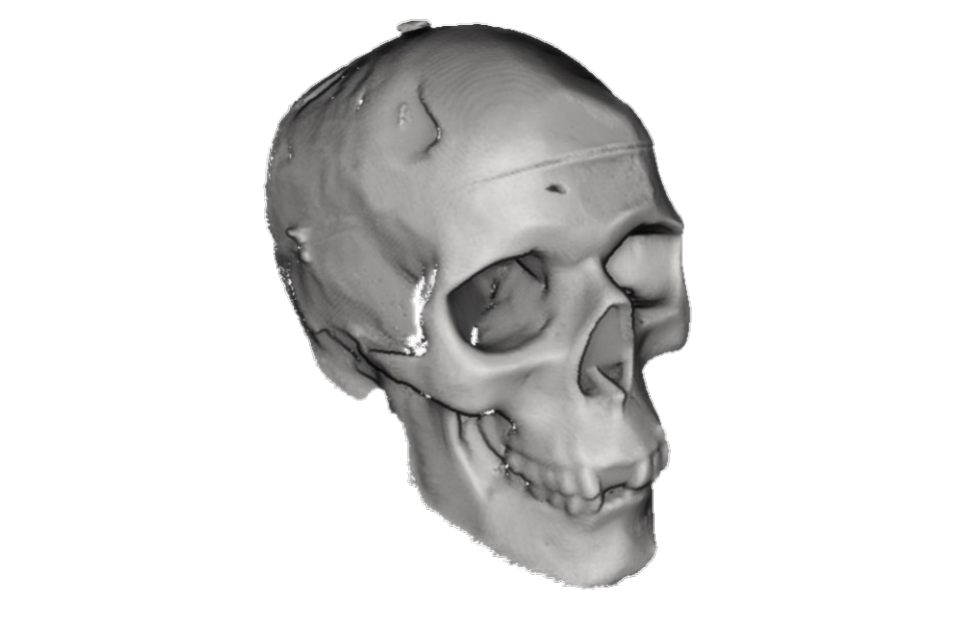}\hfill
\cellimg{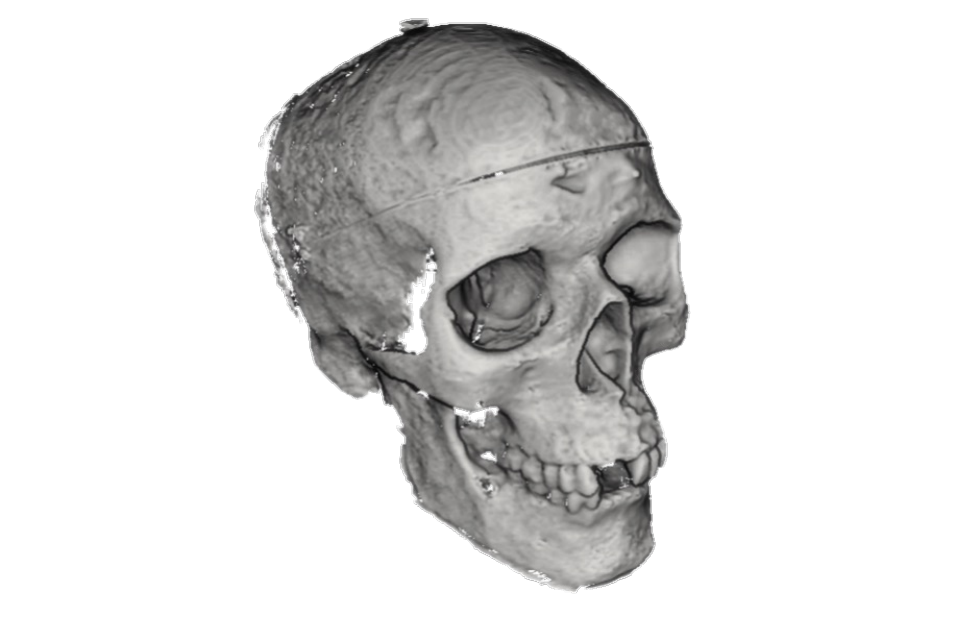}\hfill
\cellimg{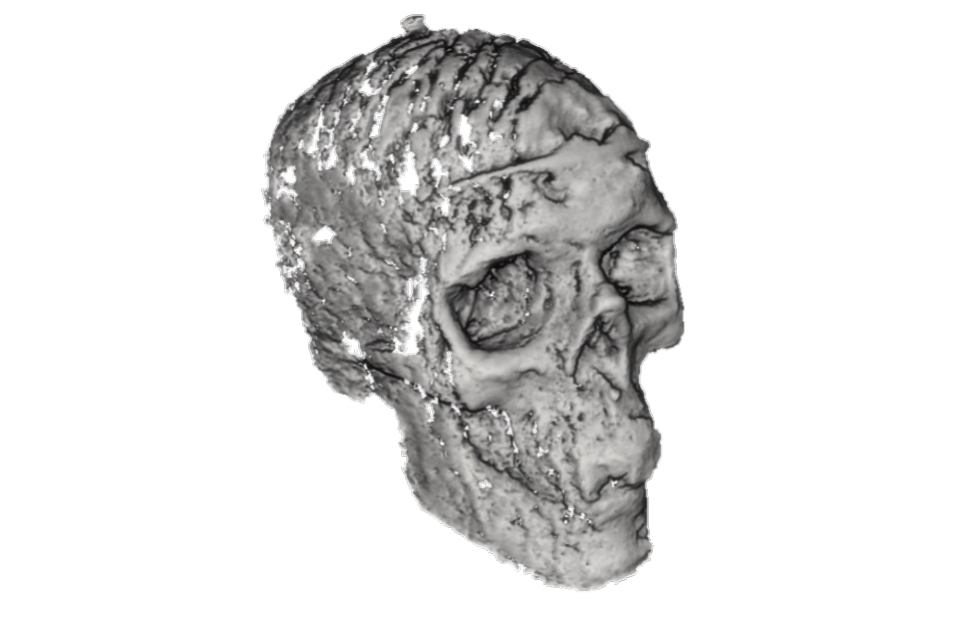}\hfill
\cellimg{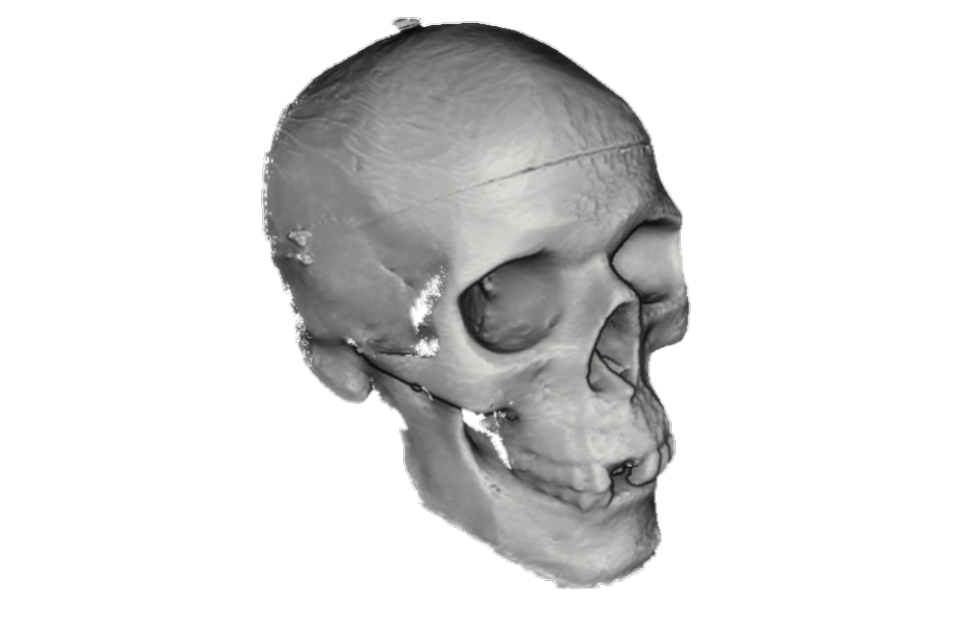}

\vspace{0.25em}
\cellimg{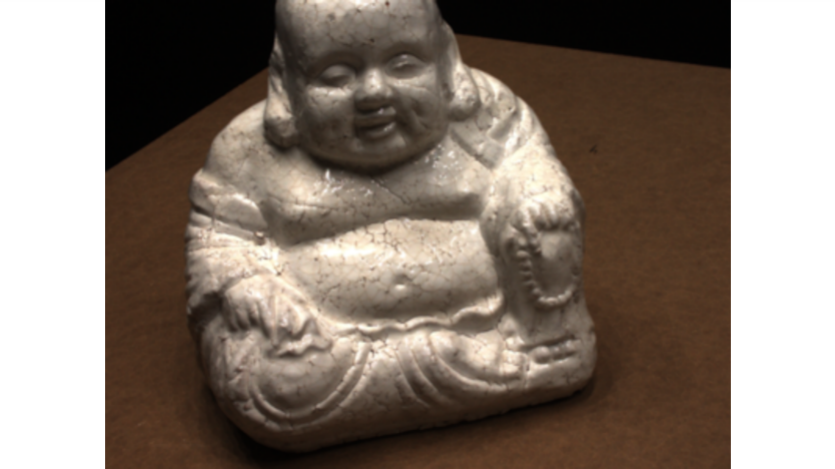}\hfill
\cellimg{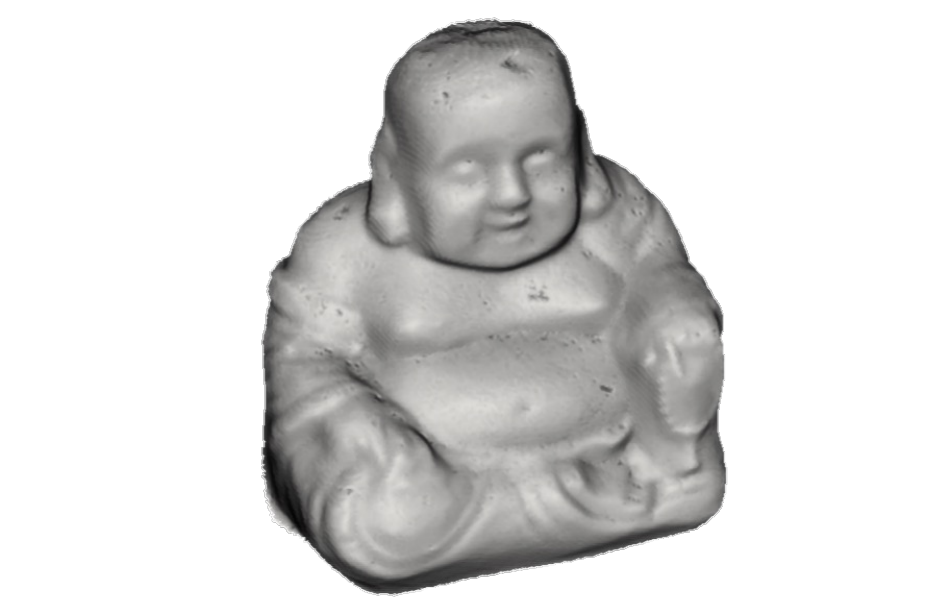}\hfill
\cellimg{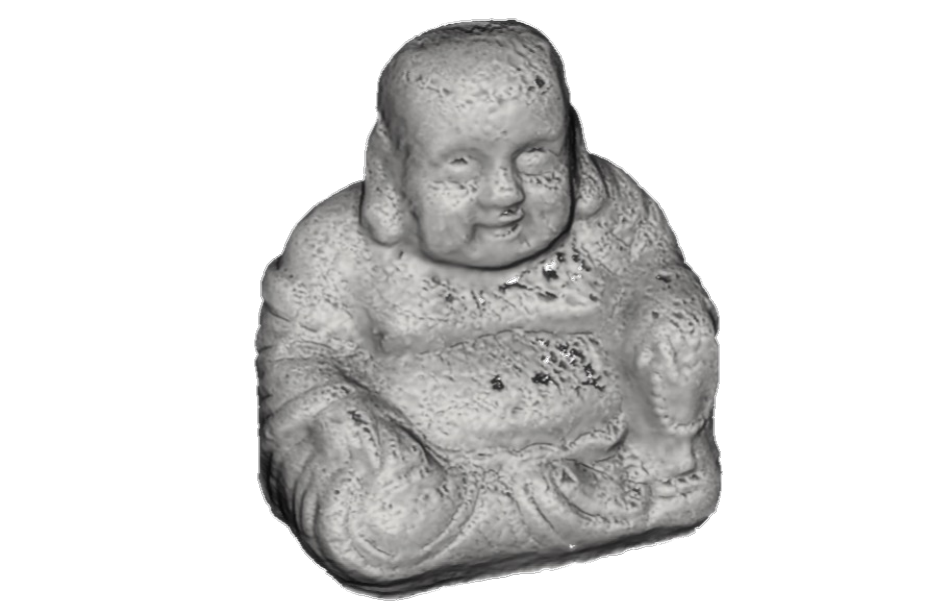}\hfill
\cellimg{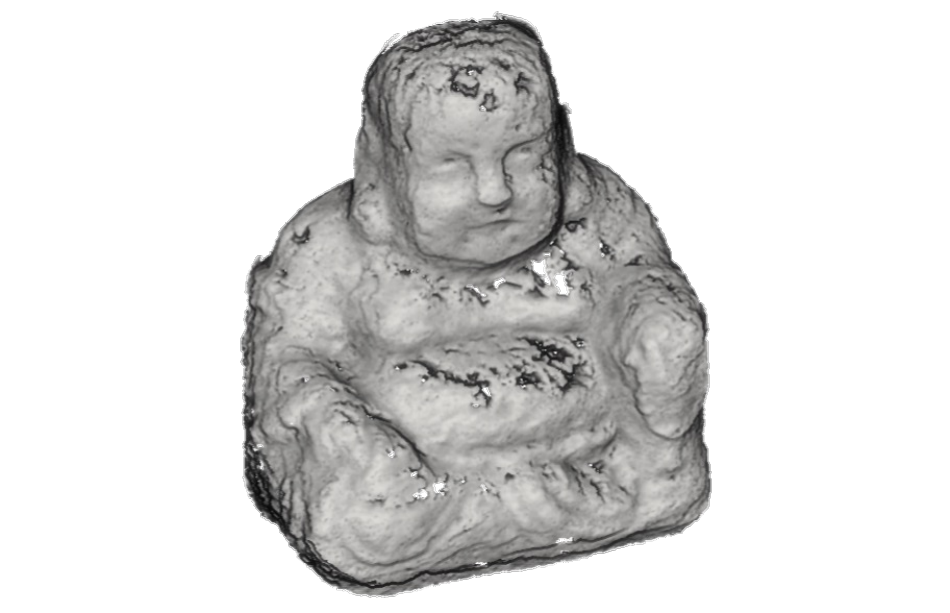}\hfill
\cellimg{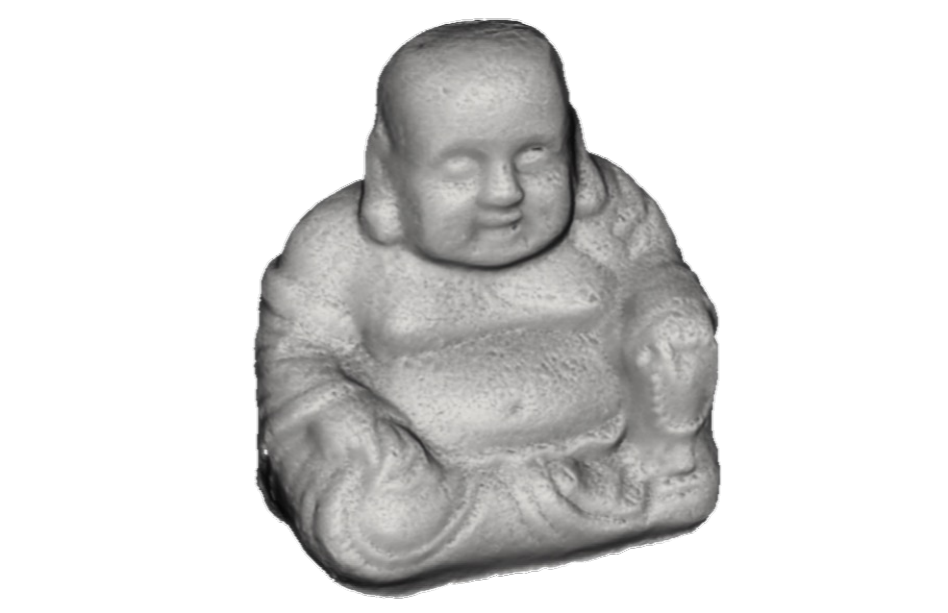}
\vspace{0.5em}
\colhead{Input}\hfill
\colhead{2DGS}\hfill
\colhead{SVRaster}\hfill
\colhead{SVRaster + SDF}\hfill
\colhead{Ours}

\vspace{-2mm}
\caption{\textbf{Reconstructed mesh comparison on DTU.}
Four scenes (rows) and four methods (columns). From top to bottom, we show scans 40, 63, 65, and 114. Our method converges to cleaner and more complete geometry while preserving details (best viewed with zoom).}
\label{fig:DTU-qual}
\vspace{-2mm}
\end{figure*}

%% file: sec/Results.tex
\section{Experiments}
\label{sec:experiments}

\input{table/05_table_tnt}

\input{figs/8_TNT_comparision/8_TNT_comparison}

\subsection{Implementation}
Our implementation uses PyTorch with custom CUDA kernels on top of SVRaster~\cite{Sun_2025_CVPR_SVRaster}.
We train for $8\mathrm{K}$ iterations on the DTU dataset~\cite{jensen2014large} and $10\mathrm{K}$ on the Tanks-and-Temples (TnT) dataset~\cite{knapitsch2017tanks}.
Because TnT covers larger scenes, we add one additional level (DTU: $2^{10}$; TnT: $2^{11}$) and allocate $2\mathrm{K}$ extra iterations to stabilize optimization at the finest level.
All experiments are done by a NVIDIA RTX~4090.


\subsection{Evaluation}

\noindent \textbf{Datasets.}
We evaluate on the DTU benchmark using the NeuS-preprocessed split~\cite{Wang_2021_NeurIPS_NeuS}
and on the Tanks-and-Temples dataset (TnT) using the 2DGS-preprocessed release~\cite{Huang_2024_SIGGRAPH_2DGS}.
All images are downsampled by $\times2$.
We compare against 2DGS~\cite{Huang_2024_SIGGRAPH_2DGS},
SVRaster~\cite{Sun_2025_CVPR_SVRaster}, and our full model (\emph{Ours}). Additionally, we measure the performance of a naive combination of two baselines `\emph{SVRaster+SDF}' since, as far as our understanding, there is no official paper that deals with Signed Distance Function on top of the Sparse voxel rasterization.
This naive variant follows the same training strategy as SVRaster, but stores SDF values at the voxel corners and renders using~\cref{eq:neus_alpha}. It also applies a per-voxel, ray-directional Eikonal loss, similar to NeuS~\cite{Wang_2021_NeurIPS_NeuS}.

\noindent\textbf{Analysis.}
For mesh extraction, we employ TSDF fusion for the DTU dataset and direct marching cubes for the TnT dataset, which we found to produce reliable meshes in practice. To incorporate the background in the TnT dataset, we define an \textit{outer\_voxel\_area} and initialize it as a large spherical shell to represent the sky. Qualitatively, as shown in ~\cref{fig:DTU-qual} and ~\cref{fig:tnt_qual}, our reconstructions are typically smoother and less noisy than those of the baselines, owing to the proposed smoothness terms and our coarse-to-fine learning schedule. Furthermore, our method consistently produces hole-free meshes. This contrasts sharply with a naive SVRaster+SDF implementation, which often falls into severe geometric local minima and exhibits significant artifacts on challenging scenes such as DTU scan 65 and the TnT Barn. These failures demonstrate that geometric initialization is critically important when applying SDFs to sparse voxels, especially as the scene’s underlying structure deviates from the simple spherical initialization. Quantitatively, as shown in ~\cref{tab:dtu} and ~\cref{tab:TnT}, our method provides a favorable accuracy-efficiency trade-off and clearly improves over vanilla SVRaster, although it does not always achieve the lowest Chamfer Distance (CD) compared to the strongest baselines using Gaussian representation~\cite{chen2024pgsr,li2024monogsdf}.

\subsection{Ablation Study}
We ablate each component to verify that the proposed design is both \emph{effective} and \emph{necessary} when combining an SDF with SVRaster.
We report one representative scene per dataset; quantitative results are summarized in~\cref{tab:ablations}.

\noindent\textbf{Variants 1-3: Before continuity (ray-Eikonal only).} We first train for 20k iterations using the SVRaster pruning strategy, without any continuity term. V1, a naive SDF swap with only the Eikonal loss, struggled to converge as we initialized the SDF as a central sphere. Consequently, complex scenes like TnT failed to converge properly, yielding a very low F1 score and requiring nearly an hour of training. Adding the PI$^3$ initialization (V2) reduces training time, but brings little accuracy gain in DTU and is more effective in TnT, indicating initialization is a critical step in outdoor complex scene reconstruction.
With the normal cue (V3) active for 18k iterations, accuracy improves noticeably, especially for DTU, showing that normals provide useful geometric guidance despite noise. However, it does not provide significant changes for TnT.

\noindent\textbf{Variants 4-5: Introducing continuity.}
We introduce continuity under a coarse-to-fine schedule with $\log s$ density scheduling.
V4 trains only up to the $2^{9}$ resolution level, applying child-level continuity and stopping at 6000 iterations. This substantially stabilizes surfaces, and we found this intermediate stage critical for large scenes like TnT.
Finally, our full model (V5) extends this process. It trains to final resolutions ($2^{10}$ for DTU, $2^{11}$ for TnT) and total iterations (8k for DTU, 10k for TnT). Crucially, after the initial 6000 iterations (which use child-only continuity), we switch to parent-child continuity for the remaining refinement. This preserves smooth transitions across fine voxels without the cost blow-up of full child propagation and yields the best overall quality.

\input{table/05_ablation}

\noindent\textbf{Effect of normal cue from point maps.}
We observe that enforcing the normal loss until the end of training can hinder convergence. 
As shown in~\cref{fig:normal-and-table}, the surface normal maps tend to have repetitive error, which is also addressed in the official repository. So in our framework, we use this noisy surface normal during the early phase of training, which still provides a meaningful signal for stable optimization of individual voxels as stated in V2 and V3 of ~\cref{tab:ablations}
\input{table/05_ablation2}
\input{table/05_ablation_parent}

\noindent\textbf{Effectiveness of parent-level continuity.}
To demonstrate the necessity of our parent-level continuity at high resolutions ($\ge 2^{10}$), we compare our full model (Ours) against two baselines: (1) one without any continuity term, and (2) one that uses only the ray-Eikonal loss at these fine levels. As summarized in~\cref{tab:ablation_continuity}, the 'w/o Continuity' baseline (1) expectedly yields the lowest F1 score (0.51). While adding the 'ray-Eikonal' loss (2) provides a slight improvement (F1 0.53), our full model (3) with parent-level continuity achieves the best results across Precision, Recall, and F1 score (0.57). This confirms our approach is an effective and efficient solution for high-resolution refinement.

\noindent\textbf{Convergence speed.}
With PI$^3$ initialization and our coarse-to-fine surface-aware training, our method converges faster per iteration than SVRaster.
This is also reflected in the iteration vs. CD curves in Fig.~\ref{fig:cd-conv}.

\input{figs/9_cd_per_iteration/9_CD_per_iteration}

%% file: table/05_table_tnt.tex
\begin{table*}[t]
\caption{
\textbf{Quantitative results on the Tanks-and-Temples Dataset}~\cite{knapitsch2017tanks}. 
We employ the F1 score for evaluation, the higher the better.
}
\label{tab:TnT}
\vspace{-2mm}
\setlength\tabcolsep{6.5pt}
\centering
\resizebox{0.8\linewidth}{!}{%
\setlength{\tabcolsep}{10.0pt}
\begin{tabular}{l|cccccc|c|c}
\toprule
Method & Barn & Caterpillar & Courthouse & Ignatius & Meetingroom & Truck & \textit{Mean} & \textit{Time} \\
\midrule

\multicolumn{9}{l}{\textbf{Implicit Representation.}} \\
NeuS \cite{Wang_2021_NeurIPS_NeuS} 
& 0.29 & 0.29 & 0.17 & 0.83 & 0.24 & 0.45 & 0.38 & $>24$h \\
Neuralangelo \cite{Li_2023_CVPR_Neuralangelo} 
& \textbf{0.70} & \textbf{0.36} & \textbf{0.28} & \textbf{0.89} & \textbf{0.32} & \textbf{0.48} & \textbf{0.50} & $>128$h \\
GeoNeuS \cite{geoneus} 
& 0.33 & 0.26 & 0.12 & 0.72 & 0.20 & 0.45 & 0.35 & $>12$h \\
MonoSDF \cite{monosdf} 
& 0.49 & 0.31 & 0.12 & 0.78 & 0.23 & 0.42 & 0.39 & 6h \\

\midrule
\multicolumn{9}{l}{\textbf{Explicit Representation (Gaussian).}} \\
2DGS \cite{Huang_2024_SIGGRAPH_2DGS} 
& 0.41 & 0.23 & 0.16 & 0.51 & 0.17 & 0.45 & 0.30 & 16m \\
GOF \cite{gof} 
& 0.51 & 0.41 & 0.28 & 0.68 & 0.28 & 0.59 & 0.46 & 24m \\
VCR-GauS \cite{chen2024vcr} 
& 0.62 & 0.26 & 0.19 & 0.61 & 0.19 & 0.52 & 0.40 & 53m \\
MonoGSDF \cite{li2024monogsdf} 
& 0.56 & 0.38 & \textbf{0.29} & 0.72 & 0.25 & 0.62 & 0.47 & 3h \\
PGSR \cite{chen2024pgsr} 
& \textbf{0.66} & \textbf{0.44} & 0.20 & \textbf{0.81} & \textbf{0.33} & \textbf{0.66} & \textbf{0.52} & 45m \\

\midrule
\multicolumn{9}{l}{\textbf{Explicit Representation (Sparse Voxel).}} \\
SVRaster \cite{Sun_2025_CVPR_SVRaster} 
& 0.35 & 0.33 & 0.29 & 0.69 & 0.19 & 0.54 & 0.40 & 10m \\
\nickname (ours) 
& \textbf{0.43} & \textbf{0.36} & \textbf{0.32} & \textbf{0.70} & \textbf{0.21} & \textbf{0.57} & \textbf{0.43} & 12m \\

\bottomrule
\end{tabular}%
}
\vspace{-2mm}
\end{table*}

%% file: figs/8_TNT_comparision/8_TNT_comparison.tex






\captionsetup[subfigure]{labelformat=empty} 

\newcommand{\colheadd}[1]{\begin{subfigure}{.24\textwidth}\centering\normalsize\textnormal{#1}\end{subfigure}}

\newcommand{\cellimgg}[1]{%
  \begin{subfigure}[b]{.24\textwidth}\centering
    \includegraphics[width=\linewidth]{#1}
  \end{subfigure}%
}

\begin{figure*}[t]
\centering
\cellimgg{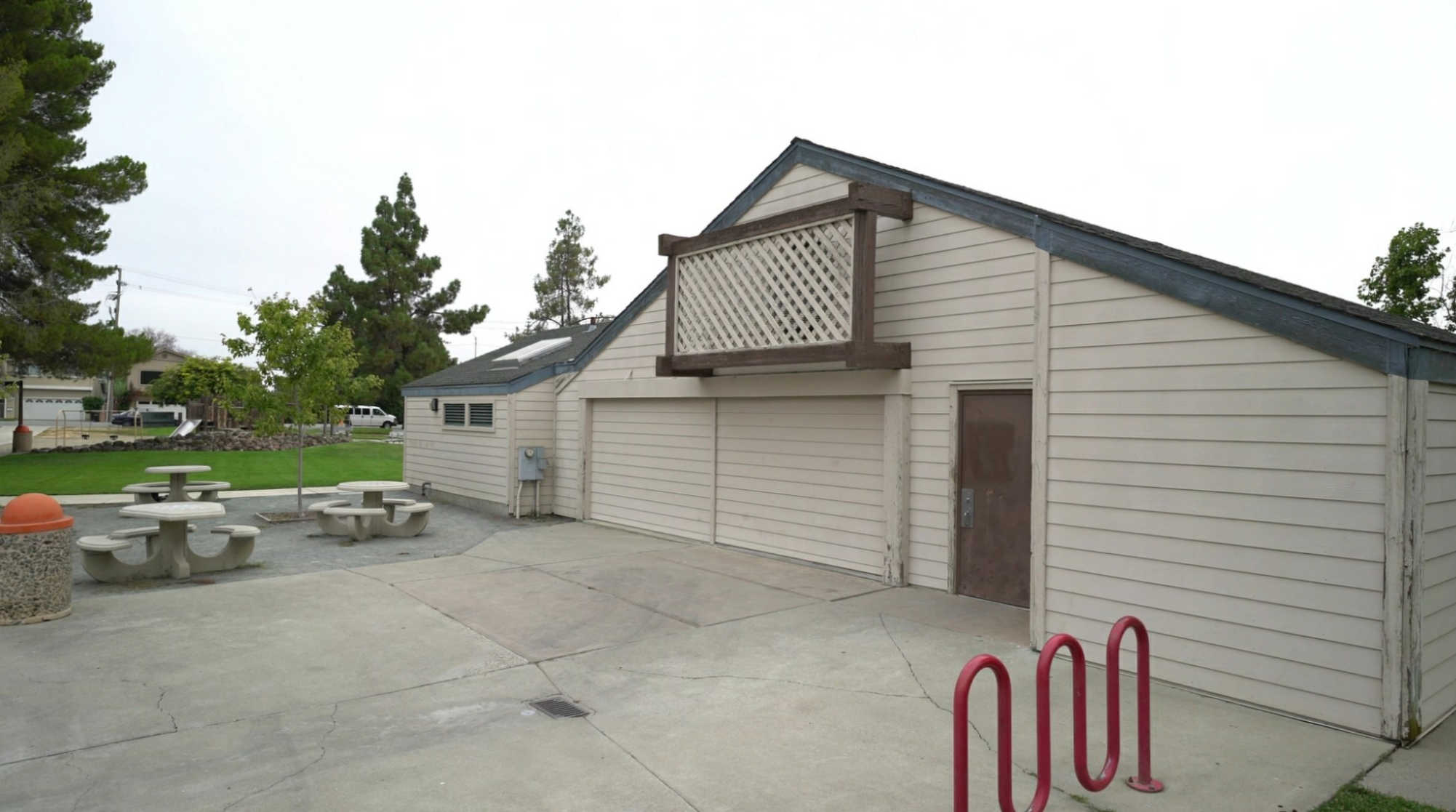}\hfill
\cellimgg{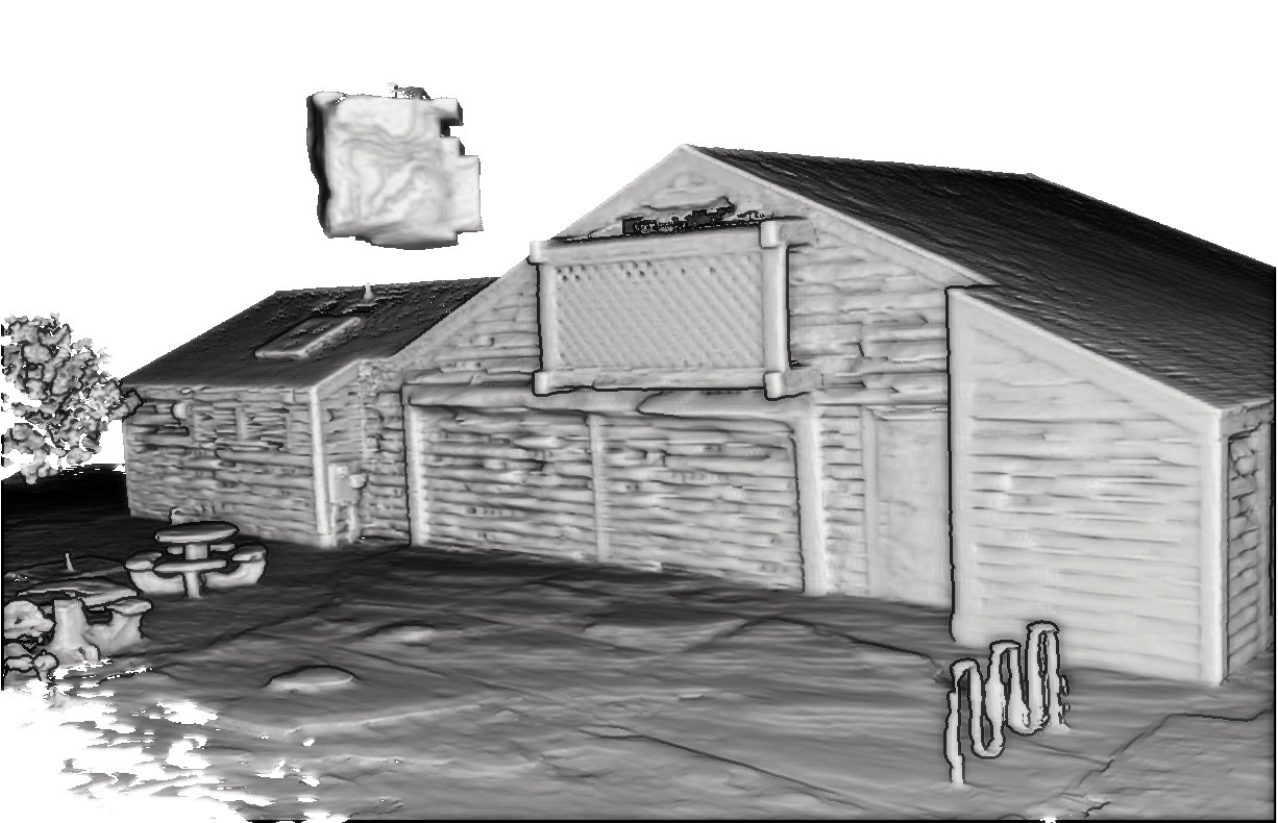}\hfill
\cellimgg{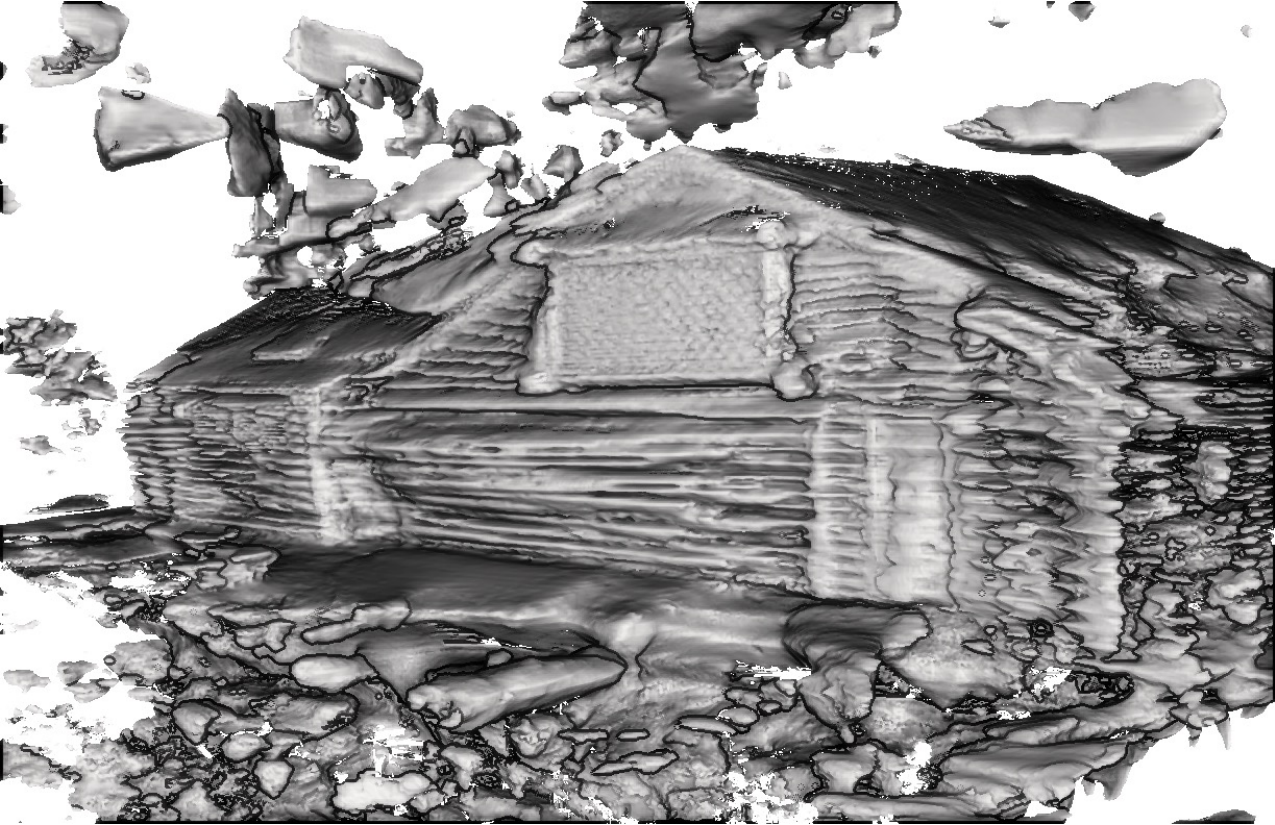}\hfill
\cellimgg{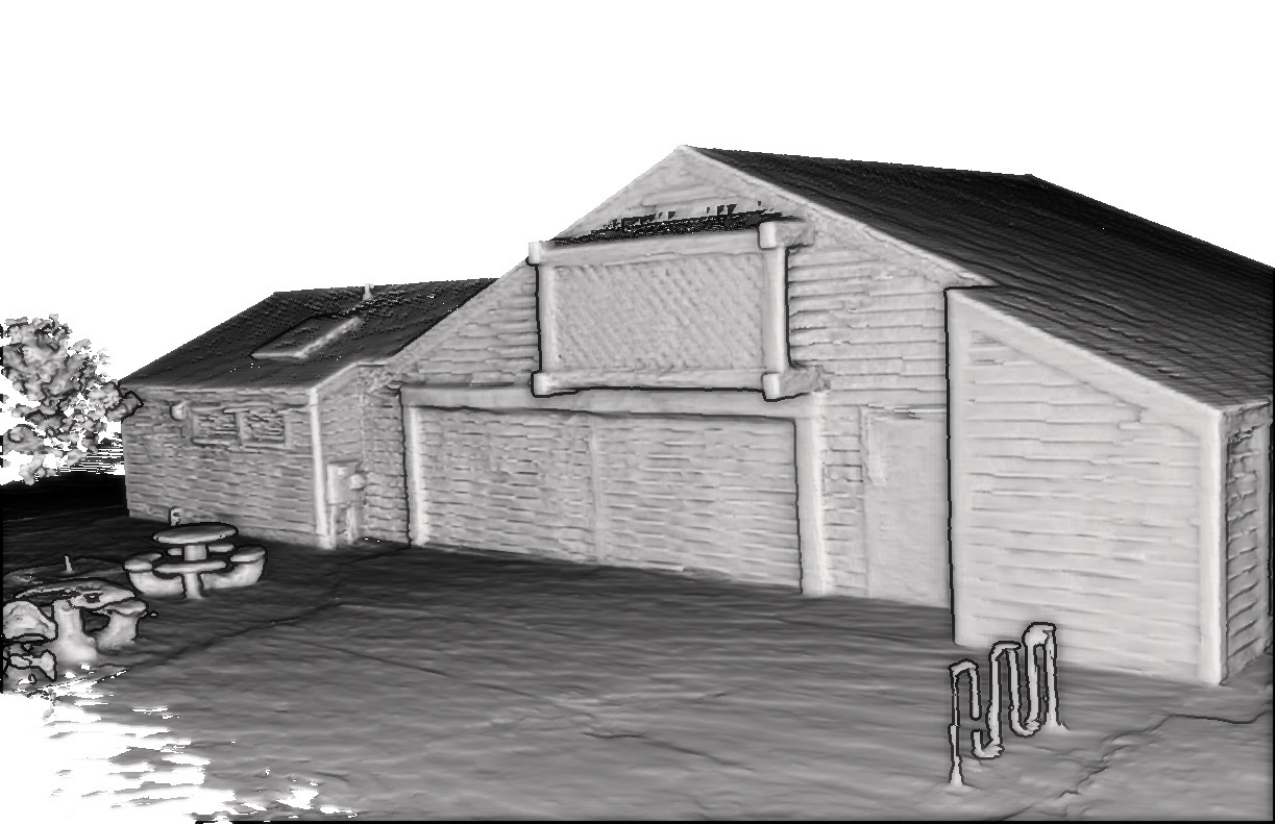}\\
\vspace{0.2em}
\cellimgg{figs/8_TNT_comparision/truck_image.pdf}\hfill
\cellimgg{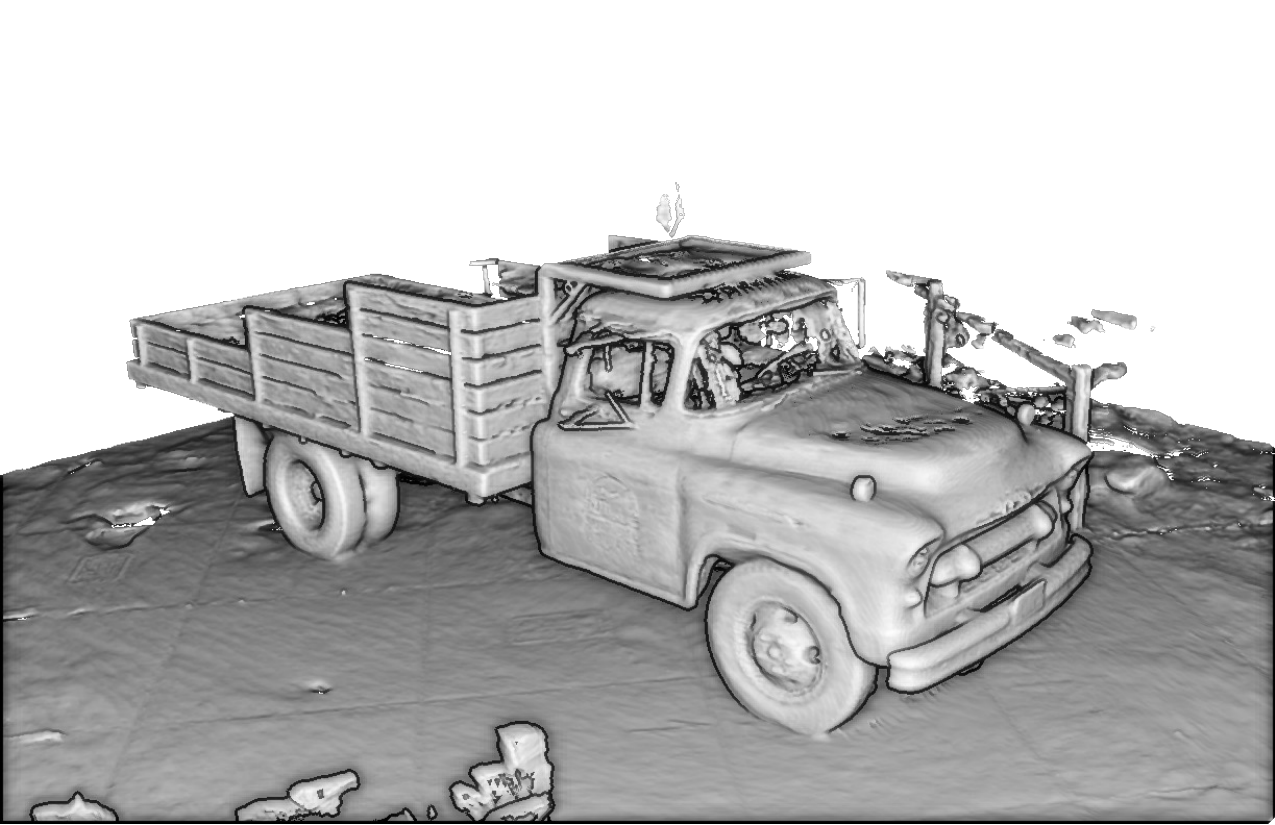}\hfill
\cellimgg{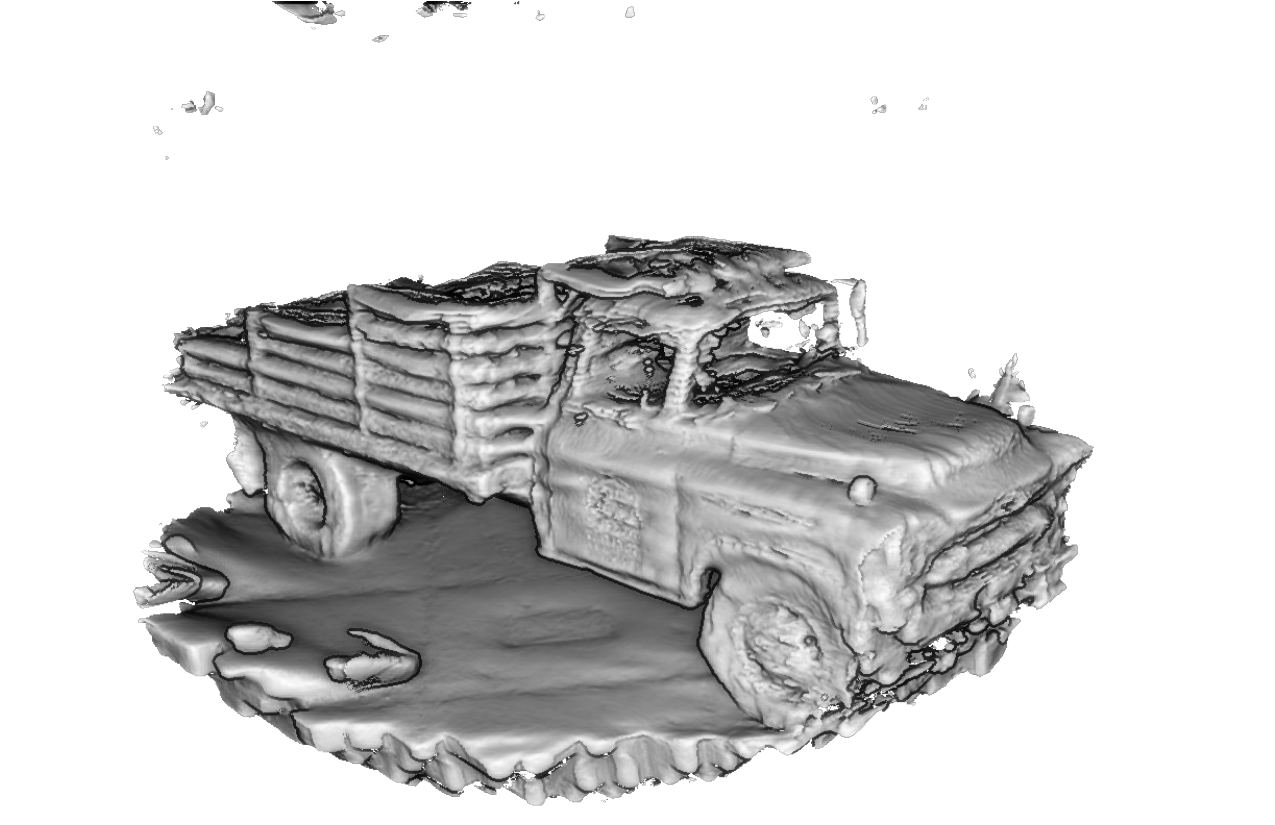}\hfill
\cellimgg{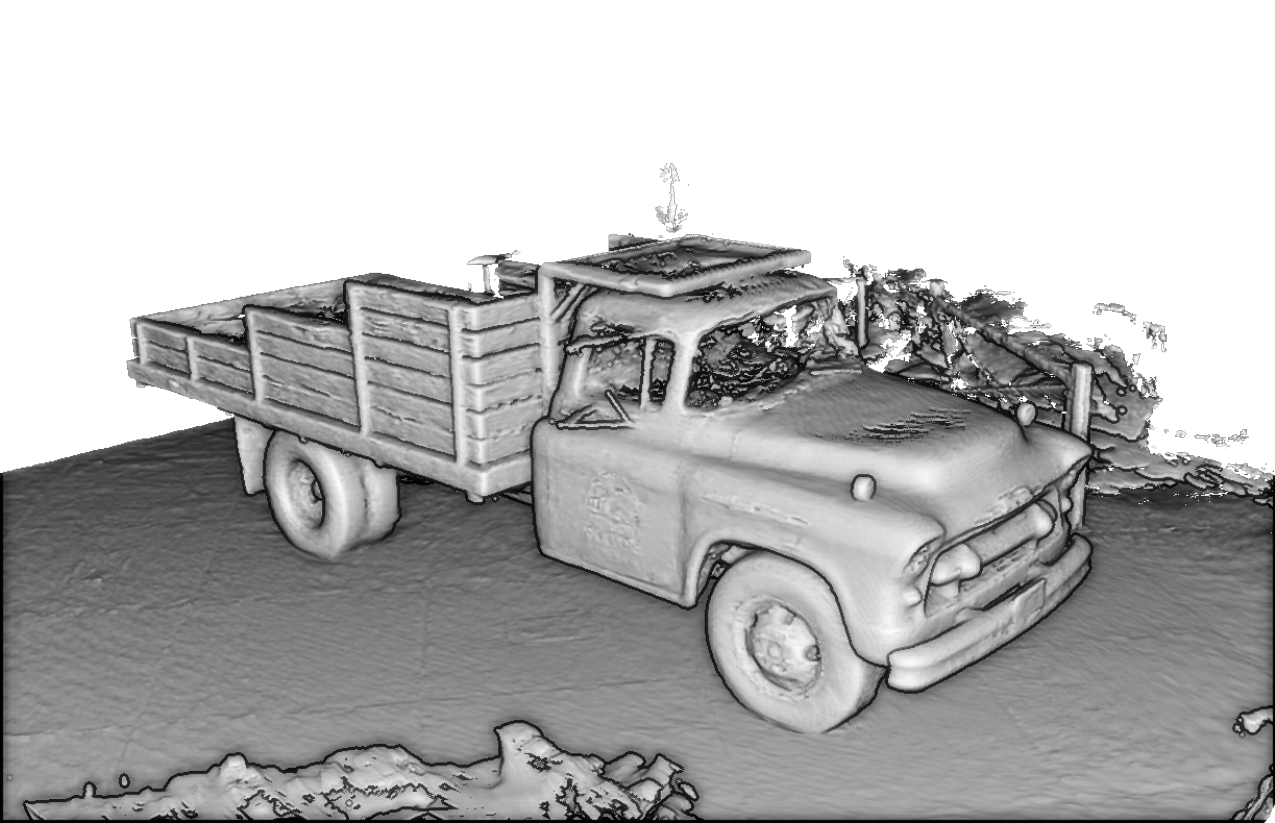}

\vspace{0.2em}

\colheadd{Input}\hfill
\colheadd{SVRaster}\hfill
\colheadd{SVRaster + SDF}\hfill
\colheadd{Ours}


\caption{\textbf{Qualitative comparison on Tanks-and-Temples.} Two scenes (rows) and three methods (columns). From top to bottom, we show scene Barn and Truck. Our method reconstructs clean geometry even in large-scale outdoor scenes. (best viewed with zoom).}
\label{fig:tnt_qual}
\vspace{-3mm}
\end{figure*}

%% file: table/05_ablation.tex
\begin{table}[t]
\centering
\setlength{\tabcolsep}{6pt}
\renewcommand{\arraystretch}{1.15}
\caption{Ablations of components. Lower CD is better; higher F1 is better. CD is for DTU scan 24, F1 score for TnT Barn}
\vspace{-2mm}
\resizebox{\columnwidth}{!}{
\setlength{\tabcolsep}{2.0pt}
\begin{tabular}{@{}c c cc cc cc cc@{}}
\toprule
\multirow{2}{*}{Step} & \multirow{2}{*}{ \shortstack{SDF \\ (NeuS)} } &
\multicolumn{2}{c}{PI$^3$ prior} &
\multicolumn{2}{c}{Continuity loss} &
\multicolumn{2}{c}{DTU} &
\multicolumn{2}{c}{Point recon} \\
\cmidrule(lr){3-4}\cmidrule(lr){5-6}\cmidrule(lr){7-8}\cmidrule(l){9-10}
& & Init. & Norm & Child($2^9$) & Parent & Time & Voxels & CD$\downarrow$ & F1$\uparrow$ \\
\midrule
V1 & \checkmark & & & & & 6.8m & 0.8M & 1.00  & 0.03 \\
V2 & \checkmark & \checkmark &  &  &  & 4.5m & 0.6M & 1.13 & 0.20 \\
V3 & \checkmark & \checkmark & \checkmark &  &  &  8.2m&  1.1M& 0.66 & 0.23 \\
\midrule
V4 & \checkmark & \checkmark & \checkmark & \checkmark & & 3.8m & 1.7M & 0.43 &  0.19\\
V5 & \checkmark & \checkmark & \checkmark & \checkmark & \checkmark & 5.6m & 4.3M & \textbf{0.40} & \textbf{0.43} \\
\bottomrule
\end{tabular}
}
\label{tab:ablations}
\vspace{-2mm}
\end{table}

%% file: table/05_ablation2.tex
\begin{figure}[t]
\centering
\begin{minipage}[t]{0.42\columnwidth}
  \vspace{0pt} 
  \centering
  \includegraphics[width=\linewidth]{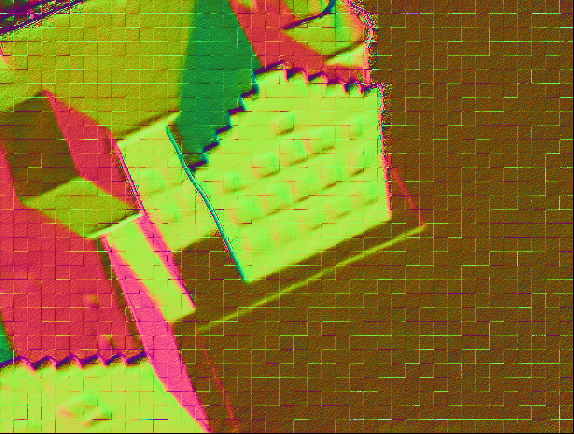}\\[-0.2em]
  {\scriptsize \textbf{Normal}}
\end{minipage}
\hfill
\begin{minipage}[t]{0.54\columnwidth}
  \vspace{0pt} 
  \centering
  {\scriptsize \textbf{CD (↓) comparison}}\\[0.2em]
  {\scriptsize
  \setlength{\tabcolsep}{4pt}%
  \resizebox{\linewidth}{!}{%
    \begin{tabular}{@{}lcc@{}}
      \toprule
      Scene & Normal (till last) & Ours \\
      \midrule
      DTU24  & 0.47 & \textbf{0.40} \\
      DTU37   & 0.81 & \textbf{0.67} \\
      DTU69   & 0.75 & \textbf{0.68} \\
      DTU83   & 1.37 & \textbf{1.32} \\
      DTU105  & 0.84 & \textbf{0.69} \\
      \bottomrule
    \end{tabular}}}
\end{minipage}
\vspace{-2mm}
\caption{\textbf{Normals vs. accuracy.} Left: A visualization of the noisy normal prior derived from the point map. 
Right: Chamfer Distance (CD$\downarrow$) comparison. Persisting the normal loss until the final iteration (using a noisy prior) degrades performance, demonstrating the importance of our annealing strategy.}
\label{fig:normal-and-table}
\vspace{-2mm}
\end{figure}

%% file: table/05_ablation_parent.tex
\begin{table}[t]
\centering
\caption{Effectiveness of parent-level continuity on the TNT Truck scene at high resolutions ($\ge 2^{10}$). Our parent-level term (Ours) is crucial for high-fidelity refinement compared to baselines.}
\label{tab:ablation_continuity}
\vspace{-2mm}
\resizebox{\columnwidth}{!}{
\begin{tabular}{lccc}
\toprule
Metric & (1) w/o Continuity & (2) w/ ray-Eikonal & (3) Ours \\
\midrule
Precision ($\uparrow$) & 0.46 & 0.46 & \textbf{0.52} \\
Recall ($\uparrow$) & 0.57 & 0.63 & \textbf{0.63} \\ 
F1 Score ($\uparrow$) & 0.51 & 0.53 & \textbf{0.57} \\
\bottomrule
\end{tabular}
}
\vspace{-2mm}
\end{table}

%% file: figs/9_cd_per_iteration/9_CD_per_iteration.tex

\begin{figure}[t]
  \centering
  \includegraphics[width=0.9\columnwidth]{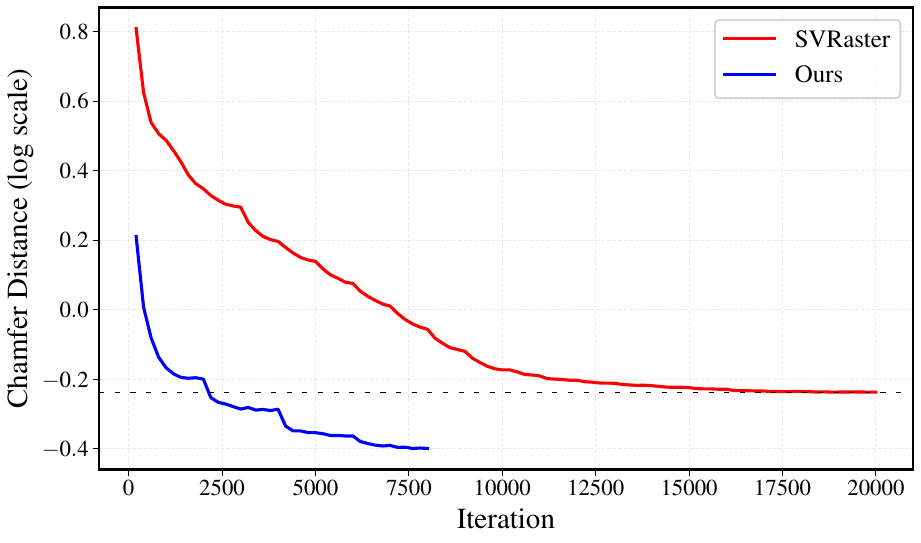}
  \vspace{-4mm}
  \caption{Convergence on DTU scan 24. Our method converges significantly faster and achieves a lower final Chamfer Distance (CD$\downarrow$) than baseline. Notably, our model surpasses the final reconstruction quality of SVRaster after only 2,200 training iterations.}
  \label{fig:cd-conv}
  \vspace{-2mm}
\end{figure}

%% file: sec/Conclusion.tex
\section{Conclusion}
We propose \nickname integrates Signed Distance Functions into sparse voxel rasterization to achieve fast, smooth, and accurate 3D surface reconstruction. By leveraging geometry-informed initialization and spatial coherence in LoD, our method overcomes the discontinuity and local minima issues that arise when applying SDFs to independently parameterized voxels. Despite these benefits, the approach remains limited by the resolution–memory trade-off inherent to sparse voxel structures, and its performance can still be affected by imperfect geometric priors, leaving room for improved initialization and more robust continuity mechanisms in future work.

%% file: supp/0_intro.tex
In this supplement, we describe the details of our method that are not included in the the main paper: We describe the voxel initialization (\cref{sec:voxel_initialization}), training losses (\cref{sec:loss_details}), implementation details (\cref{sec:implementation_details}), ablation study (\cref{sec:ablation_study_supp}), and qualitative results (\cref{sec:qualitative_results}).

%% file: supp/a_sparse_voxel_initialization.tex
\section{Voxel Initialization}
\label{sec:voxel_initialization}

Geometry initialization plays a crucial role in optimizing neural fields, especially when employing explicit representations such as 3D Gaussians~\cite{Kerbl_2023_ACM_3DGS} or sparse voxels~\cite{Sun_2025_CVPR_SVRaster}.
While density-based methods like SVRaster typically start from an empty or near-zero space, our SDF-based formulation enables faster and more efficient convergence by directly utilizing rich geometric priors from the beginning.
To achieve this stable and physically meaningful initialization, we leverage PI$^3$~\cite{wang2025pi3}, a visual geometry model that estimates 3D points in world coordinates from unposed images. 
Given a set of input images $\mathcal{I} = \set{I_i}_{i=1}^{N_{img}}$, where $N_{img}$ is the number of images, the model infers a set of camera poses $\set{[R_i| \mathbf{t}_i]}_{i=1}^{N_{img}}$ ($[R|\mathbf{t}] \in \mathbb{R}^{3\times 4}$) together with per-view 3D point maps $\set{P_i}_{i=1}^{N_{img}}$ (in estimated world coordinates) and confidence maps $\set{C_i}_{i=1}^{N_{img}}$. Note that the camera pose $[R_i|t_i]$ warps a point at the world coordinate into the camera coordinate corresponding to the $i$-th image $I_i$.

\paragraph{Camera Pose Alignment.}
Because the predicted point maps are expressed in the coordinate system defined by the model’s predicted camera parameters, they are not immediately consistent with the ground-truth world coordinate system assumed by prior methods. Given $\{[R_i\mid \mathbf{t}_i]\}_{i=1}^{N_{\mathrm{img}}}$ and
$\{[R_i^{\mathrm{gt}}\mid \mathbf{t}_i^{\mathrm{gt}}]\}_{i=1}^{N_{\mathrm{img}}}$,
we estimate a 7-DoF similarity $(s,R,\mathbf{t})$ by aligning camera centers by
$\mathbf{C}_i {=} {-}R_i^{\!\top}\mathbf{t}_i$ and $\mathbf{C}_i^{\mathrm{gt}} {=} {-}\big(R_i^{\mathrm{gt}}\big)^{\!\top}\mathbf{t}_i^{\mathrm{gt}}$
where $\mathbf{C}_i$ and $\mathbf{C}_i^{\mathrm{gt}}$ denote camera positions in world coordinates. Then, we optimize $\tilde{s},\tilde{R},\tilde{\mathbf{t}}$ by:
\begin{equation}
    \arg\min_{\tilde{s},\tilde{R},\tilde{\mathbf{t}}}
\sum_{i=1}^{N_{\mathrm{img}}}
\left\|\, s\,R\,\mathbf{C}_i + \mathbf{t} - \mathbf{C}_i^{\mathrm{gt}} \right\|_2^2.
\end{equation}

With the optimized $(s,R,\mathbf{t})$, we warp point maps into the ground truth world coordinate system as $\mathbf{p}_{\text{world}} = s\,R\,\mathbf{p} + \mathbf{t}$ where $\mathbf{p}$ is a point in point maps and $\mathbf{p}_{\text{world}}$ is a warped point. As a result, we obtain the warped point maps~$P_{\text{world}}$ at the ground truth world coordinate, which will be used for voxel allocation and SDF initialization in the following steps.

\paragraph{Voxel Allocation with Negative Distances.}
Our process begins by defining a uniform dense grid at $L=6$ ($G=2^6$) within the predefined cube region. The next step is to allocate individual SDF values inside all voxel corner points.
Before allocation, it is unknown which corner points are inside or outside the surface. We adopt a robust strategy by first assuming all voxel corners are inside the object, initializing them with a negative SDF value. The subsequent sign-flipping step will then 'carve out' the visible 'outside' (positive) space from this initial negative volume. This provides a more stable starting point than initializing with ambiguous unsigned distances.

To assign initial SDF magnitudes to all $({2^L+1})^3$ corner nodes $p_{geo}$ of this dense $L=6$ grid, we first build a KD-tree over the subsampled aligned points $P^{sub}_{\text{world}}$ for efficient nearest-neighbor queries.
For each corner node $p_{geo}$, we find the nearest point $\mathbf{p}_{p_{geo}} \in P^{sub}_{\text{world}}$ using the KD-tree and assign its negative distance:
\begin{equation}\label{eq:sdf_initialization}
\tilde{f}(p_{geo}) = {-}\Big(\min_{\mathbf{p}_{p_{geo}} \in P^{sub}_{world}}|p_{geo}-\mathbf{p}_{p_{geo}}|_2 \Big),
\end{equation}
where $\tilde{f}(p_{geo})$ function denotes the initial SDF value in $p_{geo}$. This value provides the initial 'inside' magnitude, and in the next step, we flip the sign for visible corners.



\paragraph{Voxels with Initial Signed Distance.}
Unsigned distance alone does not indicate whether a point is inside or outside the surface. To initialize a sign for each voxel, we determine if its corners are "visible" and flip their signs accordingly (e.g., to positive/outside). We check visibility using two conditions. First, we project all voxel corners and the point map $P_{\text{world}}$ onto low-resolution camera pixels. A corner's sign is flipped if its distance to the camera is less than the distance of the nearest $P_{\text{world}}$ point projecting to the same pixel (implying it is not occluded by the point map). Second, any voxel corner that is never projected to any camera (i.e., it is outside all camera frustums) is also flipped.

%% file: supp/b_loss.tex
\section{Loss Details}
\label{sec:loss_details}
In this section, we describe the loss terms that regularize our signed distance field.
We first explain how we sample query points in the dense coordinate space shared by the Eikonal and smoothness losses, and then define the global Eikonal, local Eikonal, second-order smoothness, normal prior, and ray-Eikonal losses.
The way these terms are weighted and scheduled during training is detailed in \cref{sec:implementation_details}.

\paragraph{Random Sampling for Losses.}
Several of our continuity losses (the Eikonal and smoothness terms) are evaluated at randomly sampled positions in the dense grid.
We first choose a dense cell coordinate $\mathbf{p}_i = (x_i, y_i, z_i) \in  \{0, \dots, G-1\}^3$ and add a random uniform offset $\boldsymbol{\delta} \sim \mathcal{U}(0, 1)^3$. 
Then we can get a query point $\mathbf{p}'_i = \mathbf{p}_i + \boldsymbol{\delta}$ in the unit cube volume cell $c(x_i,y_i,z_i)$. 
This random point $\mathbf{p}'_i$ serves as the root for loss propagation. To compute the numerical gradient at this location, we subsequently sample the six adjacent points (e.g., $\mathbf{p}'_i + (1, 0, 0)$) in the dense coordinate space. These samples allow us to approximate the gradient across unit cell boundaries, enforcing geometric consistency.

\paragraph{Eikonal Loss.}
To ensure the learned function $f$ behaves as a valid Signed Distance Function, we enforce the Eikonal constraint, $\|\nabla_{\mathbf{x}} f\|_2 \approx 1$, in world coordinates. We compute the gradient $\nabla_{\mathbf{x}} f$ at the perturbed dense coordinate $\mathbf{p}'_i = (x'_i, y'_i, z'_i)$.
Crucially, the SDF value $f(\mathbf{p})$ at any dense coordinate $\mathbf{p}$ is not stored directly, but must be computed using our voxel association mechanism and trilinear interpolation. This involves:
\begin{enumerate}
    \item Finding the dense cell index: $\mathrm{k} = \mathrm{idx}(\lfloor \mathbf{p} \rfloor)$ 
    \item Retrieving the enclosing voxel: $v = \mathrm{NVS}[\mathrm{k}]$ (\cref{subsec:sparse_voxel_association} of the manuscript).
    \item Interpolating the geometry: $f(\mathbf{p}) = \mathrm{interp}(geo_v, \boldsymbol{\delta})$, where $\boldsymbol{\delta} = \frac{\mathbf{x(p)} - v_{min}}{v_l} $ is the local offset and $\mathbf{x(p)}$ is a world coordinate of dense coordinate $\mathbf{p}$.
\end{enumerate}
\noindent To compute the gradient with respect to world coordinates ($\mathbf{x}$) while sampling in dense coordinates ($\mathbf{p}$), we use the central difference approximation with the correct world-space step size, $h_L$. The partial derivative $\partial_x f$ is:
\begin{equation}
    \partial_x f(\mathbf{p}'_i) = \frac{f(x'_i+1, y'_i, z'_i) - f(x'_i-1, y'_i, z'_i)}{2h_L},
\end{equation}
where $\partial_y f$ and $\partial_z f$ are computed similarly.
Here, the numerator $f(x'_i+1, \dots) - f(x'_i-1, \dots)$ queries the SDF values (in metric units) from the voxels enclosing the adjacent dense cells (found via NVS). The denominator $2h_L$ is the true world-space distance between these two sampling locations, ensuring the gradient is in metric units. The final Eikonal loss is:
\begin{equation}
    \mathcal{L}_{\text{eik}}=\sum_{i=1}^{N_\mathrm{samples}} \Bigl(\|\nabla_{\mathbf{x}} f(\mathbf{p}'_i)\|_2-1\Bigr)^2,
\end{equation}
where $N_\mathrm{samples}$ is the number of dense cell that have all six neighbor cells to calculate the loss.

\paragraph{Second-order (Laplacian) Smoothness Loss.}
We apply a Laplacian loss in world coordinates to encourage smoothness. We approximate the Laplacian $\nabla^2 f$ at the query point $\mathbf{p}'_i$ by querying the $f(\cdot)$ function at its neighbors, again normalizing by the world-space step size $h_L$:
\begin{equation}
    \partial_{xx} f(\mathbf{p}'_i) \approx \frac{f(x'_i+1, y'_i, z'_i) - 2f(\mathbf{p}'_i) + f(x'_i-1, y'_i, z'_i)}{h_L^2},
\end{equation}
where $\partial_{yy} f$ and $\partial_{zz} f$ are computed similarly. This term computes the curvature in metric units, where the value for each neighbor is found via NVS and interpolation. 
The final smoothness loss is: 
\begin{equation}
    \mathcal{L}_{\text{smooth}}=\sum_{i=1}^{N_\mathrm{samples}} \bigl\|\,\nabla^2 f(\mathbf{p}'_i)\,\bigr\|_1.
\end{equation}

\paragraph{Local Eikonal Loss.}
Whereas the Eikonal loss operates on randomly sampled dense coordinates, we replace it with a local Eikonal constraint at voxel centers after $L>9$. For each active voxel $v$ with center $\mathbf{c}_v$, we analytically compute the gradient $\nabla_{\mathbf{x}} f(\mathbf{c}_v)$ by differentiating the trilinear interpolation function of its eight corner SDF values ($geo_v$).
Specifically, the $x$-component of the gradient is computed by averaging the differences along the $x$-edges:

\begin{equation}
\frac{\partial f}{\partial x} \approx \frac{1}{4 h_v} \sum_{j,k \in \{0,1\}} \left( f_{1,j,k} - f_{0,j,k} \right)
\end{equation}
where $f_{i,j,k} \in geo_{v}$ are the corner SDF values and $h_v$ is the voxel size. We enforce $\|\nabla_{\mathbf{x}} f(\mathbf{c}_v)\|_2 \approx 1$ at the voxel center:
\begin{equation}
    \mathcal{L}_{\text{le}}
=
\sum_{i}^{N_\mathrm{vox}}
\Bigl(\,\|\nabla_{\mathbf{x}} f(\mathbf{c}_v)\|_2 - 1\,\Bigr)^2,
\end{equation}
where $N_\mathrm{vox}$ is the number of voxels including all parents and children. For efficiency, this term is evaluated only on a random subset of active voxels, as described in the implementation details.

\paragraph{Normal Loss with Geometric Prior.}
In addition to the internal consistency losses ($\mathcal{L}_{\text{eik}}$, $\mathcal{L}_{\text{smooth}}$), we leverage the explicit geometric priors from the PI$^3$ point maps (introduced in \cref{subsec:sparse_voxel_initialization_using_pointmap}) to guide the surface orientation. For each camera view $i$, we compute a 2D prior normal map $\mathbf{N}_{\text{prior}}$ from the per-view point map $P_i$.

To compare against this prior, we must render our model's normals to the image plane. We first define the normal vector $\mathbf{n}_v$ for a voxel $v$ by taking the analytical gradient of the trilinear interpolation function (with respect to the local voxel coordinates $\mathbf{q}$) and evaluating it at the voxel center $\mathbf{q}_c=(0.5,0.5,0.5)$:
\begin{equation}
    \mathbf{n}_v = \mathrm{normalize}\bigl(\nabla_{\mathbf{q}} \, \mathrm{interp}(geo_v, \mathbf{q}_c)\bigr),
\end{equation}
where $\operatorname{normalize}(\cdot)$ denotes the operation of scaling the input vector to unit length (i.e., $L_2$ normalization).

We then render the expected normal map by accumulating these per-voxel normals. For a ray corresponding to a pixel $\mathbf{u}$, we use the same volumetric rendering equation as the color pass \cref{eq:vol_color}, simply replacing the per-voxel color $\mathbf{c}_i$ with the per-voxel normal $\mathbf{n}_i$:
\begin{equation}
    \mathbf{N}_{\text{rendered}}(\mathbf{u}) = \sum_{i=1}^{N_\mathrm{ray}} T_i \alpha_i \mathbf{n}_i.
    \label{eq:rendered_normal}
\end{equation}

The final normal loss $\mathcal{L}_{\text{normal}}$ is a robust cosine distance loss between the rendered normal map and the prior normal map, computed on all valid pixels $\mathbf{u}$ in the image:
\begin{equation}
    \mathcal{L}_{\text{normal}} = \frac{1}{|\mathcal{M}|} \sum_{\mathbf{u} \in \mathcal{M}} \Bigl( 1 - \mathbf{N}_{\text{rendered}}(\mathbf{u}) \cdot \mathbf{N}_{\text{prior}}(\mathbf{u}) \Bigr),
\end{equation}
where $\mathcal{M}$ is the set of valid pixels and $|\mathcal{M}|$ denotes the number of such pixels.

While this loss provides a valuable geometric guide, the $\mathbf{N}_{\text{prior}}$ map is derived from imperfect point maps and can be noisy. Therefore, we apply a weight decay schedule, annealing the loss to zero in the final fine-tuning stages, allowing the high-fidelity internal losses to define the final geometry.

\paragraph{Ray-Eikonal Loss.}
Ray-eikonal loss $\mathcal{L}_\mathrm{re}$ is a ray-based Eikonal regularizer, which is implemented only for the naive baseline: SVRaster+SDF. For each ray segment that intersects a voxel, we determine the entry and exit coordinates, $\mathbf{p}_{\text{in}}$ and $\mathbf{p}_{\text{out}}$, and evaluate the SDF values $f(\mathbf{p}_{\text{in}})$ and $f(\mathbf{p}_{\text{out}})$ via trilinear interpolation of the voxel's corner SDFs. Let $s = \|\mathbf{p}_{\text{out}} - \mathbf{p}_{\text{in}}\|_2$ be the segment length. We approximate the directional gradient along the ray as:
\begin{equation}
g = \frac{f(\mathbf{p}_{\text{out}}) - f(\mathbf{p}_{\text{in}})}{s}.
\end{equation}
We then penalize deviations from the expected SDF slope (assuming the ray traverses free space or enters the surface):
\begin{equation}
\mathcal{L}_\mathrm{re} = \sum_{i=1}^{N_\mathrm{ray}} w_i \bigl(g_i + 1\bigr)^2,
\end{equation}
where $w_i = T_i \alpha_i$ is the standard visibility weight for the segment, and $N_\mathrm{ray}$ is the number of intersected voxels. This loss encourages a consistent signed gradient ($g \approx -1$, decreasing SDF along the ray) along occupied segments.

%% file: supp/c_additional_implementation_detail.tex
\section{Implementation Details}
\label{sec:implementation_details}

\paragraph{Overall Loss Functions and Schedules.}
\input{figs/12_optimization_overview/12_optimization_overview}
We first draw the overall optimization strategy in ~\cref{fig:optimization-overview}. For DTU scenes we train for $8{,}000$ iterations with the following objective:
\begin{multline}
\label{eq:loss_dtu}
    \mathcal{L}_\mathrm{total}
    = \mathcal{L}_\mathrm{photo}
    + \lambda_n \mathcal{L}_\mathrm{normal}
    + \lambda_e \mathcal{L}_\mathrm{eik} \\
    + \lambda_{le} \mathcal{L}_{le}
    + \lambda_s \mathcal{L}_\mathrm{smooth}
    + \lambda_m \mathcal{L}_\mathrm{mask} \\
    + \lambda_{\text{n-dmean}} \mathcal{L}_{\text{n-dmean}}
    + \lambda_{\text{n-dmed}} \mathcal{L}_{\text{n-dmed}},
\end{multline}
where lambda is specified as below.

The photometric term $\mathcal{L}_\mathrm{photo}$ follows SVRaster~\cite{Sun_2025_CVPR_SVRaster} and combines the rendered RGB reconstruction loss, SSIM loss, and the $\mathcal{L}_R$ color concentration loss. All weights for these components are kept identical to SVRaster. The mask term $\mathcal{L}_{\text{mask}}$ enforces the transmittance of pixels outside the mask to be 1 (transparent) and those inside the mask to be 0 (opaque). This term is included to improve quantitative reconstruction results, with the weight set to $\lambda_m = 1.0$.

\paragraph{Normal Supervision.}
The normal loss $\mathcal{L}_\mathrm{normal}$ is applied with
\begin{equation}
    \lambda_n = \left\{ \begin{array}{cl}
        0.10 & : \ 0 \le \tau < 4000 \\
        0.01 & : \ 4000 \le \tau < 6000 \\
        0.00 & : \ \mathrm{else}
\end{array} \right.,
\end{equation}
where $\tau$ denotes iterations.

\paragraph{Global Eikonal Loss.}
The dense Eikonal term $\mathcal{L}_\mathrm{eik}$ is weighted by $\lambda_e = 10^{-8}$. Since this loss is evaluated on dense grid cells, the number of samples roughly quadruples whenever the dense resolution increases by one level. To keep the overall contribution of this loss balanced, we decay its weight by a factor of $0.25$ every $2{,}000$ iterations, and we apply it only up to the $2^9$ resolution level, that is, until $\tau = 6{,}000$.

\paragraph{Local Eikonal Loss.}
We additionally use a local Eikonal regularizer $\mathcal{L}_{le}$ at the centers of parent and child cells to encourage unit gradient magnitude near the surface. For DTU scenes, we set $\lambda_{le} = 10^{-11}$ and activate this term only in the fine stages, for iterations $6{,}000 \le \tau \le 8{,}000$. For stability, we evaluate $\mathcal{L}_{le}$ on each eligible cell with probability $0.5$ and scale the weight accordingly so that the expected contribution matches $\lambda_{le}$.

\paragraph{Smoothness Loss.}
The smoothness term $\mathcal{L}_\mathrm{smooth}$ uses $\lambda_s = 10^{-10}$. As with the dense Eikonal loss, the number of samples increases with the dense resolution, so we decay $\lambda_s$ by a factor of $0.25$ every $2{,}000$ iterations during the coarse levels. After the $2^9$ level (from $\tau = 6{,}000$ to $8{,}000$), we stop decaying and keep $\lambda_s$ fixed while applying the loss in the parent–child hierarchy.

\paragraph{Additional Regularizers.}
Following SVRaster~\cite{Sun_2025_CVPR_SVRaster}, we also include the $\mathcal{L}_{\text{n-dmean}}$ and $\mathcal{L}_{\text{n-dmed}}$ regularizers with
\[
\lambda_{\text{n-dmean}} = 0.001, \quad
\lambda_{\text{n-dmed}} = 0.001.
\]
We start applying $\mathcal{L}_{\text{n-dmed}}$ from iteration $\tau = 1{,}000$ and $\mathcal{L}_{\text{n-dmean}}$ from $\tau = 2{,}000$. Their definitions are identical to those in SVRaster~\cite{Sun_2025_CVPR_SVRaster}.

\paragraph{Tanks-and-Temples Training.}
For Tanks-and-Temples scenes, we train for $10{,}000$ iterations with the same loss terms and base coefficients as in Eq.~\eqref{eq:loss_dtu} except using $\mathcal{L}_{\text{mask}}$. The only difference is the schedule of the regularizers. We use
\begin{equation}
    \lambda_n = \left\{ \begin{array}{cl}
        0.01 & : \ 0 \le \tau < 4000 \\
        0.005 & : \ 4000 \le \tau < 8000 \\
        0.00 & : \mathrm{else}
\end{array} \right..
\end{equation}
 The local Eikonal term $\mathcal{L}_{le}$ is enabled for $6{,}000 \le \tau \le 10{,}000$, and the smoothness loss $\mathcal{L}_\mathrm{smooth}$ is kept active until $\tau = 10{,}000$ without further decay. All other settings follow the DTU configuration described above.

\paragraph{Implementation of Baseline.}
For a fair comparison, we implement a voxel-based SDF baseline within our framework. The initialization is deliberately simple: we first define a uniform dense grid at $L=6$ ($G=2^6$) within the predefined cube region. We then assign SDF values to the voxel corners by measuring the Euclidean distance from the center of this cube region to each corner and subtracting one-fifth of the cube edge length. This produces an approximate spherical zero-level set inside the volume, without using any external geometric priors.

Following SVRaster~\cite{Sun_2025_CVPR_SVRaster}, we train this baseline for 20,000 iterations, applying pruning and subdivision every 1,000 iterations. The loss function is defined as:
\begin{multline}
\mathcal{L}_\mathrm{total} =
\mathcal{L}_\mathrm{photo}
+ \lambda_\mathrm{re} \mathcal{L}_\mathrm{re} \\
+ \lambda_{\text{n-dmean}} \mathcal{L}_{\text{n-dmean}}
+ \lambda_{\text{n-dmed}} \mathcal{L}_{\text{n-dmed}},
\end{multline}
where $\mathcal{L}_\mathrm{photo}$ (photometric loss), $\mathcal{L}_{\text{n-dmean}}$, and $\mathcal{L}_{\text{n-dmed}}$ follow the same weights as in our method (\cref{eq:loss_dtu}), and we set $\lambda_\mathrm{re} = 10^{-3}$.

%% file: figs/12_optimization_overview/12_optimization_overview.tex
\begin{figure*}
    \centering
    \includegraphics[width=1.0\linewidth]{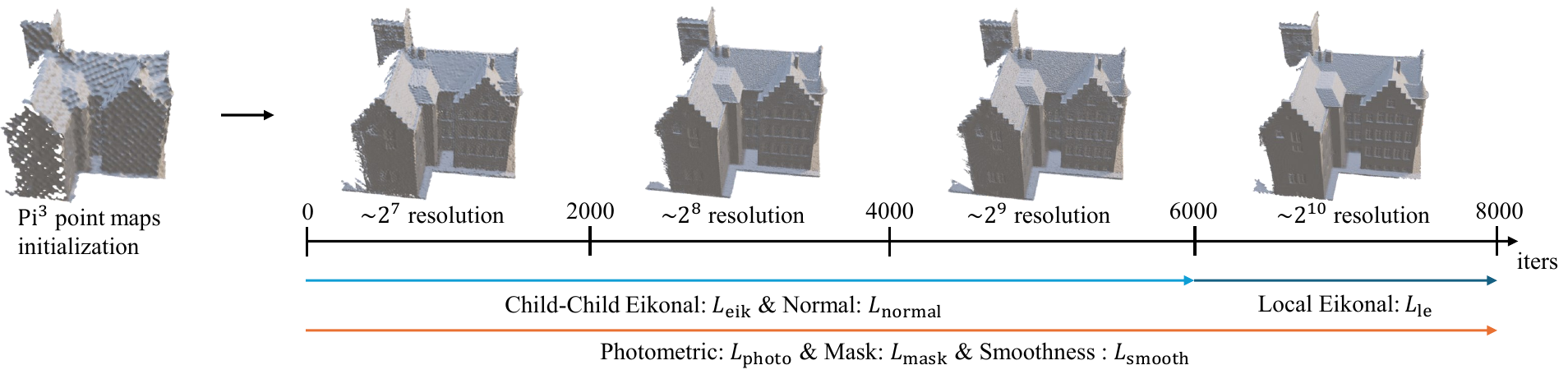}
    \caption{Optimization strategy.}
    \label{fig:optimization-overview}
\end{figure*}

%% file: figs/13_normal_until_end/13_normal_until_end.tex

\begin{figure*}[t]
  \centering
  \includegraphics[width=1.0\linewidth]{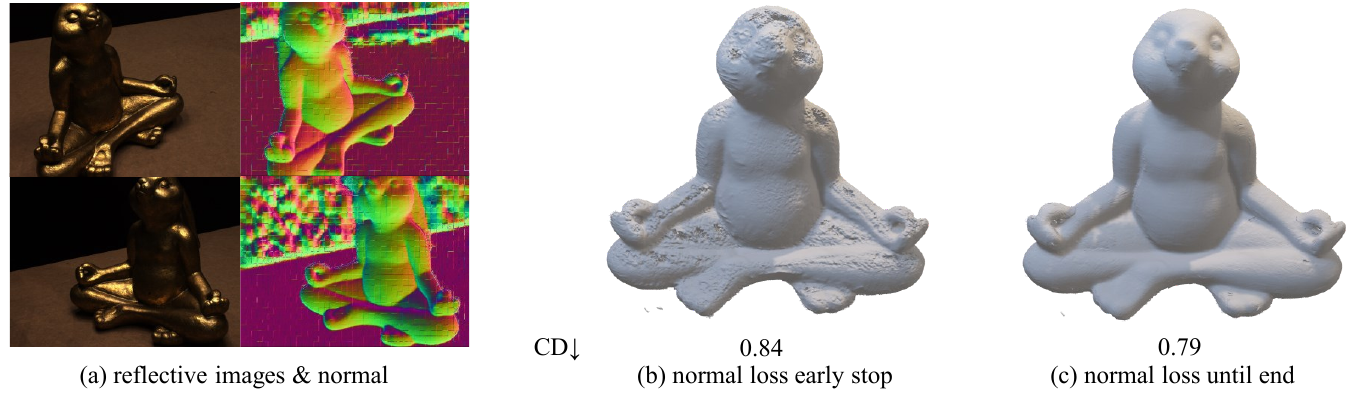}
  \vspace{-4mm}
  \caption{(a) Reflective views and corresponding normal cue from DTU scan 110. (b) Our original normal-loss scheduling. (c) Extending the normal loss until the last iteration. Extending the normal-loss schedule enables our model to better handle reflective surfaces, improving the Chamfer Distance from 0.84 to 0.79 and yielding visually cleaner reconstructions.
  }
  \label{fig:normal_until}
\end{figure*}

%% file: figs/14_real-world_result/14_real_world_result.tex
\begin{figure*}[t]
  \centering
  \includegraphics[width=1.0\linewidth]{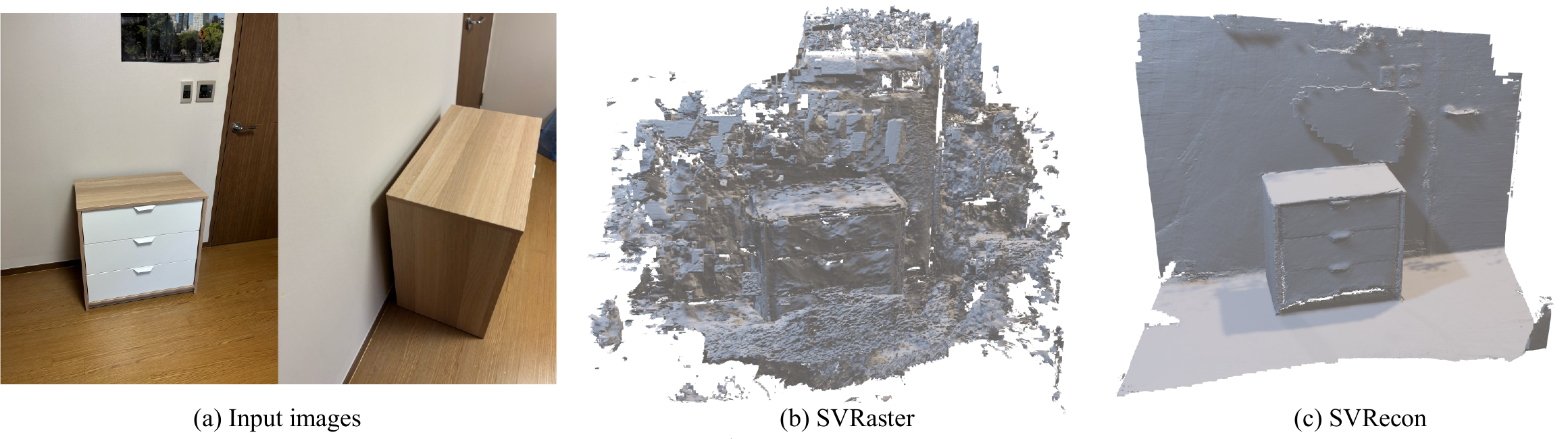}
  \vspace{-6mm}
  \caption{\textbf{Real-world reconstruction on an indoor drawer scene.}
(a) Input images, (b) reconstruction by SVRaster, and (c) reconstruction by our SVRecon, which yields a cleaner and more coherent mesh with fewer artifacts.}
  \label{fig:real-world}
\end{figure*}

%% file: supp/e_additional_ablation.tex
\section{Ablation Study}
\label{sec:ablation_study_supp}

\paragraph{Initialization in Outdoor Scenes.}
We find that, for outdoor scenes, initializing the region outside the inner reconstruction box is critical for mesh quality. When the scene is learned with SDF values, background regions that should carry non-negligible density can be initialized with very large SDF magnitudes. In that case, it becomes difficult for density to emerge there during optimization, and parts of the background are instead explained as extensions of the target mesh. 
\input{table/c_ablation_outside}    
\input{figs/11_outdoor_initialization/11_outdoor_initialization}

For the Tanks-and-Temples \textit{Truck} scene, we conduct an ablation over three settings: (1) training without any outside box, (2) extending the PI$^3$-based initialization to the outside box, and (3) additionally adding a spherical shell that encloses the sky (ours); see ~\cref{tab:ablation_outside}. However, our approach still has failure cases. For example, in the TnT \textit{Caterpillar} scene (see ~\cref{fig:caterpillar-compare}), the sky color varies significantly across the image, and parts of the sky are still reconstructed as geometry attached to the caterpillar mesh. This is a limitation of our current design.

\paragraph{Handling Reflective Objects.}
Reflective objects in DTU (e.g., scan 110) reveal a limitation of our method: the network can overfit to specular highlights, leading to noisy geometry. As shown in ~\cref{fig:normal_until}, we address this by simply extending the normal loss until the end of training. This extended normal supervision better constrains the surface orientation within reflective regions, reduces floating artifacts, and yields cleaner reconstructions. On scan 110, the Chamfer Distance improves from 0.84 to 0.79, and the resulting mesh is visually smoother and more stable on reflective surfaces.

%% file: table/c_ablation_outside.tex
\begin{table}[t]
\centering
\caption{\textbf{Effect of outside-scene initialization on TnT \textit{Truck}.}
Adding an outside bounding region and further initializing a sky sphere
improves both precision and F1.}
\label{tab:ablation_outside}
\resizebox{\columnwidth}{!}{
\begin{tabular}{lccc}
\toprule
Metric & (1) w/o outside box & (2) w/ outside box only & (3) Ours \\
\midrule
Precision ($\uparrow$) & 0.33 & 0.46 & \textbf{0.52} \\
Recall ($\uparrow$)    & 0.45 & 0.61 & \textbf{0.63} \\ 
F1 Score ($\uparrow$)  & 0.42 & 0.52 & \textbf{0.57} \\
\bottomrule
\end{tabular}
}
\end{table}

%% file: figs/11_outdoor_initialization/11_outdoor_initialization.tex
\begin{figure}[t]
  \centering
  \includegraphics[width=1.0\columnwidth]{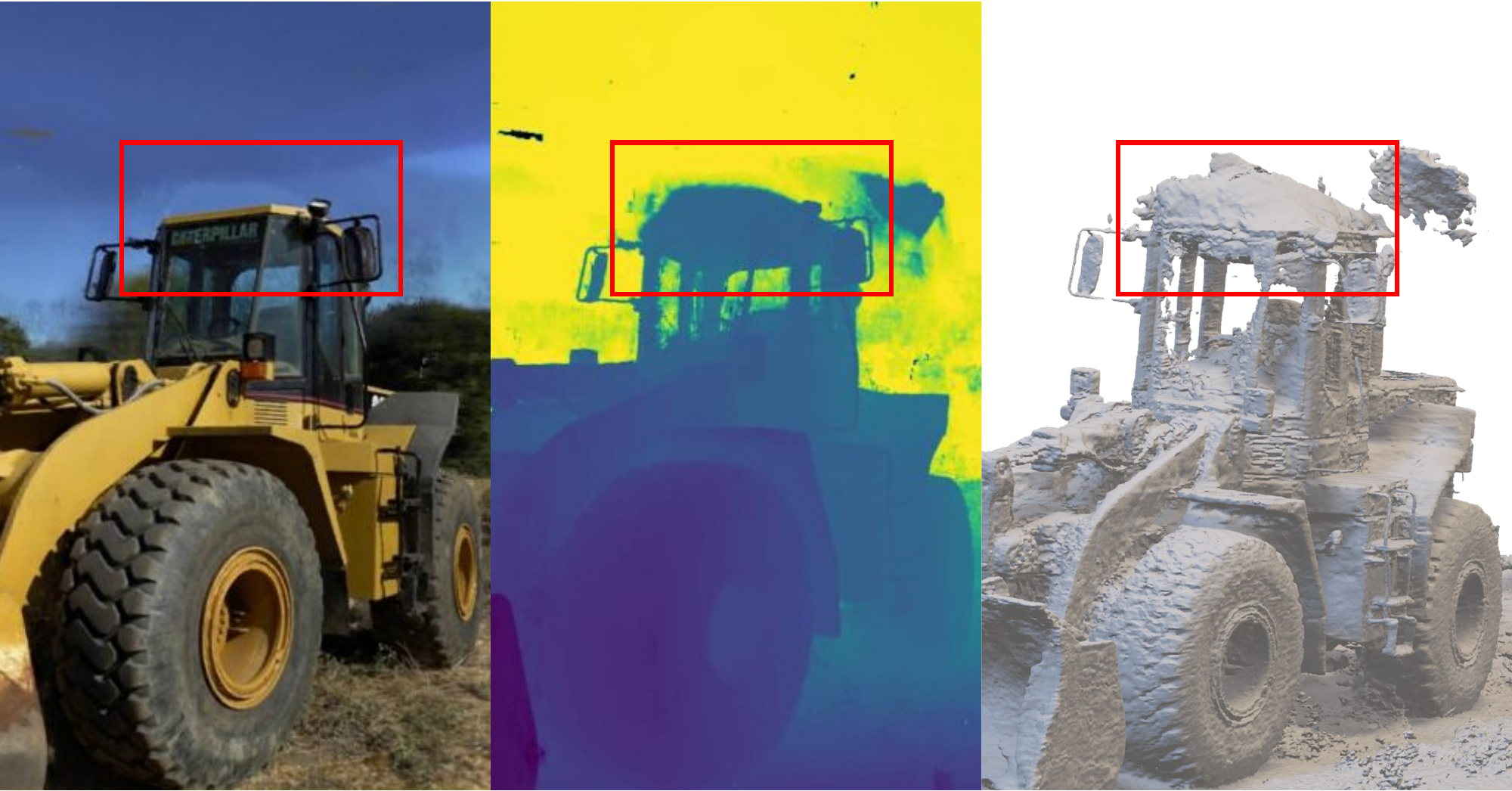}
  \caption{\textbf{Failure case in outdoor initialization.} From left to right, we
  show the rendered RGB image, rendered depth, and extracted mesh. In the TnT \textit{Caterpillar} scene, parts of the sky are incorrectly
  reconstructed as geometry attached to the object due to incomplete
  initialization of the outer (background) region. }
  \label{fig:caterpillar-compare}
\end{figure}

%% file: supp/d_full_qualitative_result.tex
\section{Qualitative Results}
\label{sec:qualitative_results}

\paragraph{Real-world Reconstruction}
We also evaluate our method on a real-world indoor scene captured with 36 handheld RGB images focusing on a drawer, as shown in \cref{fig:real-world}. Compared to SVRaster, which produces a highly noisy and fragmented reconstruction in this setting, our method recovers a clean and coherent mesh that faithfully represents both the drawer and the surrounding structures, indicating improved robustness to real-world noise and calibration imperfections.
\input{figs/10_full_qualitative/10_full_qualitative}
\input{figs/10_1_full_qualitativeTnT/10_1_full_qualitativeTnT}
\paragraph{Qualitative Result of DTU \& TnT}
For completeness, we provide full qualitative visualizations of our
reconstructed meshes on all DTU and Tanks-and-Temples scenes in
\cref{fig:qual_mesh_dtu} and \cref{fig:qual_mesh_tnt}.

Across both indoor (DTU) and outdoor (Tanks-and-Temples) scenes, the
reconstructions exhibit largely hole-free, high-fidelity surfaces, thanks
to the SDF-based representation and our regularization losses. Fine
geometric details such as thin structures and concave regions are preserved,
while spurious floaters and fragmented components are largely suppressed.

%% file: figs/10_full_qualitative/10_full_qualitative.tex
\begin{figure*}[t]
    \centering

    \begin{subfigure}{.19\linewidth}
        \centering
        \includegraphics[width=\linewidth]{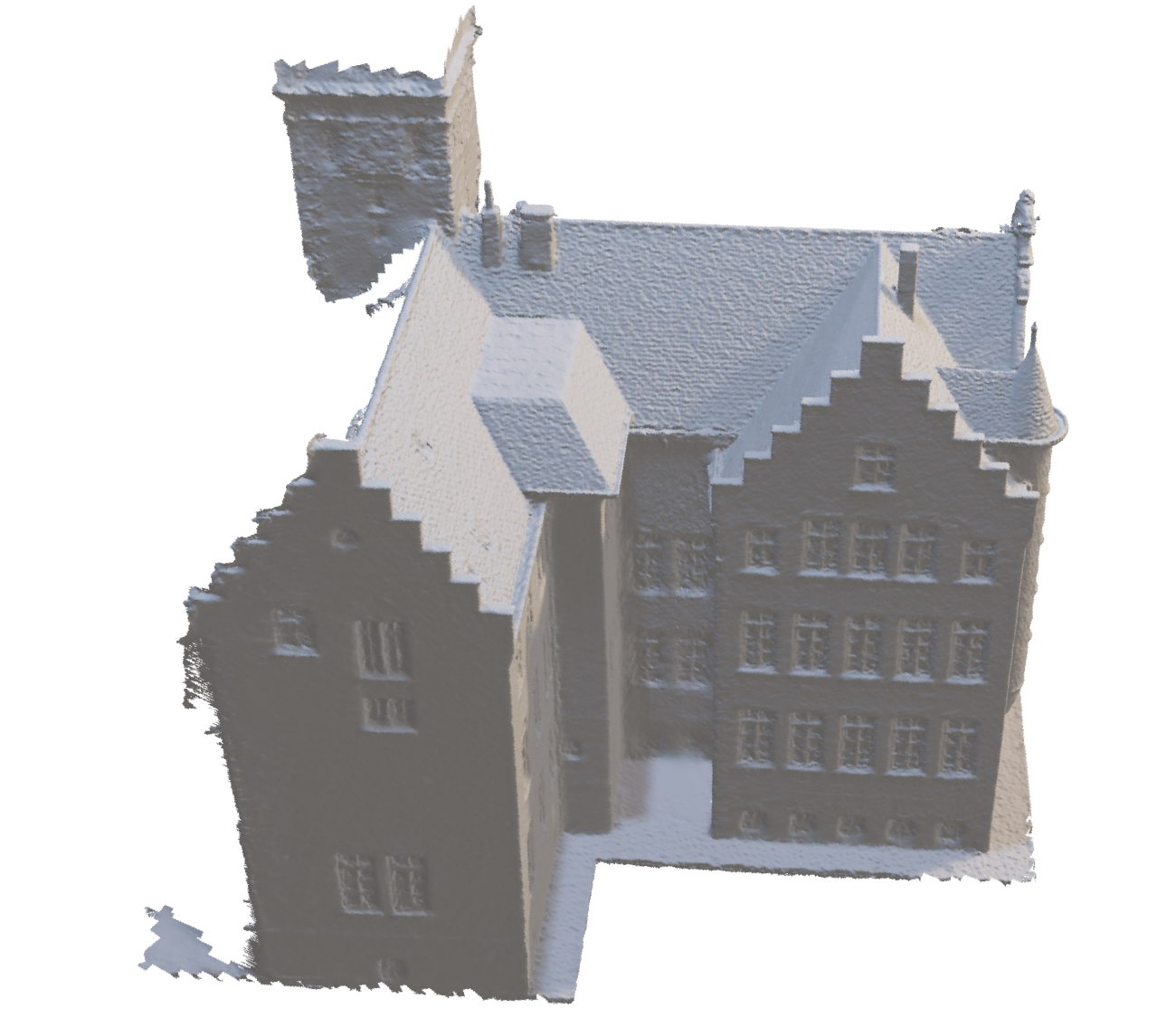}
    \end{subfigure}\hfill
    \begin{subfigure}{.19\linewidth}
        \centering
        \includegraphics[width=\linewidth]{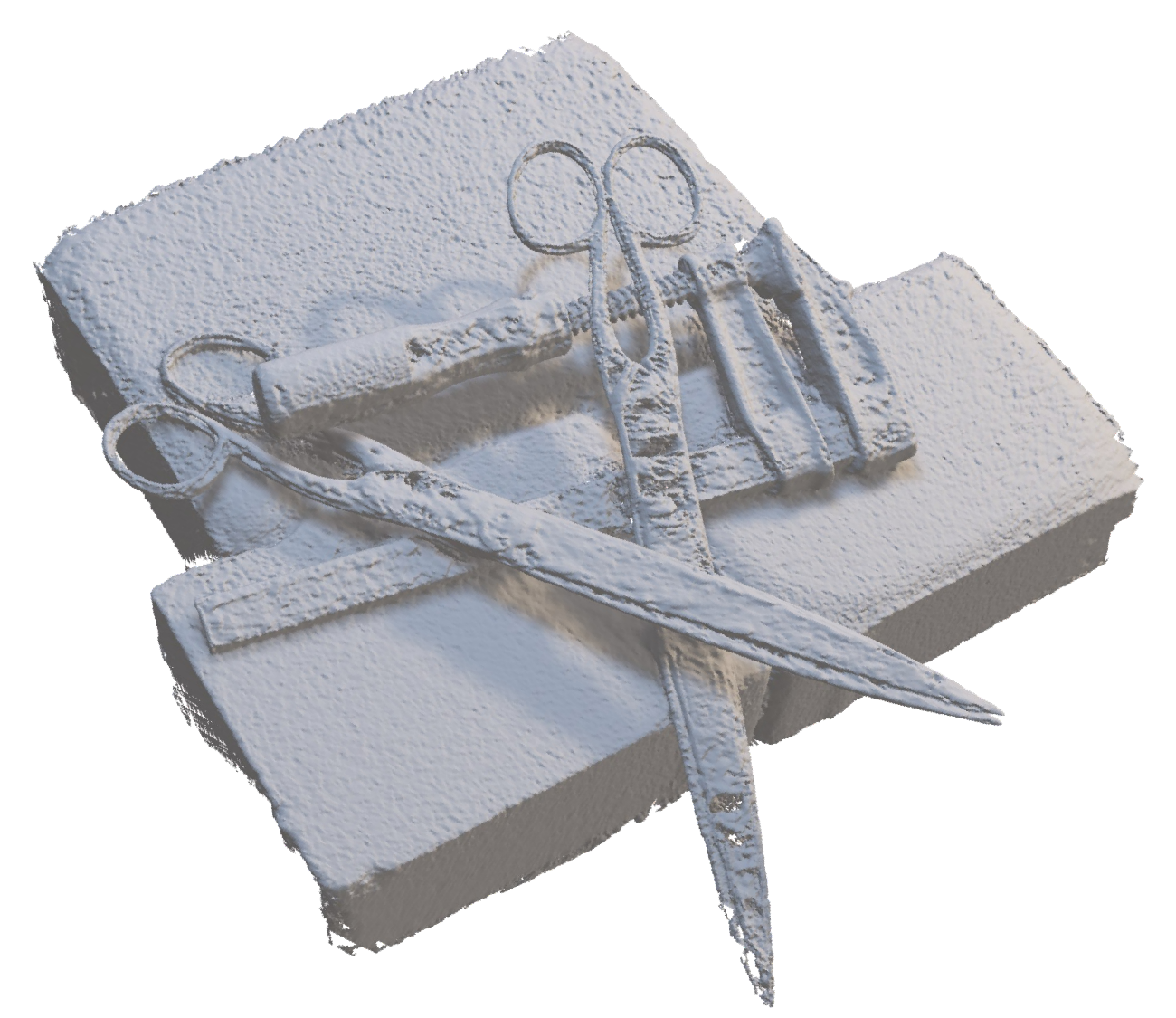}
    \end{subfigure}\hfill
    \begin{subfigure}{.19\linewidth}
        \centering
        \includegraphics[width=\linewidth]{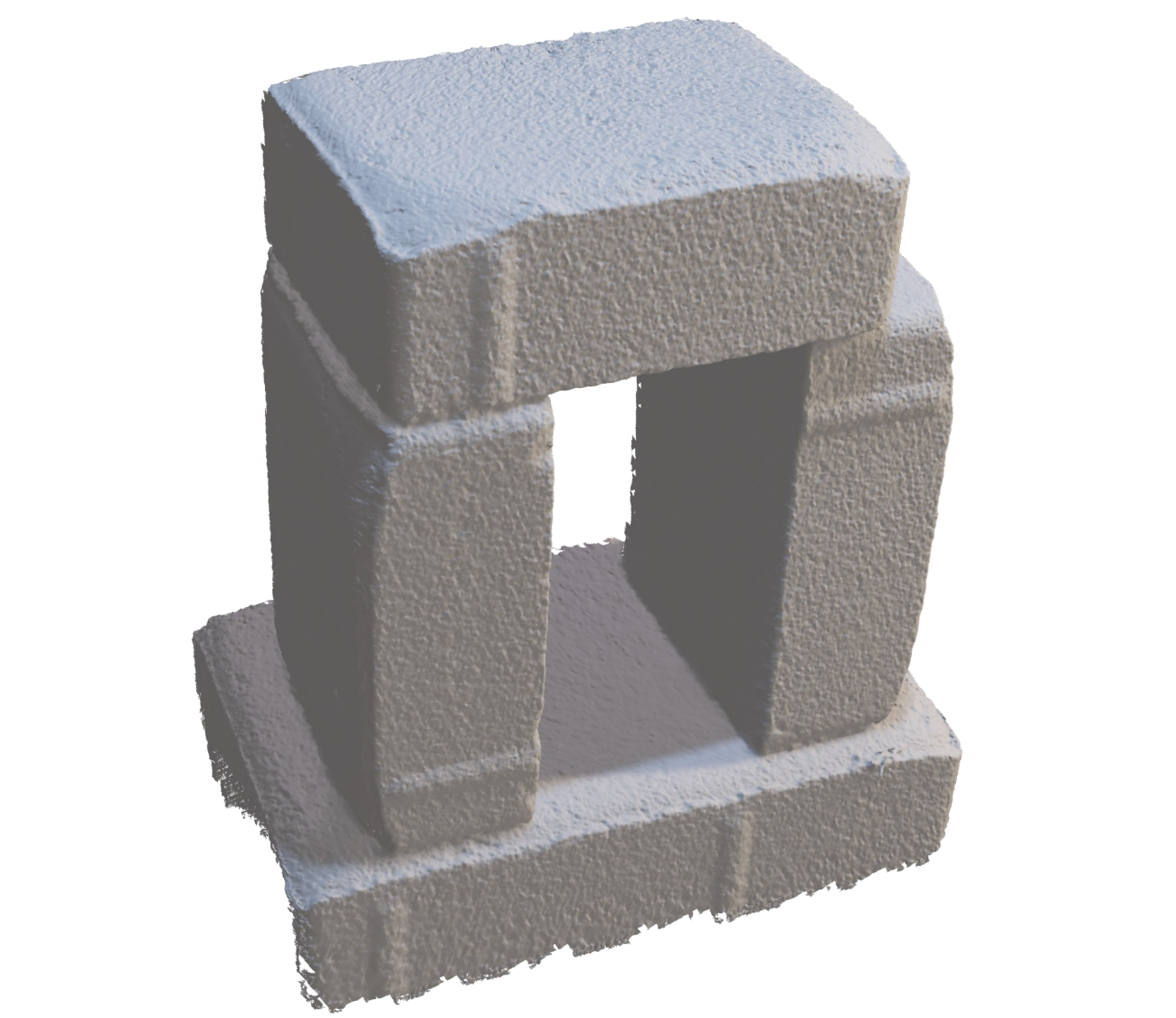}
    \end{subfigure}\hfill
    \begin{subfigure}{.19\linewidth}
        \centering
        \includegraphics[width=\linewidth]{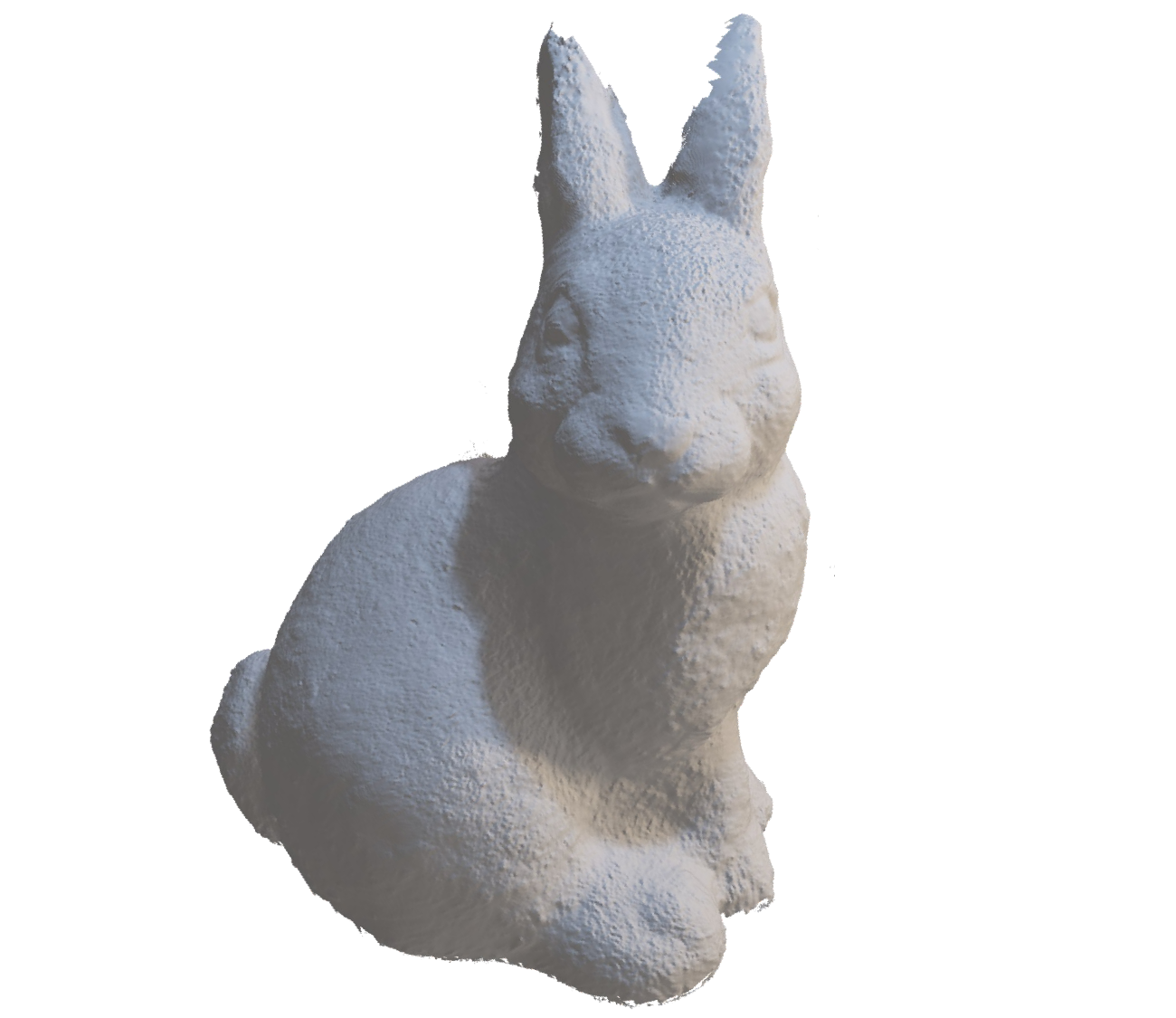}
    \end{subfigure}\hfill
    \begin{subfigure}{.19\linewidth}
        \centering
        \includegraphics[width=\linewidth]{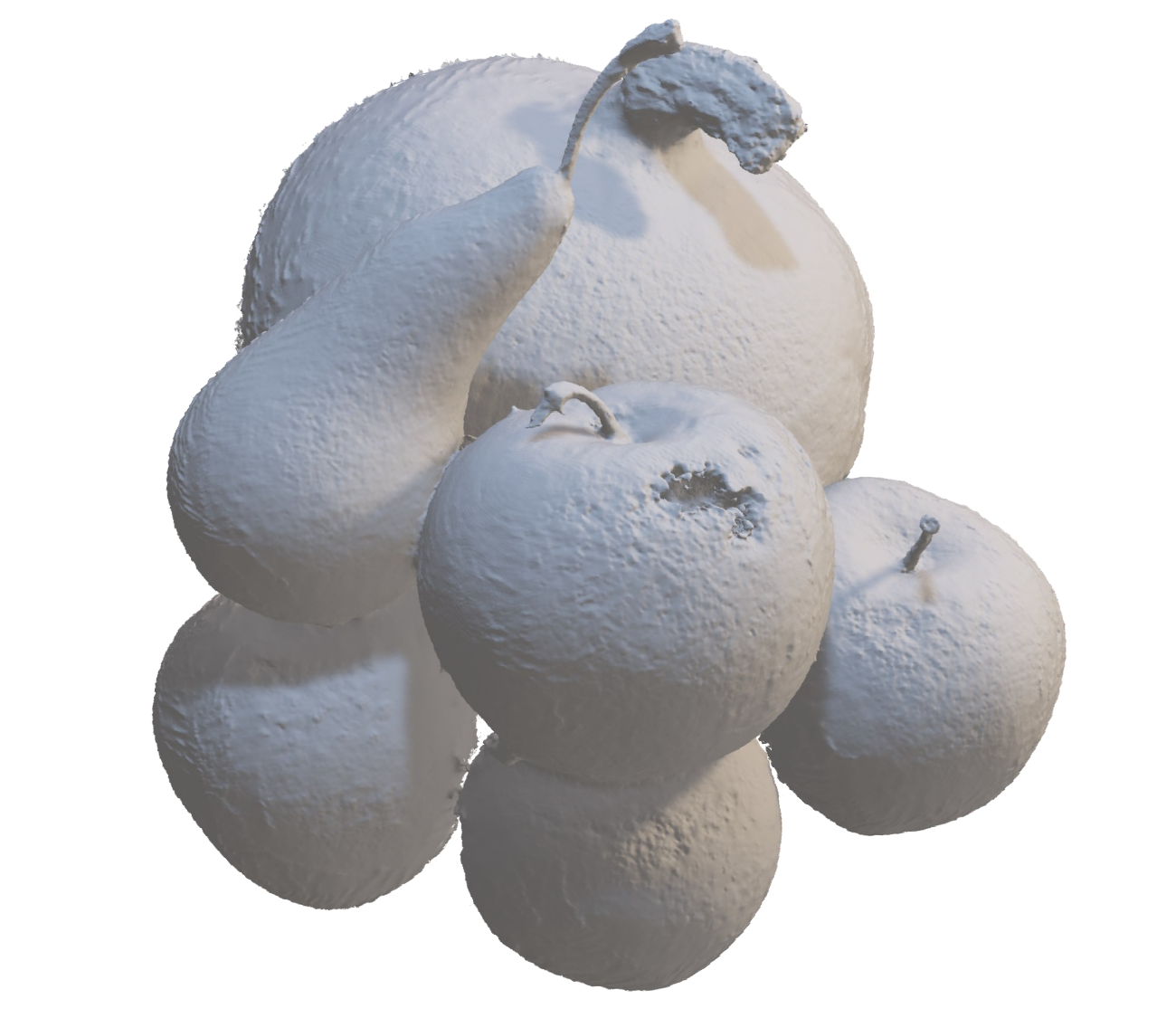}
    \end{subfigure}
    \\[0.3em]

    \begin{subfigure}{.19\linewidth}
        \centering
        \includegraphics[width=\linewidth]{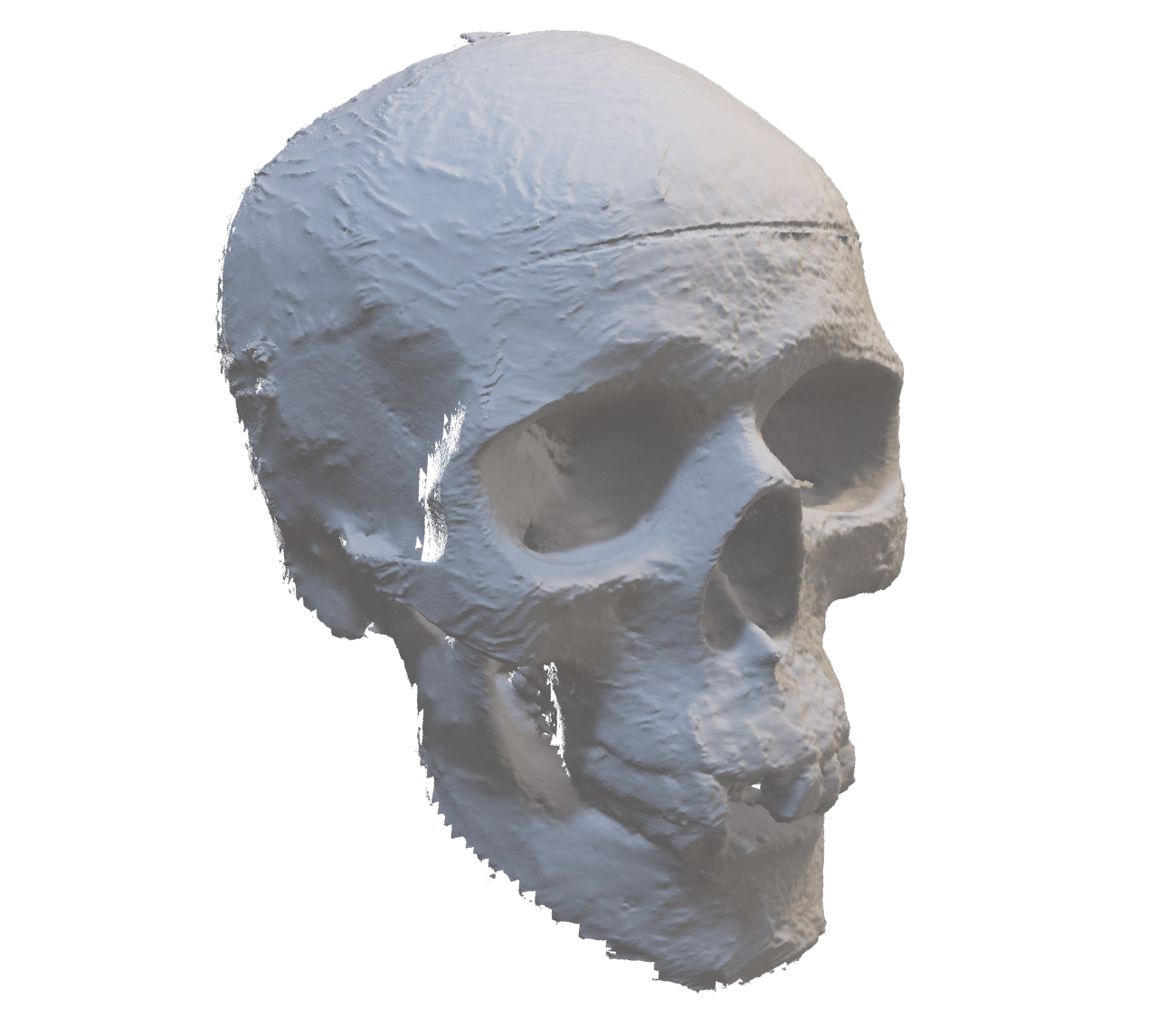}
    \end{subfigure}\hfill
    \begin{subfigure}{.19\linewidth}
        \centering
        \includegraphics[width=\linewidth]{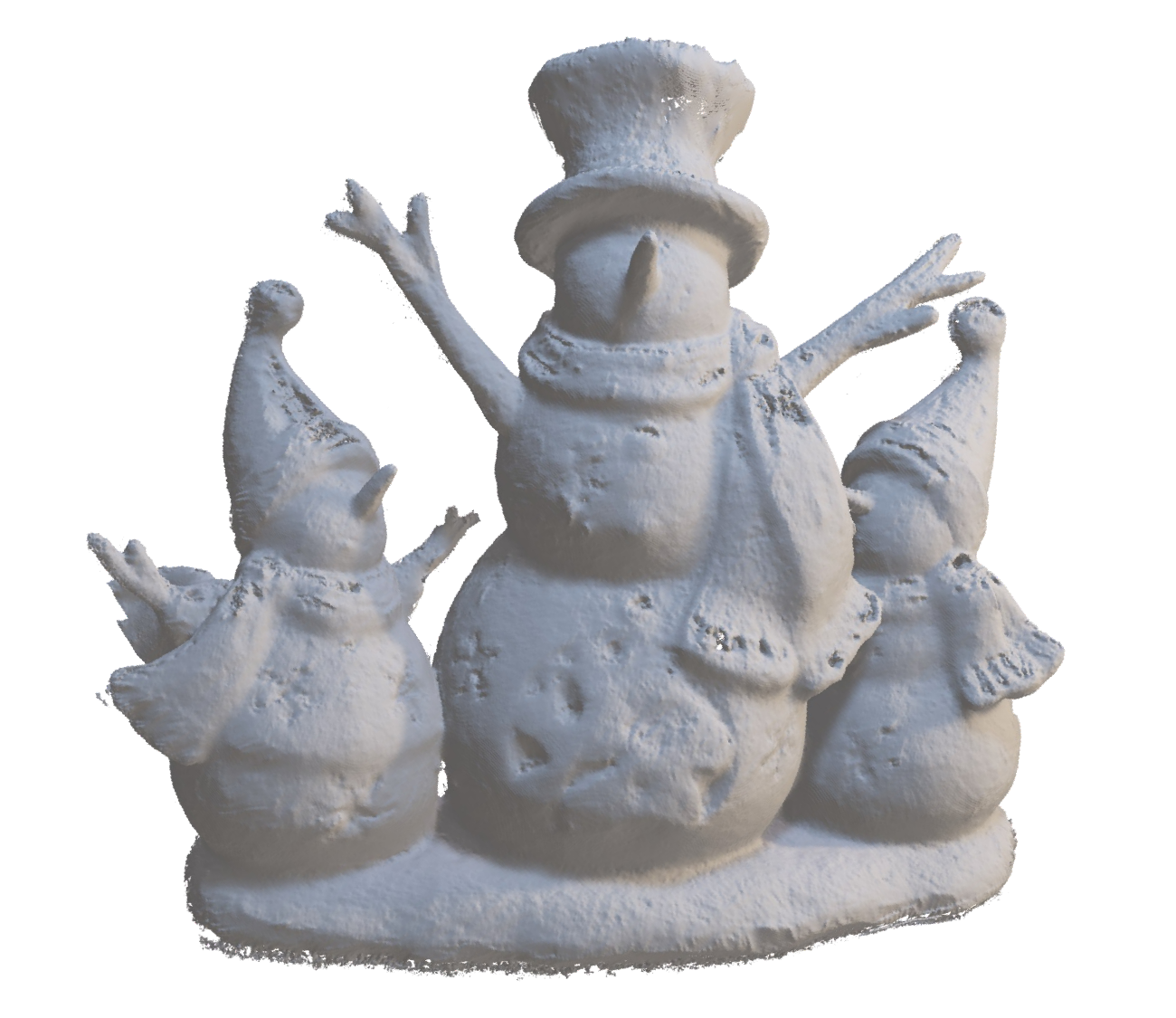}
    \end{subfigure}\hfill
    \begin{subfigure}{.19\linewidth}
        \centering
        \includegraphics[width=\linewidth]{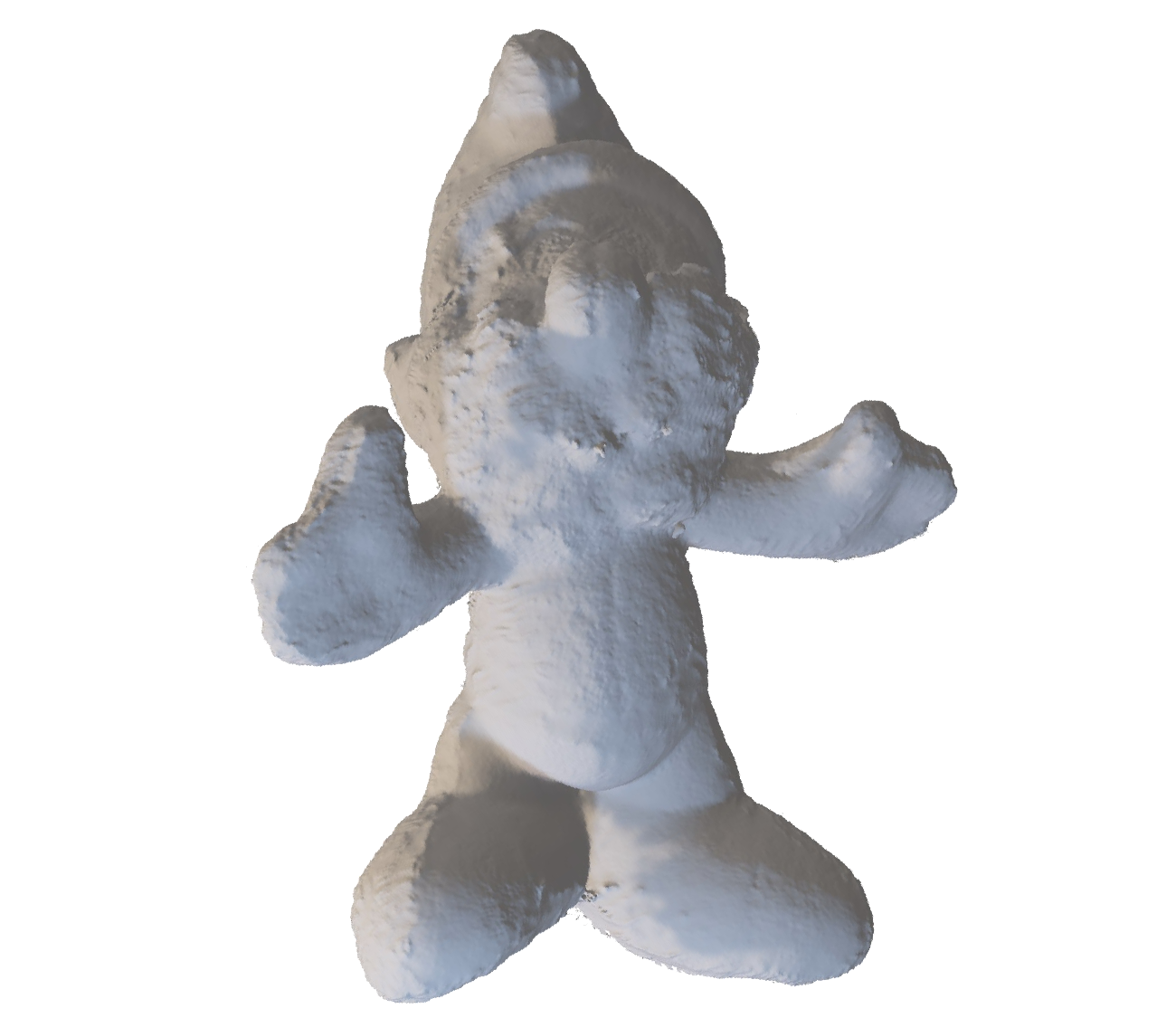}
    \end{subfigure}\hfill
    \begin{subfigure}{.19\linewidth}
        \centering
        \includegraphics[width=\linewidth]{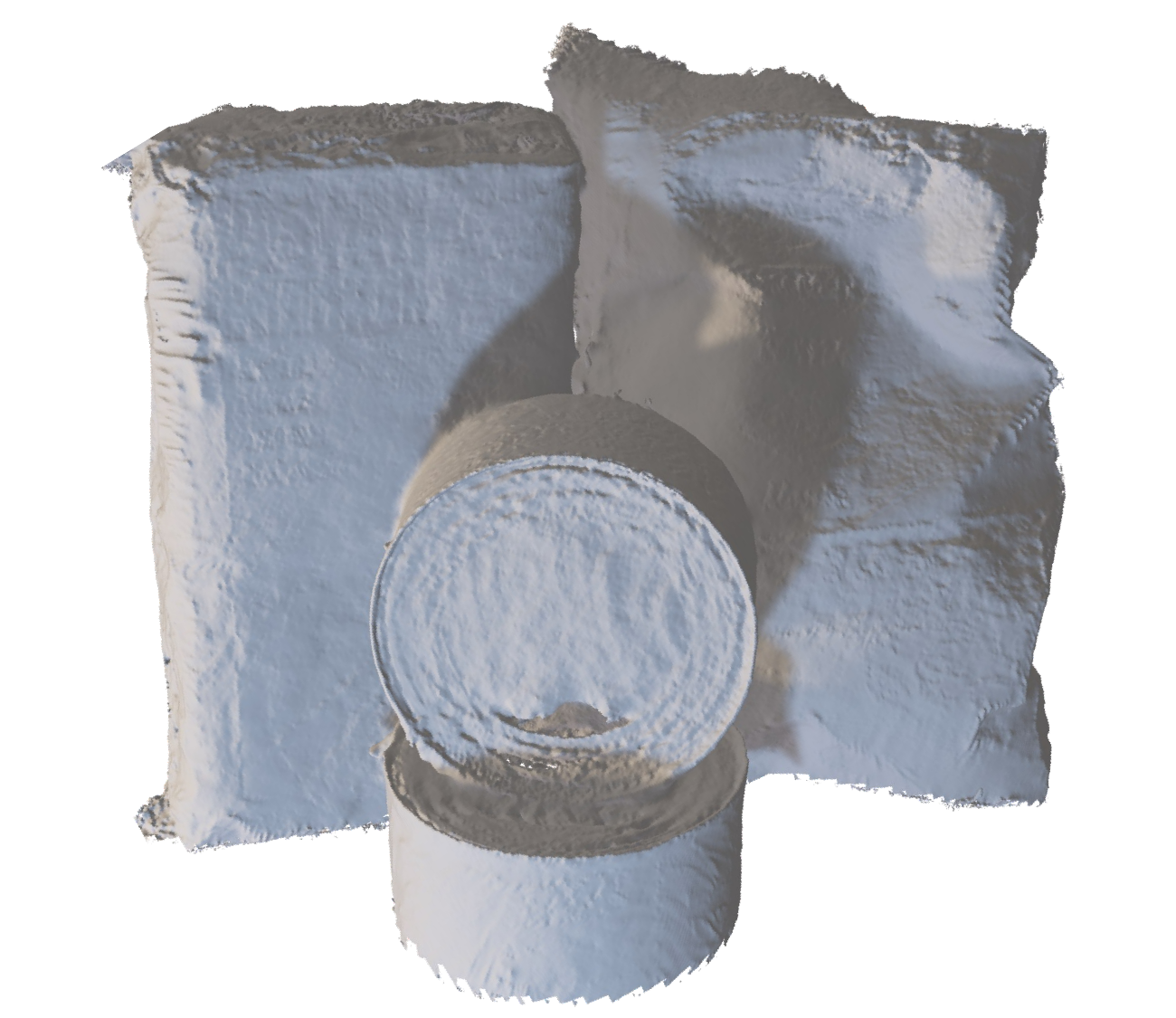}
    \end{subfigure}\hfill
    \begin{subfigure}{.19\linewidth}
        \centering
        \includegraphics[width=\linewidth]{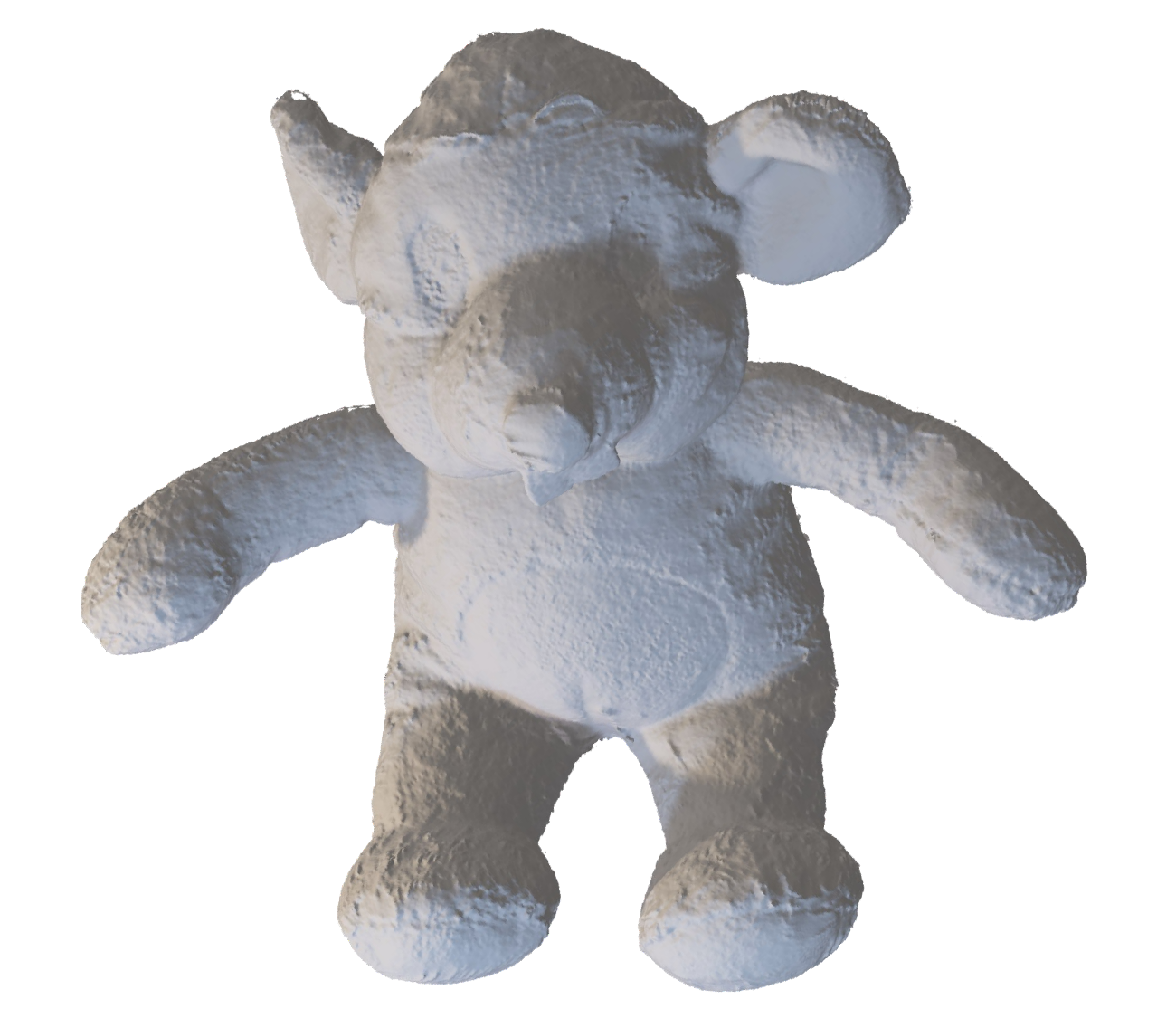}
    \end{subfigure}
    \\[0.3em]

    \begin{subfigure}{.19\linewidth}
        \centering
        \includegraphics[width=\linewidth]{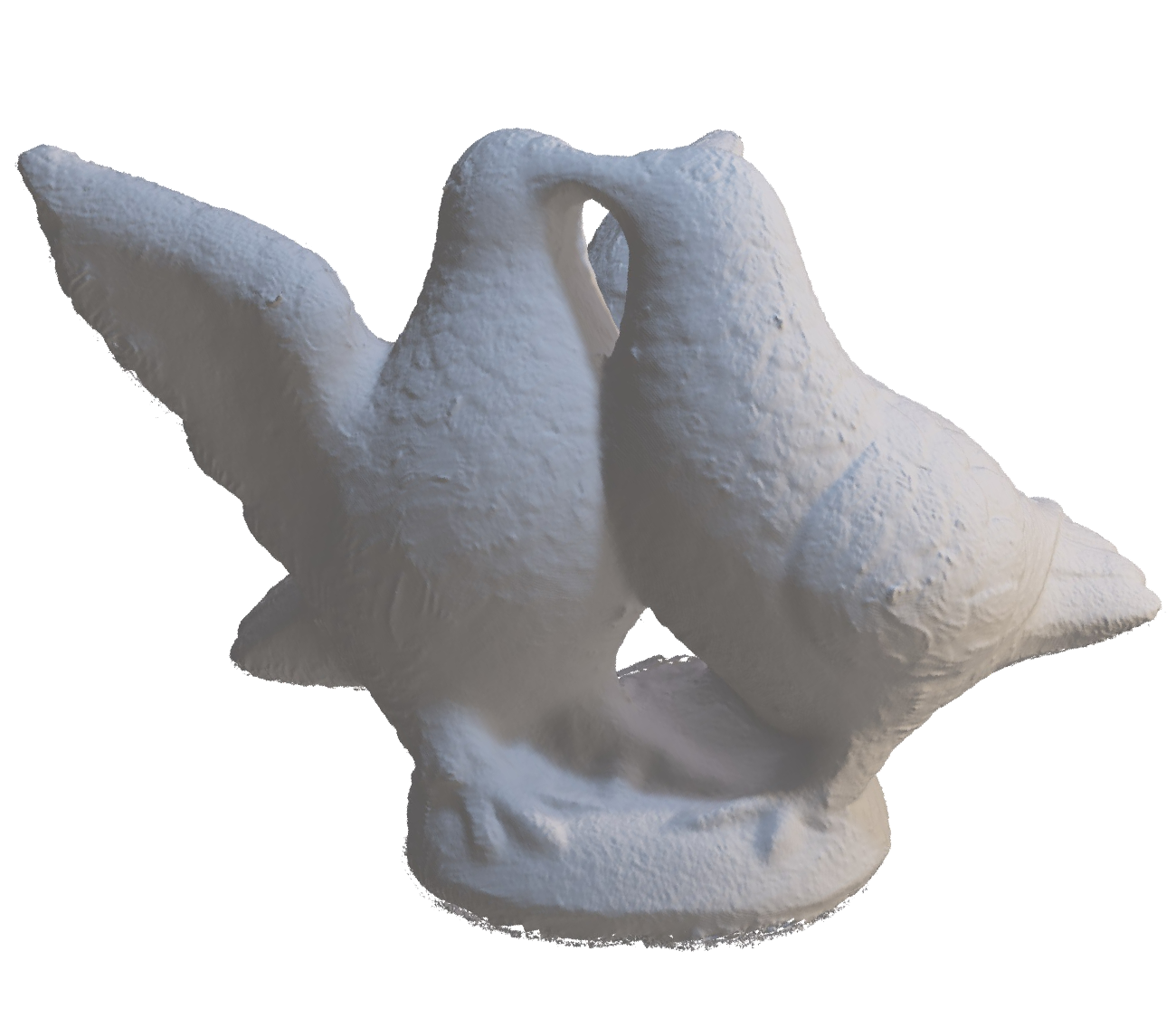}
    \end{subfigure}\hfill
    \begin{subfigure}{.19\linewidth}
        \centering
        \includegraphics[width=\linewidth]{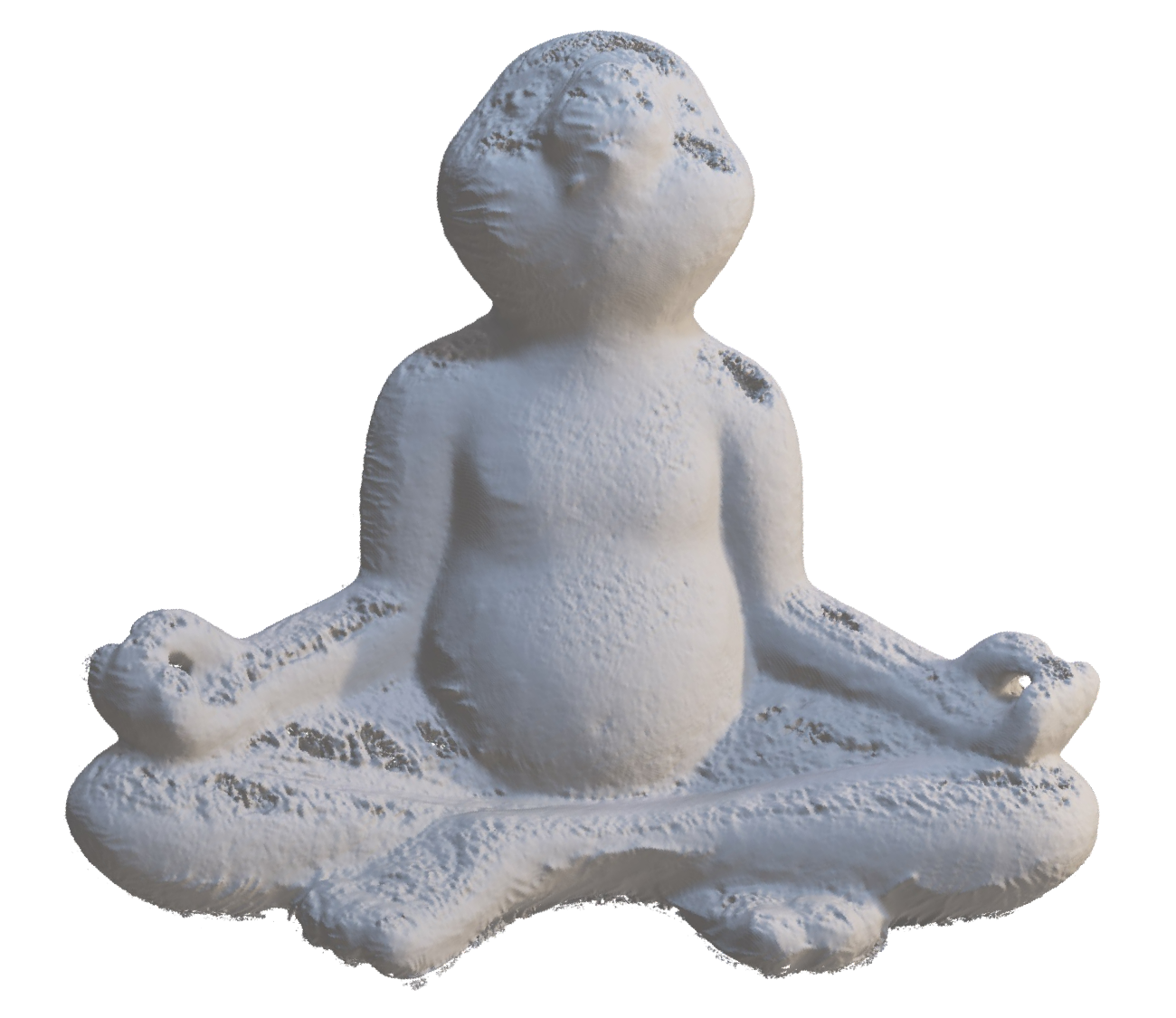}
    \end{subfigure}\hfill
    \begin{subfigure}{.19\linewidth}
        \centering
        \includegraphics[width=\linewidth]{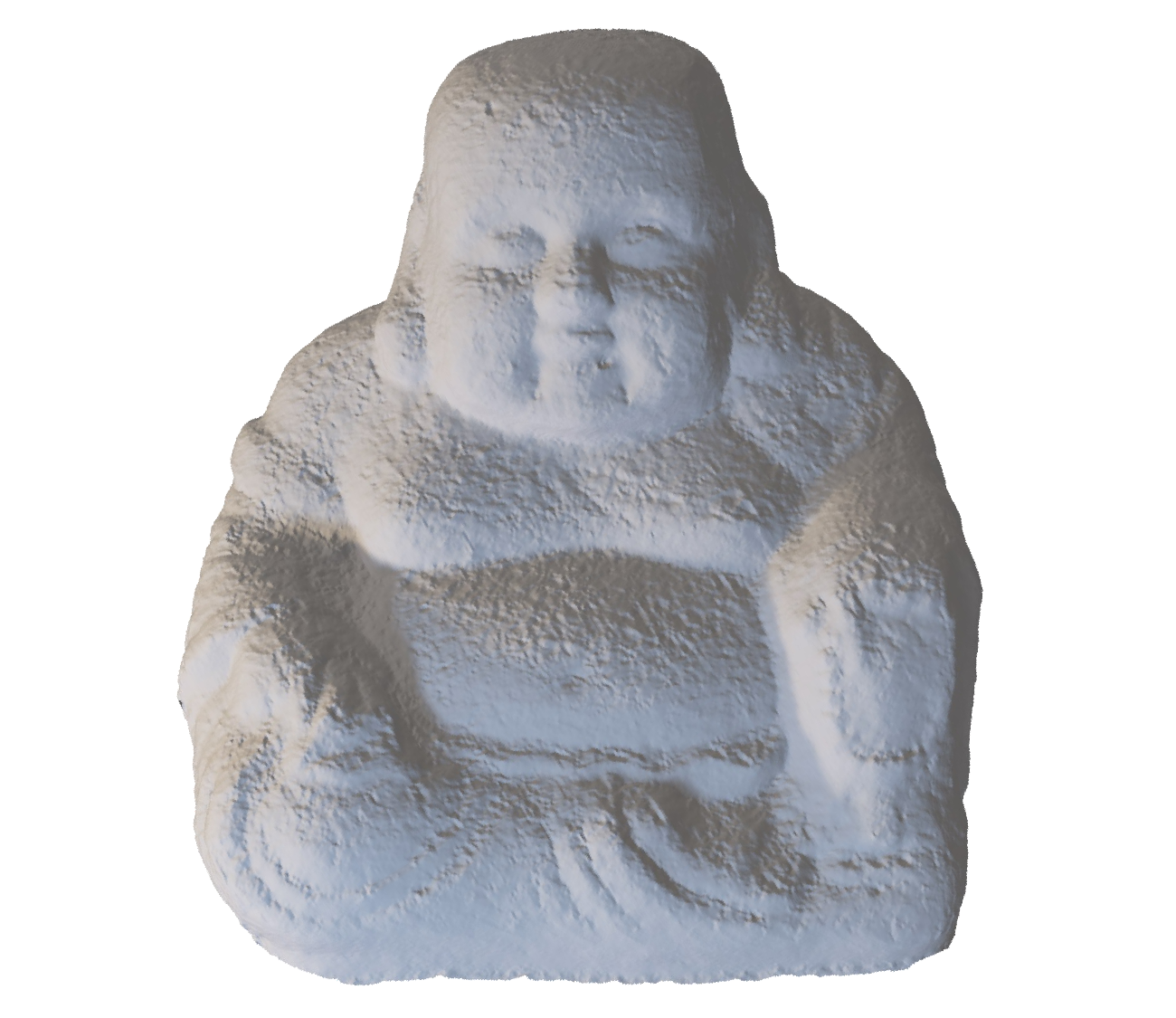}
    \end{subfigure}\hfill
    \begin{subfigure}{.19\linewidth}
        \centering
        \includegraphics[width=\linewidth]{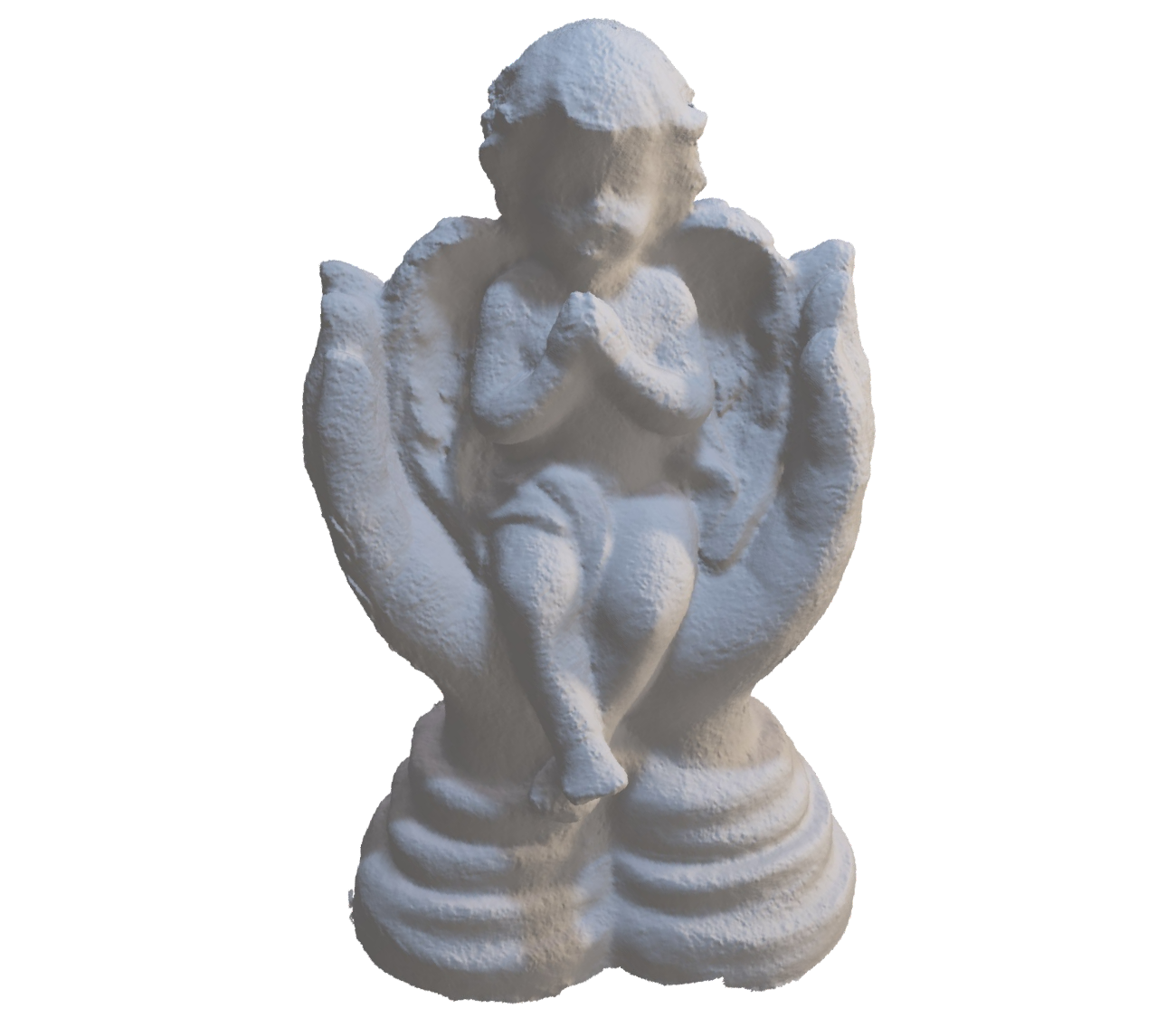}
    \end{subfigure}\hfill
    \begin{subfigure}{.19\linewidth}
        \centering
        \includegraphics[width=\linewidth]{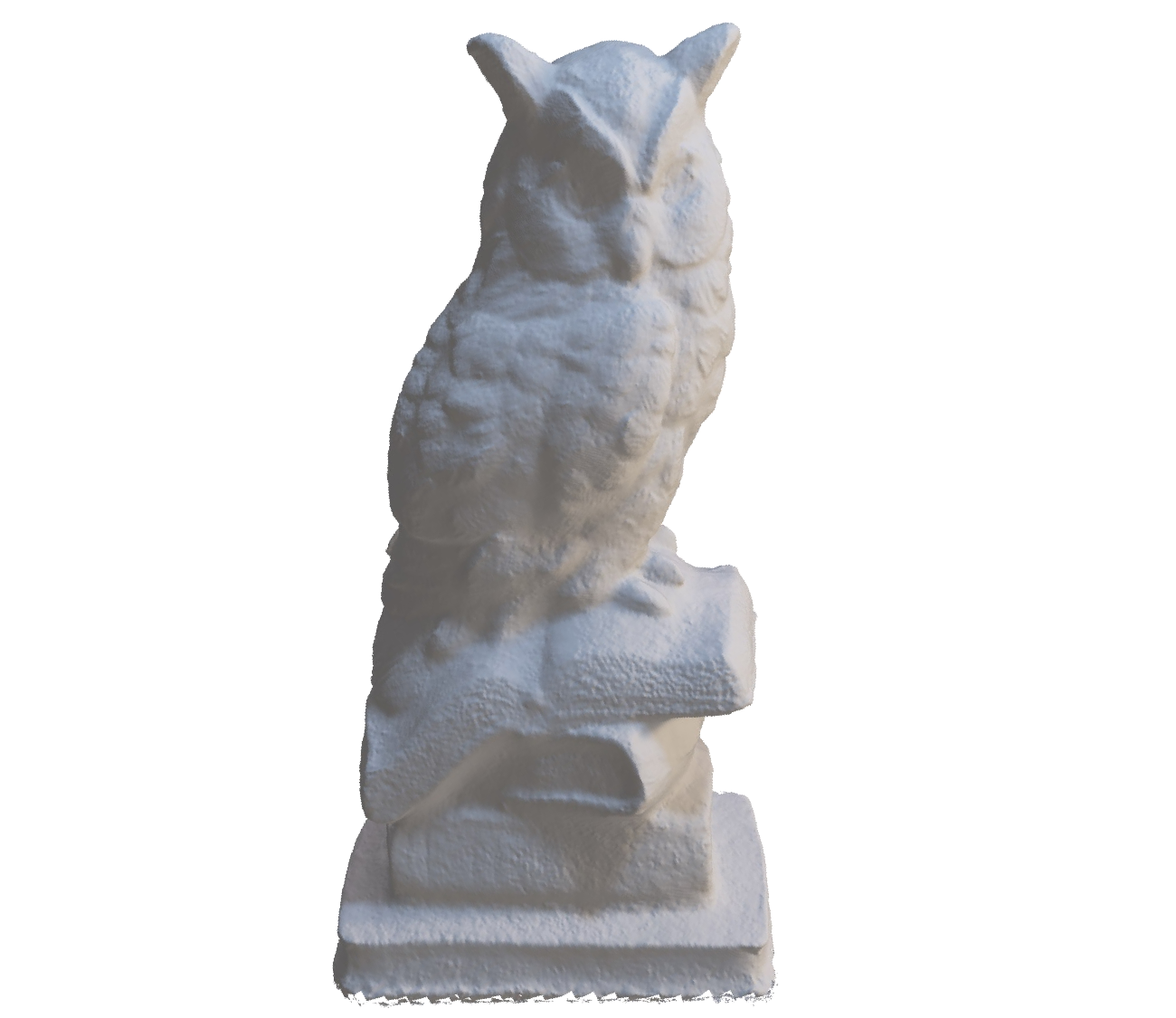}
    \end{subfigure}

    \caption{\textbf{Qualitative results of the reconstructed mesh on DTU.}
    We show 15 representative DTU scenes, where our method produces clean, watertight surfaces while preserving fine geometric details and sharp structures.}
    \label{fig:qual_mesh_dtu}
\end{figure*}

%% file: figs/10_1_full_qualitativeTnT/10_1_full_qualitativeTnT.tex
\begin{figure*}[t]
    \centering

    \begin{subfigure}{.32\linewidth}
        \centering
        \includegraphics[width=\linewidth]{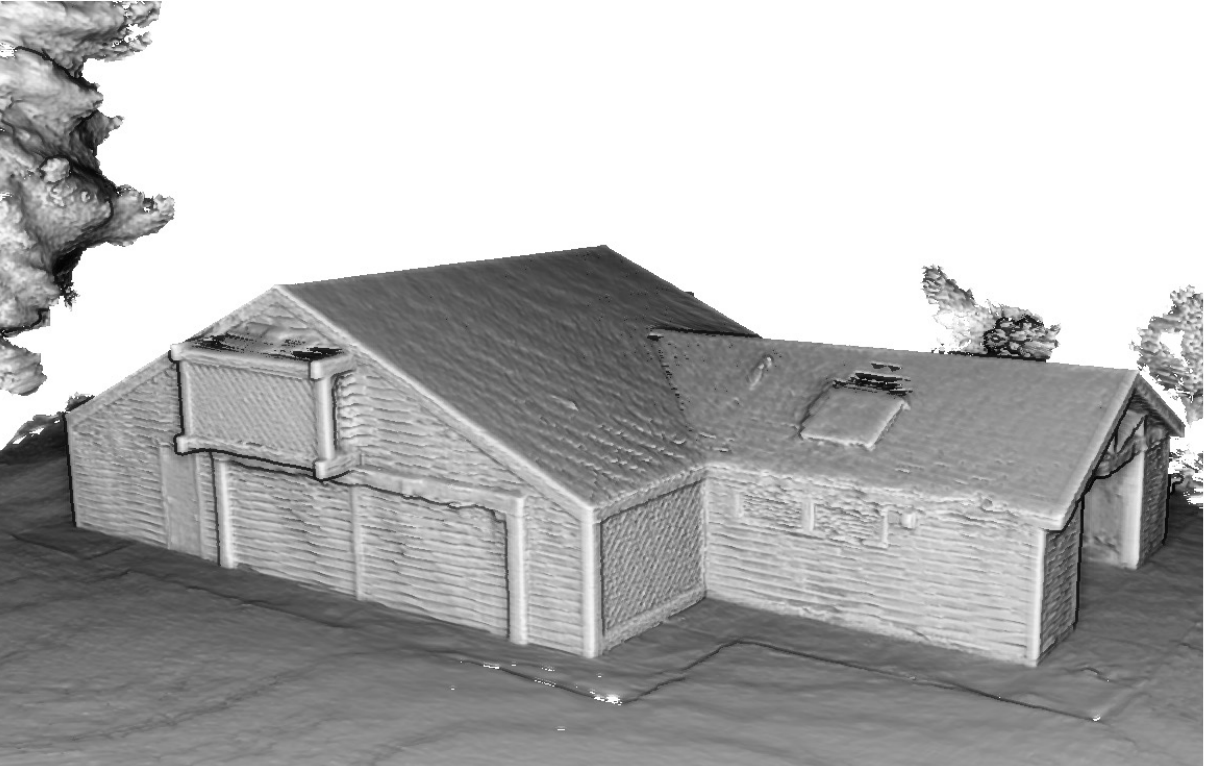}
    \end{subfigure}\hfill
    \begin{subfigure}{.32\linewidth}
        \centering
        \includegraphics[width=\linewidth]{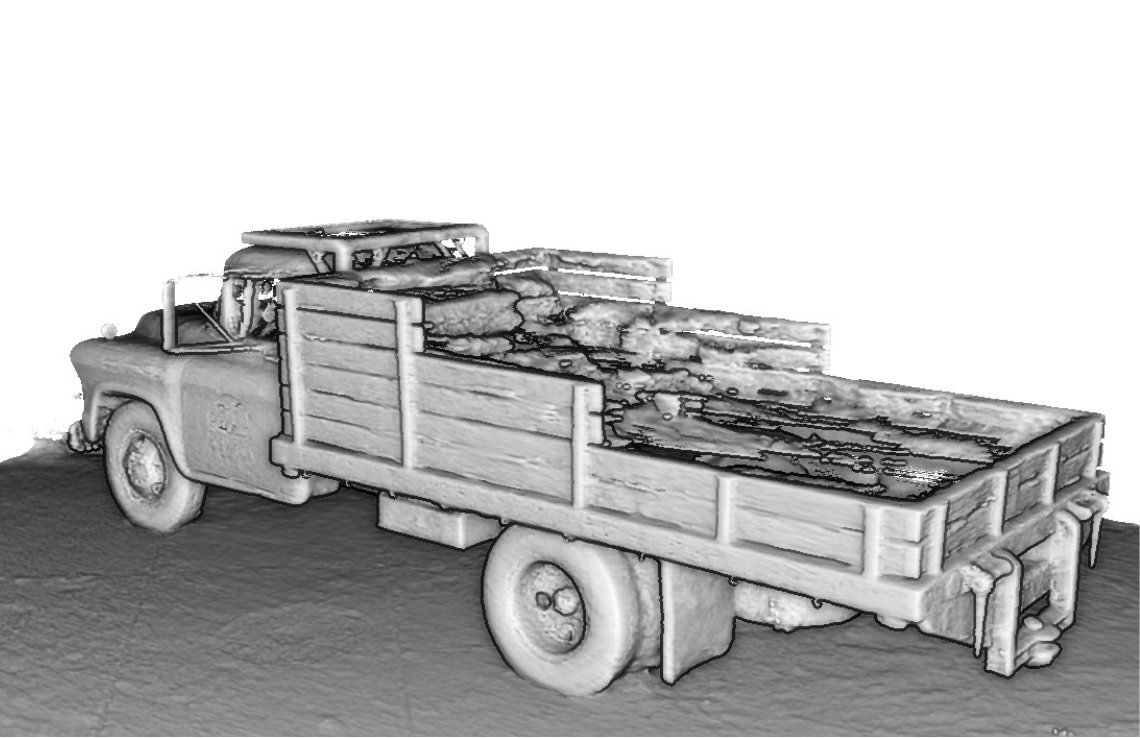}
    \end{subfigure}\hfill
    \begin{subfigure}{.32\linewidth}
        \centering
        \includegraphics[width=\linewidth]{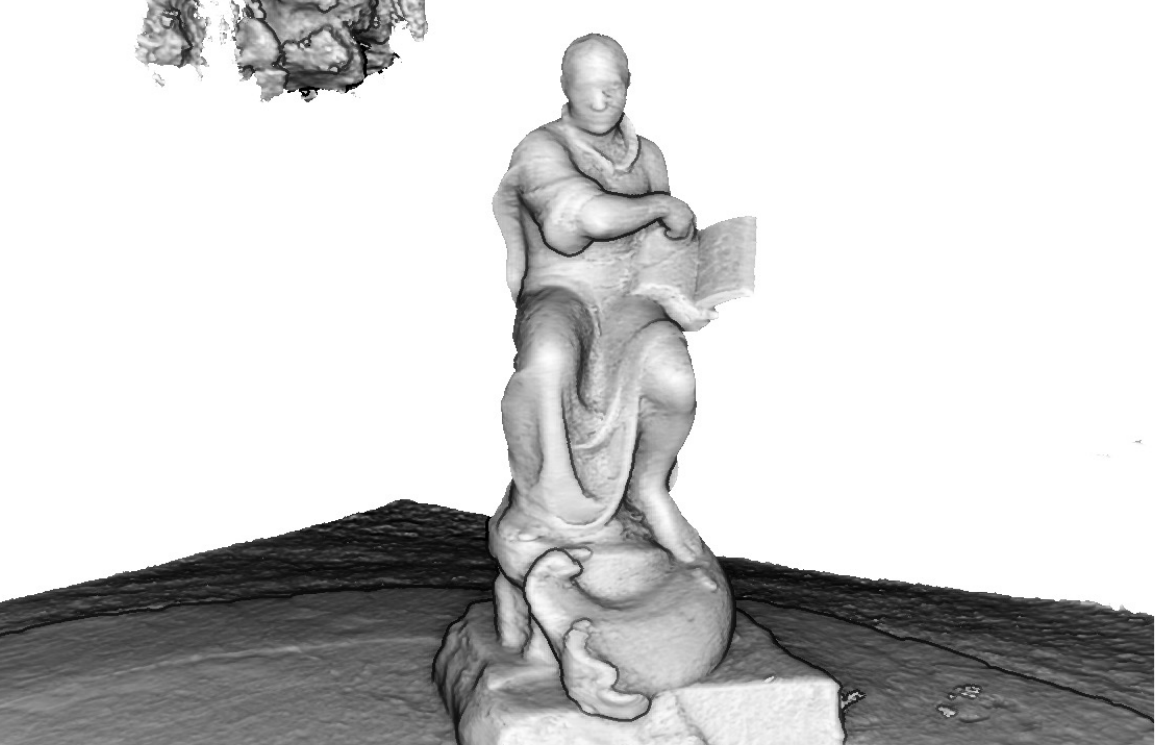}
    \end{subfigure}
    \\[0.3em]

    \begin{subfigure}{.32\linewidth}
        \centering
        \includegraphics[width=\linewidth]{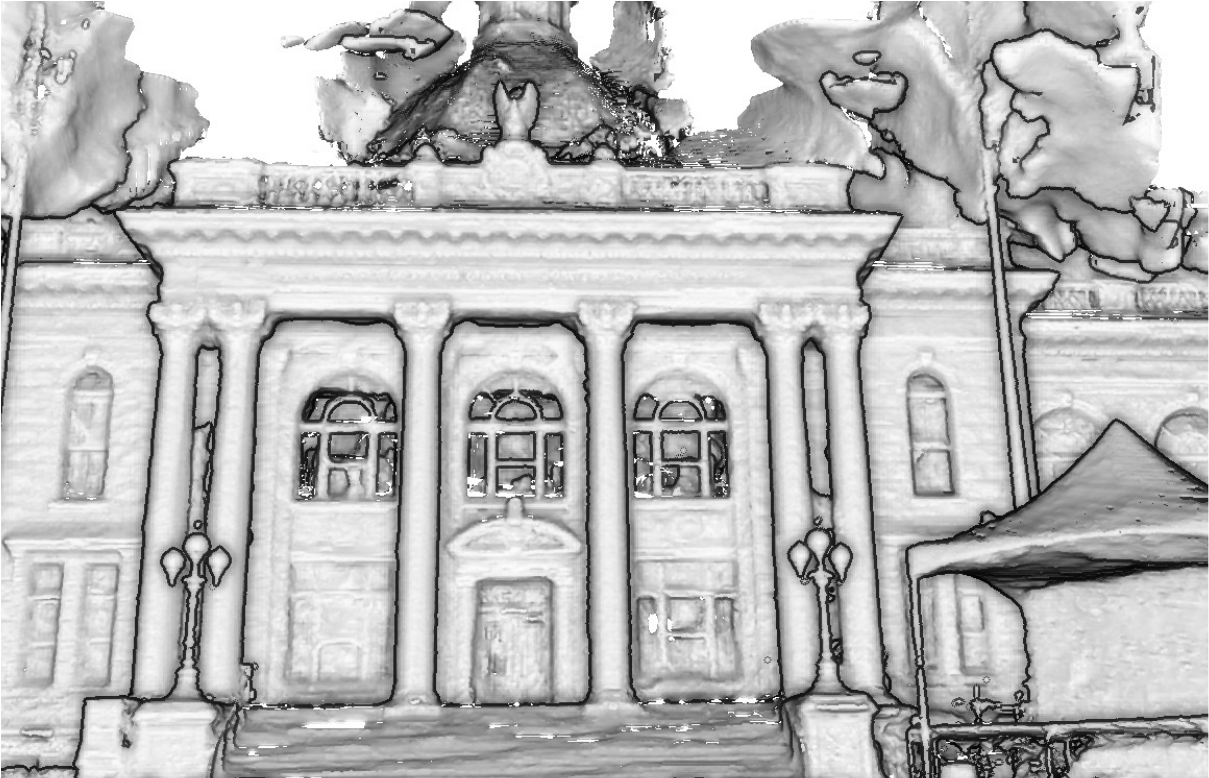}
    \end{subfigure}\hfill
    \begin{subfigure}{.32\linewidth}
        \centering
        \includegraphics[width=\linewidth]{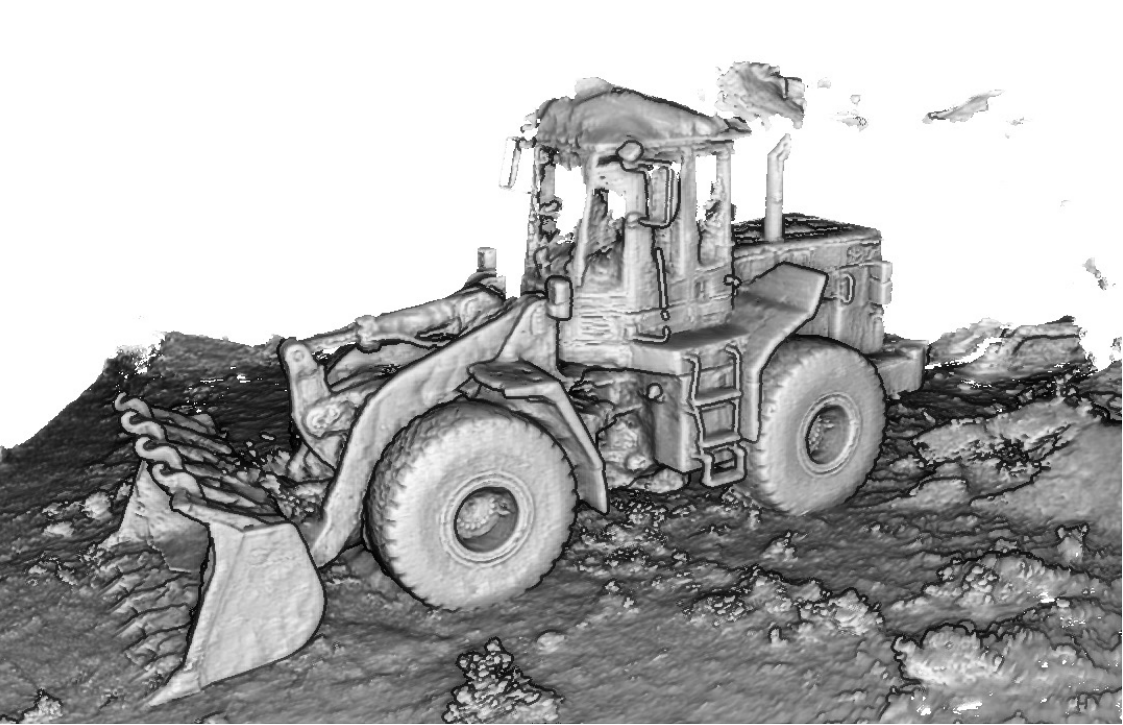}
    \end{subfigure}\hfill
    \begin{subfigure}{.32\linewidth}
        \centering
        \includegraphics[width=\linewidth]{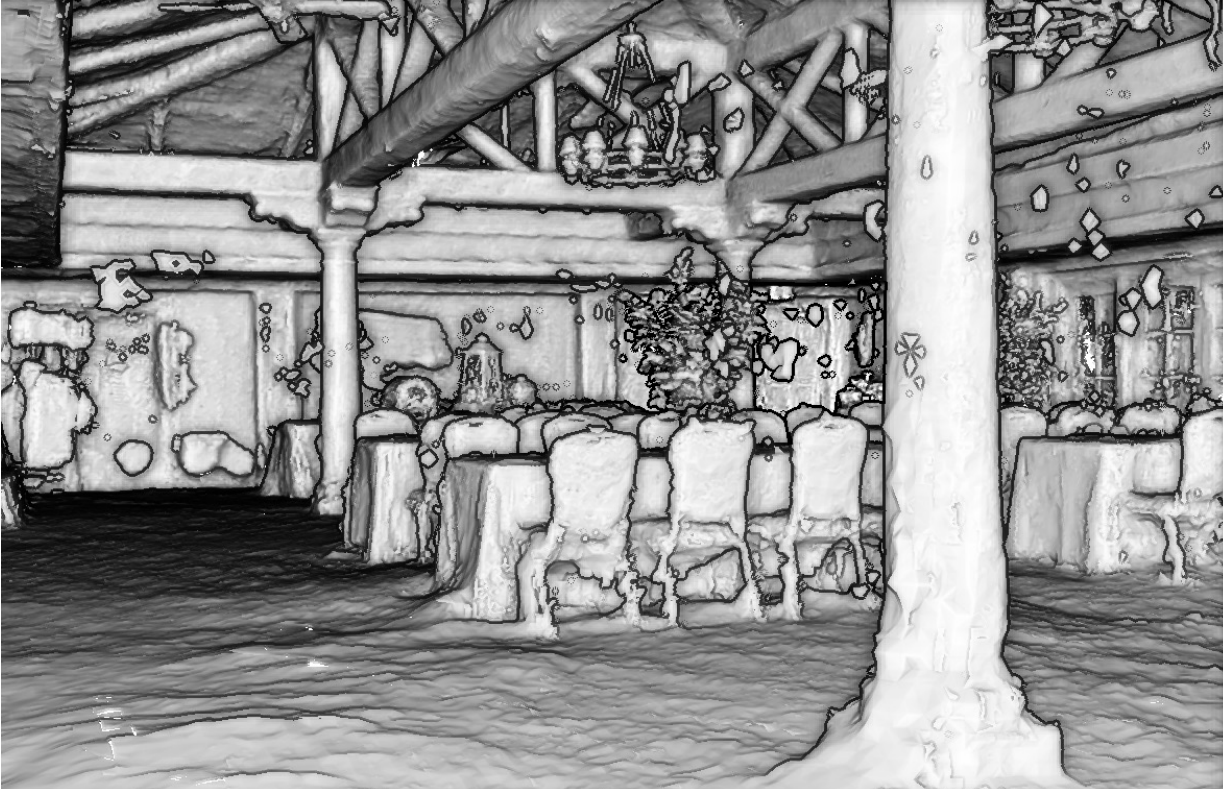}
    \end{subfigure}

    \caption{\textbf{Qualitative results of the reconstructed mesh on Tanks-and-Temples.}
    Our method scales to large, complex indoor and outdoor scenes (Barn, Truck, Ignatius, Courthouse, Caterpillar, Meetingroom), producing detailed and structurally consistent meshes.}
    \label{fig:qual_mesh_tnt}
\end{figure*}